\documentclass[Afour,sageh,times]{sagej}

\usepackage{moreverb,url}

\usepackage[colorlinks,bookmarksopen,bookmarksnumbered,citecolor=red,urlcolor=red]{hyperref}

\newcommand\BibTeX{{\rmfamily B\kern-.05em \textsc{i\kern-.025em b}\kern-.08em
T\kern-.1667em\lower.7ex\hbox{E}\kern-.125emX}}

\usepackage{natbib}
\usepackage{times}
\usepackage{graphics}
\usepackage{amsmath}
\usepackage{algorithm}
\usepackage[noend]{algpseudocode}
\usepackage{sidecap}
\usepackage{amsthm}
\usepackage{comment}
\usepackage{array}
\usepackage{enumitem}
\usepackage{booktabs}
\usepackage{multirow}
\usepackage{tabularx}
\usepackage{makecell}
\usepackage{colortbl}
\usepackage[dvipsnames]{xcolor}
\definecolor{lightgrey}{rgb}{0.8,0.8,0.8}
\definecolor{singlecontact}{rgb}{.7, 0.89, 0.78}  
\definecolor{paralleljaw}{rgb}{0.72, 0.77, 0.96}  
\definecolor{prior}{rgb}{0.96, 0.85, 0.96} 
\usepackage{threeparttable}
\usepackage[export]{adjustbox}
\usepackage{bm}
\usepackage{hhline}
\usepackage{subcaption}
\usepackage{enumitem}
\setlist[enumerate]{label*=\arabic*.}

\newcommand{\secref}[1]{Section~\ref{#1}}

\newcommand{\myparagraph}[1]{\vspace{0.05in}\noindent\textbf{#1}}

\newcommand{\hreff}[2]{\href[pdfnewwindow=true]{#1}{\nolinkurl{#2}}}

\newboolean{draft-mode}
\setboolean{draft-mode}{true}
\newcommand{\sidenote}[1]{\ifthenelse{\boolean{draft-mode}}{\marginpar{\tiny\raggedright\textsf{\hspace{0pt}#1}}}{}}
\DeclareRobustCommand{\arnote}[1]{\ifthenelse{\boolean{draft-mode}}{\textcolor{red}{\textbf{AR: #1}}}{}}
\DeclareRobustCommand{\ncdnote}[1]{\ifthenelse{\boolean{draft-mode}}{\textcolor{magenta}{\textbf{NCD: #1}}}{}}
\DeclareRobustCommand{\rhnote}[1]{\ifthenelse{\boolean{draft-mode}}{\textcolor{green}{\textbf{RH: #1}}}{}}

\newboolean{show-changes}
\setboolean{show-changes}{false} 
\newcommand{\change}[1]{\ifthenelse{\boolean{show-changes}}%
 {\textcolor{blue}{#1}}{#1}}

\newcommand{\methodd}{Tac2Pose }
\newcommand{\method}{Tac2Pose}

\begin{document}
\runninghead{Bauza et al.}

\title{Tac2Pose: Tactile Object Pose Estimation from the First Touch. }

\author{Maria Bauza*, Antonia Bronars*, and Alberto Rodriguez}
\affiliation{Massachusetts Institute of Technology, Cambridge, MA 02139, USA \\ * Authors with equal contribution.}

\corrauth{Maria Bauza, \\
Massachusetts Institute of Technology, Cambridge, MA 02139, USA.}
\email{bauza@mit.edu}

\begin{abstract}
In this paper, we present \method, an object-specific approach to tactile pose estimation from the first touch for known objects. Given the object geometry, we learn a tailored perception model in simulation that estimates a probability distribution over possible object poses given a tactile observation. To do so, we simulate the contact shapes that a dense set of object poses would produce on the sensor. Then, given a new contact shape obtained from the sensor, we match it against the pre-computed set using an object-specific embedding learned using contrastive learning. We obtain contact shapes from the sensor with an object-agnostic calibration step that maps RGB tactile observations to binary contact shapes. This mapping, which can be reused across object and sensor instances, is the only step trained with real sensor data.
%
%
%
%
%
%
%
This results in a perception model that localizes objects from the first real tactile observation. 
Importantly, it produces pose distributions and can incorporate additional pose constraints coming from other perception systems, multiple contacts, or priors.

We provide quantitative results for 20 objects. 
\methodd provides high accuracy pose estimations from distinctive tactile observations while regressing meaningful pose distributions to account for those contact shapes that could result from different object poses. 
We extend and test \methodd in multi-contact scenarios where two tactile sensors are simultaneously in contact with the object, as during a grasp with a parallel jaw gripper.
We further show that when the output pose distribution is filtered with a prior on the object pose, \methodd is often able to improve significantly on the prior. This suggests synergistic use of \methodd with additional sensing modalities (e.g. vision) even in cases where the tactile observation from a grasp is not sufficiently discriminative. Given a coarse estimate of an object's pose, even ambiguous contacts can be used to determine an object's pose precisely.

We also test \methodd on object models reconstructed from a 3D scanner, to evaluate the robustness to uncertainty in the object model. We show that even in the presence of model uncertainty, \methodd is able to achieve fine accuracy comparable to when the object model is the manufacturer's CAD model. Finally, we demonstrate the advantages of \methodd compared with three baseline methods for tactile pose estimation: directly regressing the object pose with a neural network, matching an observed contact to a set of possible contacts using a standard classification neural network, and direct pixel comparison of an observed contact with a set of possible contacts.

Website: \hreff{http://mcube.mit.edu/research/tactile_loc_first_touch.html}{mcube.mit.edu/research/tac2pose.html}
\end{abstract}


\maketitle
{\let\thefootnote\relax\footnote{{This paper extends the paper that appeared in the proceedings of the 2020  Conference on Robot Learning~\cite{bauza2020}.}}}
\vspace{-0.5in}

\section{Introduction}
\label{sec:introd}
Robotics history sends a clear lesson: accurate and reliable perception is an enabler of progress in robotics. 
From depth cameras to convolutional neural networks, we have seen how advances in perception foster the development of new techniques and applications.
For instance, the invention of high-resolution LIDAR fueled self-driving cars,  
and the generalization capacity of deep neural networks has dominated progress in perception and grasp planning in warehouse automation~\citep{zeng_2017,milan2018semantic,schwarz2018fast}.
The long term goal of our research is to understand the key role that tactile sensing plays in that progress. 
In particular, we are interested in robotic manipulation applications, where occlusions challenge accurate object pose estimation, and where object dynamics are dominated by contact interactions. 

\begin{figure*}[!ht]
\centering
\includegraphics[width=0.9\linewidth]{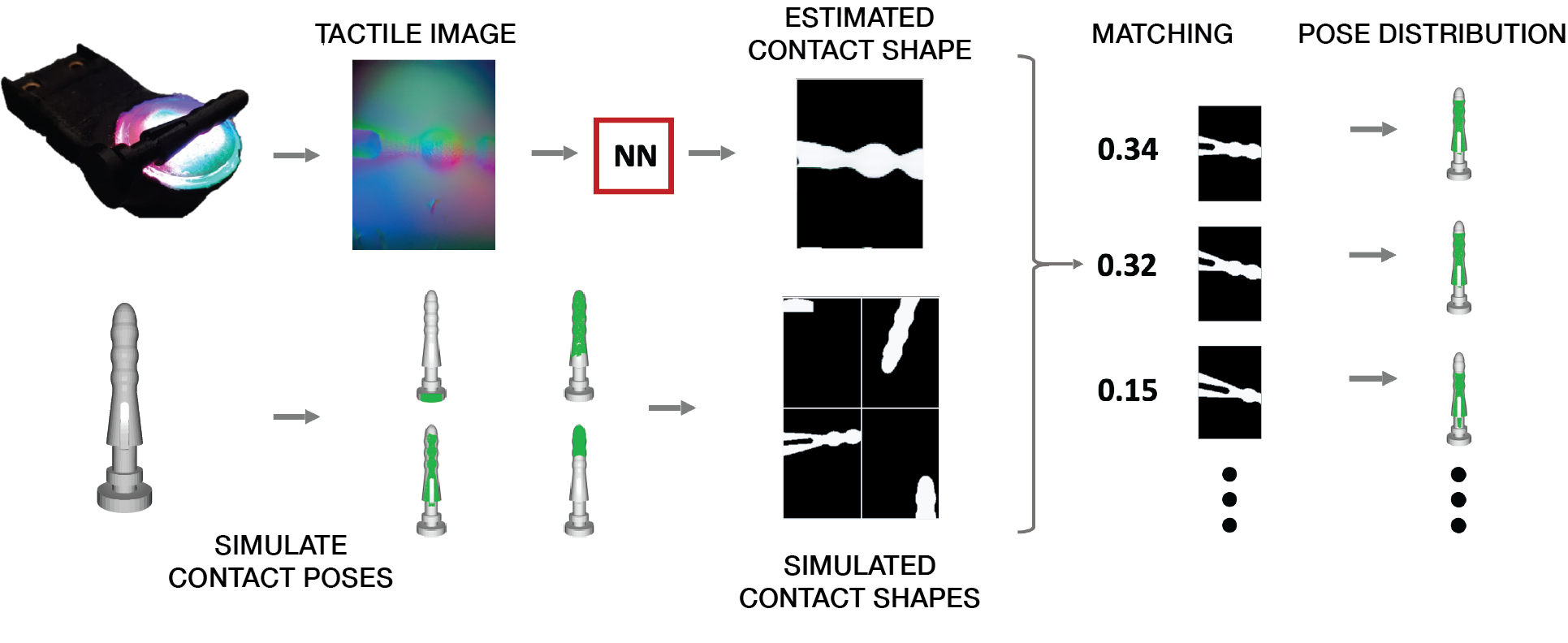}
\centering
\caption{\textbf{Tactile pose estimation with \method.} (Bottom row) In simulation, we render geometric contact shapes of the object from a dense set of possible contacts between object and tactile sensor. (Top row) The real sensor generates a tactile image from which we estimate its geometric contact shape. We then match it against the simulated set of contact shapes to find the distribution of contact poses that are more likely to have generated it. For efficiency and robustness, we do the contact shape matching in an embedding learned for that particular object. }\label{fig:motivation}

\end{figure*}

In this paper, we propose \method, a framework to estimate the pose of a touched object, as illustrated in Figure~\ref{fig:motivation}.
Given a 3D model of the object, \methodd learns an object-specific perception model in simulation, tailored at estimating the pose of the object from one--or possibly multiple--tactile images. 
%
%
As a result, the approach localizes the object from the first touch, i.e., without requiring any previous interaction. 
The perception model is based on merging two key ideas:~\looseness=-1
\begin{itemize}
    \item \textit{Geometric contact rendering:} we use the object model to render the contact shapes that the tactile sensor would observe for a dense set of contact poses. 
    \item \textit{Contact shape matching:} given the estimated contact shape from a tactile observation, we match it against a precomputed dense set of simulated contact shapes. 
    %
    %
    The comparison happens in an object-specific embedding for contact shapes learned in simulation using contrastive learning tools~\citep{moco}. 
    This provides robustness and speed compared to other methods based on direct pixel comparisons.
\end{itemize}
Enabled by the discriminative power of tactile sensing, the proposed approach is motivated by scenarios where the main requirement is estimation accuracy and where object models will be available beforehand. 
Many industrial scenarios fit this category.
We also consider cases where the object model is not known exactly, but reconstructed apriori with a 3D scanner. 
This broadens the set of scenarios to which \methodd can be applied.
Many previous solutions to tactile pose estimation require prior exploration of the object~\citep{li_2014,bauza2019tactile}.
Acquiring this tactile experience can be expensive, and in many cases, unrealistic. 
In this paper, instead, we learn the perception model directly from the object mesh model. 
The results in Sec.~\ref{sec:real_results}  show that the model learned in simulation directly transfers to the real world.~\looseness=-1
%

We attribute this to the object-specific nature of the learned model, the high resolution nature of the tactile sensors, and the contrastive-based perception framework. \methodd uses an intermediate representation of the tactile image, a binary contact mask, which we are able to generate with high fidelity both in simulation and using the real sensor. Furthermore, because \methodd matches observed contact shapes to a discrete set of simulated contacts, the match doesn't need to be perfect - just better than the other possible options - for the localization to be accurate.

Also key to the approach is that, by simulating a dense set of tactile imprints, the algorithm can reason over pose distributions, not only the best estimate.
The learned embedding allows us to efficiently compute the likelihood of each contact shape in the simulated dense set to match with the predicted contact shape from the tactile sensor. %
Predicting distributions is key given that tactile sensing provides local observations, which often do not discriminate pose globally (see, for example, the first contact in Figure \ref{fig:long_pencil_figure}).~\looseness=-1

Finally, by maintaining probability distributions in pose space, we can incorporate extra constraints over the likelihood of each pose.
We illustrate it in the case of multi-contact, where information from multiple tactile observations must be combined simultaneously, and in the case of filtering the distribution with a prior estimate of the object pose.
In practice, this prior estimate could come from previous tactile observations, kinematics, or other perception systems (e.g., vision).
%

We evaluate the performance of \methodd on real datasets collected for 20 different real objects of varying size and complexity in Section \ref{sec:real_results}. The labelled datasets evaluated in this work are available on the project website.
For 5 of the 20 objects, we also reconstruct the object models using a 3D scanner. We compare the localization accuracy on the same real datasets, when using a reconstructed model versus the manufacturer's CAD model, in Section \ref{sec:scans}.

Finally, we compare the performance of \methodd with three baseline methods for 5 objects in Section \ref{sec:baselines}: direct regression of the object pose using a neural network, matching an observed contact to a set of possible contacts using a standard classification network, and matching an observed contact to a set of possible contacts using direct pixel comparison.

In summary, the main contribution of this work is a framework for tactile pose estimation for objects with known geometry, with the following primary strengths:
\begin{itemize}
    \item[1.] Provides accurate pose estimation from the first touch, without requiring any previous interactions with the object.~\looseness=-1 
    \item[2.] Reasons over pose distributions by efficiently computing probabilities between a real contact shape and a dense set of simulated contact shapes. This allows for integration of pose constraints, such as those arising from multi-contact scenarios or prior estimates of the object pose.~\looseness=-1
\end{itemize}

We evaluate \method's accuracy with respect to object geometry, uncertainty (from reconstructed object geometry and tactile sensor noise), and other methods for tactile localization. In particular, we show: 

\begin{itemize}
    \item[1.] Quantitative results for 20 objects on three ablations of \methodd using real data: single contact, parallel jaw grasping, and parallel jaw grasping with a pose prior (Section~\ref{sec:real_results}). We achieve high localization accuracy when making contact with distinct object features. Complete results for each object can be seen in Table~\ref{table:main_all_results}.
    \item[2.] Comparisons for results on 5 objects using reconstructed object shapes instead of manufacturer's CAD models (Section~\ref{sec:scans}). The localization error on reconstructed models increases between $0.2$ to $1.5$mm compared to manufacturer's CAD models. Complete results can be seen in Table~\ref{table:scan_results}.~\looseness=-1
    \item[3.] Better performance of \methodd when compared with three baseline methods (Section \ref{sec:baselines}). Results on real contact shapes for 5 objects can be seen in Table~\ref{table:baseline_results}.
\end{itemize}

Finally, this paper extends our previous work on tactile localization~\citep{bauza2020} in several ways. First we provide results for 20 different objects (instead of 4) and analyze Tac2Pose in the case of a single contact, as well as when localizing objects using the two contacts from a parallel-jaw grasp, or when there is a prior pose distribution (that could come from vision). Moreover, we also greatly expand the number of possible contacts that we test per object, and the generality of grids that we match against, which simulates more closely a scenario with truly no prior information. This work also includes comparisons against sensible baselines, uses a better sensor and novel algorithm for extracting contact masks. Finally, we provide extensive details over the existence of non-unique contacts, the importance of 3D models (by analyzing the effect of using reconstructed rather than perfect object models), and quality of the resulting pose distributions, which further exposes the complexity of solving tactile localization.

\section{Related Work}
\label{sec:related}

Tactile sensing has been extensively explored in the robotics community. 
Relevant to this paper, this has resulted in the development of high-resolution tactile sensors and their use in a wide range of robotic manipulation applications. 
In this section, we review works that study tactile pose estimation and refer the reader to~\citep{Luo2017review} for a more in-depth review of other applications of tactile sensing.~\looseness=-1 

While we propose to use high-resolution tactile sensors that are discriminative and rich in contact information, most initial works in tactile localization were focused on low-resolution tactile sensors~\citep{schaeffer2003methods,corcoran2010tracking,petrovskaya2011,chalon2013online,Bimbo2016,saund2017touch,javdani2013efficient, chebotar2014learning}. 
%
%
Some works explore how to combine multiple tactile readings and reason in the space of contact manifolds~\citep{Koval2017, koval2015}. 
However, these are based on binary contact/no-contact signals, and require many tactile readings to narrow pose estimates.

Given the challenges from the locality of tactile sensing, recent works have gravitated towards two different approaches. 
Combining tactile and vision to obtain better global estimates of the object pose or using higher-resolution tactile sensors that can better discriminate different contacts. 
Among the solutions that combine vision and tactile, most rely on tactile sensors as binary contact detectors whose main purpose is to refine the predictions from vision~\citep{bimbo2015global, Allen1999,Ilonen2014, Falco2017,yu2018realtime}.

Other works, more in line with \method, have focused on using high-resolution tactile sensors as the main sensing source for object localization. 
Initial works in this direction
used image-based tactile sensors to recover the contact shape of an object and then use it to filter the object pose~\citep{platt2011using,pezzementi2011, Luo2017}. 
However, these approaches only provide results on planar objects and require previous tactile exploration.
There has also been some recent work on highly deformable tactile sensors for object localization~\citep{Kuppuswamy2019FastMC}.
These sensors are large enough to fully cover the touched objects, which eases localization. 

In this work, we use the image-based tactile sensor GelSlim 3.0~\citep{taylor2021}. 
The sensing capabilities of high-resolution sensors of this kind have already proven useful in multiple robotic applications, including assessing grasp quality~\citep{Hogan2018}, improving 3D shape perception~\citep{Wang2018} or directly learning from tactile images how to do contour following~\citep{lepora2019pixels} or tactile servoing~\citep{tian2019manipulation}.~\looseness=-1

For the task of tactile object localization, \citet{li_2014} proposed to extract local contact shapes from objects to build a map of the object and then use it to localize new contacts. 
The approach is meant to deal with small parts with discriminative features. 
Later \citet{izatt_2017} proposed to compute pointclouds from the sensor and use them to complement a vision-based tracker. 
Their tracker is fused with vision to deal with the uncertainty that arises from the locality of tactile sensing. 
In previous work~\citep{bauza2019tactile}, we proposed to extract local contact shapes from the sensors and match them to the tactile map of the objects to do object pose estimation.  
This approach requires the estimation of a tactile map for each object by extensively exploring them with the sensor. 

\citet{sodhi2020} estimated object pose during planar pushing from a stream of tactile imprints through a factor graph-based estimation framework. Their tactile observation model is trained to predict the relative pose of the object between a pair of non-sequential tactile images. This approach is designed to track the drift of an object from an initial well-known pose.
Similarly, \citet{sohdi2021} estimated 3D object pose over a contact sequence through a factor graph-based estimation framework. The tactile observation model maps from tactile images to surface normals using an image-to-image translation network, and uses ICP to determine the relative pose between 3D contact geometry in a contact sequence. This approach is designed to track the drift of object from an initial well-known pose, where the object itself is arbitrary and unknown. 
In comparison, \methodd assumes known object geometry and estimates the pose of the object from scratch, while also being able to exploit prior information when available. 

\methodd moves all object-specific computations to simulation and only requires an object-agnostic calibration step of the tactile sensor to predict contact shapes from RBG images from the GelSlim sensor~\citep{bauza2020}.
As a result, we can render in simulation contact shapes and learn object-specific models for pose estimation that translate well to the real world and achieve good accuracy from the first contact.~\looseness=-1

Recent work on tactile sensing applied to object insertion \citep{kim2021active} also uses an intermediate representation of the tactile image as the input to their insertion policy. Because this intermediate representation (pose of the contact line between the object and the cavity during an insertion attempt) can be simulated easily, the insertion policy can be trained entirely in simulation. Both this work and ours demonstrate the value of relying on geometric intermediate representations of contact shapes to allow high fidelity transfer of models trained on simulated data to real robotic systems.

Finally, \methodd to tactile pose estimation is related to methods recently explored in the computer vision community where they render realistic images of objects and learn how to estimate the orientation of an object given a new image of it~\citep{sundermeyer2018implicit,Jeon2020}. 
~\cite{DeepIM} designs a network to predict a relative pose between an observed image, and a rendered image of the object in a known pose. The method is used to iteratively improve on a coarse initial estimate of object pose.
~\cite{CosyPose} leverages a modified version of~\cite{DeepIM} to estimate the pose of multiple known objects from a sequence of images from multiple, unknown, viewpoints. The approach consists of predicting object poses relative to the camera within single views, then robustly matching objects and poses between views, before performing a scene-level refinement of both object and camera pose. Although this method does not reason explicitly over pose distributions for a single-view estimate of pose, it is able to handle occlusions and inaccurate single estimates by combining information from multiple viewpoints. ~\looseness=-1

Other methods address occlusion and inaccurate estimates by reasoning over pose distributions directly. ~\cite{PoseRBPF} estimates the 6D pose (3D translation, and a distribution over a discretization of 3D rotations) of an object in a Rao-Blackwellized particle filtering framework. The distributions generated using this method function as scores on a codebook of possible poses, rather than as well-defined probability distributions. \method, in contrast, regresses probability distributions which can be combined with distributions coming from additional contacts, sensing modalities, or priors.
%
%
%


\section{Method}

We present an approach to object pose estimation based on tactile sensing and known object models; illustrated in Fig.~\ref{fig:motivation}.
In an object-specific embedding, we match a dense set of simulated \textit{contact shapes} against the estimated contact shape from a real tactile observation. A contact shape is a binary mask over contact regions on the tactile sensor.
This results in a probability distribution over contact poses that can be later refined using other pose constraints. For example, we can combine information from the two contact shapes and the gripper opening obtained by grasping an object with a parallel jaw gripper.

We predict real contact shapes directly from the raw RGB tactile images that the sensor outputs (\secref{sec:tactile_mapping}). As seen in the top left quadrant of Figure \ref{fig:motivation}, we predict a binary mask over the region of contact (a.k.a. the contact shape) from an RBG tactile image with high fidelity.

The next steps of \methodd exploit the object model to estimate the contact pose, and are learned in simulation without using any real tactile observations. 
First, we render the contact shape for a given object pose, using the object model (\secref{sec:tactile_rendering}). Examples of the correspondence between object pose and contact shape obtained through geometric contact rendering can be seen in the bottom half of Figure \ref{fig:motivation}.
Next, we use geometric contact rendering to generate a dense set of object poses and their respective contact shapes, and use contrastive learning to learn an embedding to match contact shapes depending on the closeness of their object poses (\secref{sec:pose_estimation}). 
As a result, given the estimated contact shape from a real tactile observation, we can match it against this pre-computed dense set to obtain a probability distribution over contact poses. 

To predict contact poses beyond the resolution of the pre-computed set, it is possible to combine \methodd with registration techniques on the contact shapes. While we do not explore this refinement step in detail in this paper, we showed preliminary results on the efficacy of using FilterReg~\citep{gao2019filterreg} to further increase the localization resolution in~\citep{bauza2020}.
Finally, because our perception model outputs distributions over contact poses, is not restricted to single tactile observations. The distribution can be filtered or combined with other information coming from multi-contact scenarios, other sensing modalities, or additional pose constraints (\secref{sec:multi_contact}).

\subsection{Contact shape prediction from tactile observations}\label{sec:tactile_mapping}

Given a tactile observation, our goal is to extract the contact shape that produces it. 
To that aim, we train an image translation network (pix2pix) based on \citet{isola2018imagetoimage} to map RGB tactile observations to contact shapes. 
%
%
%
We train pix2pix using pairs of real RGB tactile images and their corresponding contact shapes. The training data is collected autonomously in a controlled 4-axis stage that generates planned touches on known 3D-printed shapes, following the approach proposed in~\citet{bauza2019tactile}. The dataset we use to train pix2pix contains 10,000 RGB tactile image/contact shape pairs from 32 distinct contact geometries. The dataset does not contain examples of contact geometries from any of the objects we use to evaluate tactile localization in Section \ref{sec:data_collection}.
Note that the map between RGB tactile images and contact shapes is independent of the object, and therefore we only need to gather labelled data once. Empirically, we found that a model trained on images from a single GelSlim sensor generalizes well across multiple instances of GelSlim sensors.~\looseness=-1

We provide further implementation details in ~\secref{sec:append_label} of the Appendix, including visualizations of the 32 contact geometries used for training (Figure~\ref{fig:data_collection}).

\subsection{Contact shape rendering in simulation}\label{sec:tactile_rendering}

Given the geometric model of an object and its pose w.r.t. the sensor, we use rendering techniques to simulate contact shapes from object poses (Figure~\ref{fig:motivation}, bottom). 
We refer to this process as \textit{geometric contact rendering}.
Below, we describe how we compute contact poses, i.e., object poses w.r.t. the sensor that would result in contact without penetration; and their associated contact shapes:

\begin{enumerate}
    \item We create a rendering environment using the open-source library Pyrender~\citep{pyrender}. 
    In this environment, we place a virtual depth camera at the origin looking in the positive z-axis. 
    The sensor can be then imagined as a flat surface (a rectangle in our case) orthogonal to the z-axis and at a positive distance $d$ from the camera. 
    \item We place the object in any configuration (6D pose) such that all points in the object are at least at a distance $d$ in the z-axis from the origin. We then compute the smallest translation in the z-axis that would make the object contact the surface that represents the sensor. 
    \item Finally, we move the object accordingly and render a depth image. 
    The smallest pixel value of the depth image corresponds to a depth $d$, and we consider that only pixels between distances $d$ and $d + \Delta d$ are in contact with the sensor, where $\Delta d$ represents the maximum deformation of the sensor. 
    The rest are marked as non-contact. 
    We convert the depth image to a binary image, where pixels with contact have a value of 1 and pixels with no contact have a value of 0. This binary image is the contact shape.
\end{enumerate}

For our sensor, depth images have a width and height of 640x640 (later re-scaled to 160x160 for faster to compute), the distance to the origin is $d=10$mm, and the maximum sensor deformation is $\Delta d = 1.3$mm. 
%

\subsection{Global tactile pose estimation}\label{sec:pose_estimation}

Once we know how to compute contact shapes both in simulation and from real tactile imprints, we reduce the problem of object pose estimation to finding what contact poses are more likely to produce a given contact shape. 
We solve this problem by first discretizing the space of possible contact poses as a parametrized grid, and then learning a similarity function that compares contact shapes.

\myparagraph{Object-dependent grids.}
\label{sec:grid} Using the 3D model of an object, we discretize the space of object poses in a multidimensional grid. 
Building a grid in the space of poses is a well-studied problem~\citep{yershova2010generating,rocsca2014new} that makes finding nearest neighbors trivial. 
It also allows each point on the grid to be seen as the representative of a volumetric part of the space which helps to reason over distributions. 
We prune the grid by only keeping object poses that result in contact, and then pair each of them with their respective contact shape.
Since we only consider poses that result in contact, this reduces the space of 6D poses to a 5D manifold. 
%
Using a grid, a discrete structured set of poses, allows us to easily account for object symmetries which can significantly reduce the grid size.

\begin{figure*}[h]
\centering
	\includegraphics[width=0.9\linewidth]{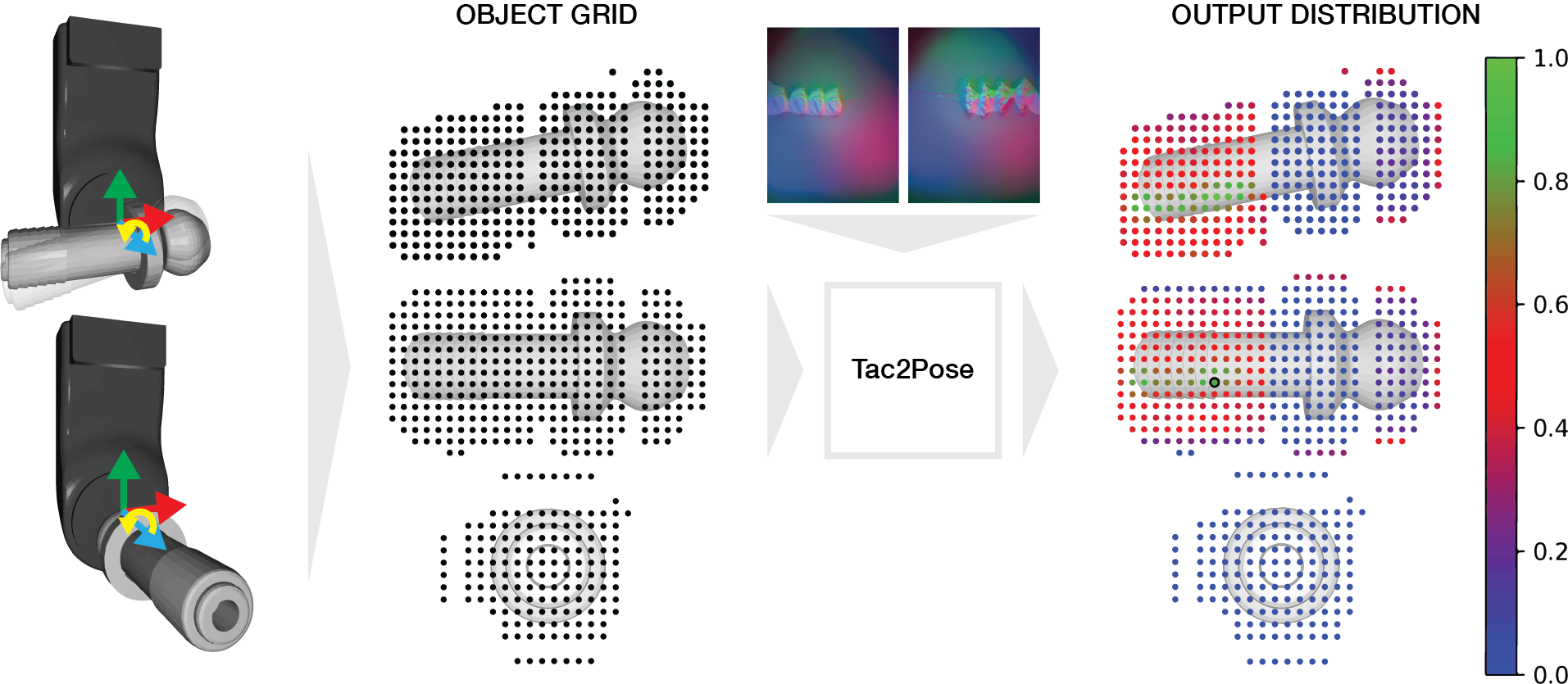}
\centering
\caption{\textbf{Object grids.} The four dimensions of the grid with respect to the object are visualized (left): {\fontfamily{lmtt}\selectfont grasp approach direction}, which is defined as the direction of the axis of the grasp (\textcolor{blue}{blue} arrow), x translation (\textcolor{red}{red} arrow), y translation (\textcolor{Green}{green} error), and angle in the plane of the grasp (\textcolor{yellow}{yellow} arrow). Samples of grid elements on the object {\fontfamily{lmtt}\selectfont long grease} are shown as black dots, where each black dot represents the center location of the gripper during a grasp. \methodd assigns likelihoods to each gripper location (right), given an observed contact (top). The most likely gripper locations for an observed contact are colored \textcolor{Green}{green}, while the least likely are colored \textcolor{blue}{blue}. The ground truth gripper location is shown as a black dot (right, center).
} \label{fig:grid dimensions}
\end{figure*}

Our grids cover regions of the object which correspond to feasible grasp locations from the set of stable resting poses of the object on a surface. 
In particular, we include faces that correspond to feasible \textit{grasp approach directions}, where the grasp approach direction specifies the axis of the grasp relative to the object. 
For some objects, we also include additional grasp approach directions that have interesting tactile features.
For each grasp approach direction, We compute a dense set of contacts with 2.5mm translational resolution, and 6 degrees of rotational resolution around the axis of the grasp. 
As a result, the grid spans four coordinates (grasp approach direction, x, y, $\theta$) and a contact pose will be no more than 1.25mm from an element of the grid, and often closer. 
A visualization of the grid dimensions can be seen in Figure~\ref{fig:grid dimensions}.
The number of elements in each grid varies between 3.8K and 181.3K, depending on the object size, shape, and number of grasp approach directions. We render each triplet of elements (two contacts, and the gripper opening) at 15 Hz, which can be parallelized. When grids are rendered serially, generating them takes from several minutes to many hours, depending on the grid size.

\myparagraph{Similarity metric for contact shapes.} 
Given a new contact shape, we compare it to all pre-computed contact shapes in the grid to find what poses are more likely to produce it. 
To that aim, we modify Momentum Contrast~(MoCo)~\citep{moco}, a widely-used algorithm in contrastive learning, to encode contact shapes into a low dimensional embedding based on the distance between contact poses. 

MoCo is able to learn unsupervised embeddings by building "a dynamic dictionary with a queue and a moving-averaged encoder". 
Instead, \methodd is supervised and the elements in the queue are fixed and assigned to each of the poses in the object's grid. Given a new contact shape, our model learns to predict the likelihood that each pose in the grid has produced the given shape. This likelihood is computed by taking the softmax over the distances between the embedding of the new contact shape and the embeddings saved in the queue, which correspond to the embeddings of each contact shape in the grid. Compared to the original MoCo, \methodd is supervised because during training, given a new contact shape, we can compute which element in the queue is closest to it.

To implement the encoder in our model, we use a ResNet-50~\citep{he2016} cropped before the average-pooling layer to preserve spatial information, making it a fully-convolutional architecture. 
The loss function is the categorical cross-entropy loss which allows us to ensure that the output of the softmax is a well-defined probability distribution. 
The training data comes from selecting a random contact pose and finding its closest element in the dense grid. 
Then, we use as desired probabilities a vector of all zeros except for the closest element which gets assigned to probability one (see Figure~\ref{fig:encoder}).
\secref{sec:append_sim} in the Appendix contains further details on the learning method.

Once we have created a dense grid and trained a similarity encoder for an object, given a new contact shape at test time, we can estimate which poses from the grid are more likely to generate it.
To run \methodd in real-time, we first encode the given contact shape and then compare it to all pre-computed encodings from the grid, which requires a single matrix-vector multiplication. Finally, we perform a softmax over the resulting vector of distances to obtain a probability distribution over the contact poses in the grid. We take the best match to be the contact pose with the highest probability, after (if applicable) incorporating additional pose constraints.~\looseness=-1
\begin{figure*}[ht]
\centering
	\includegraphics[width=0.9\linewidth]{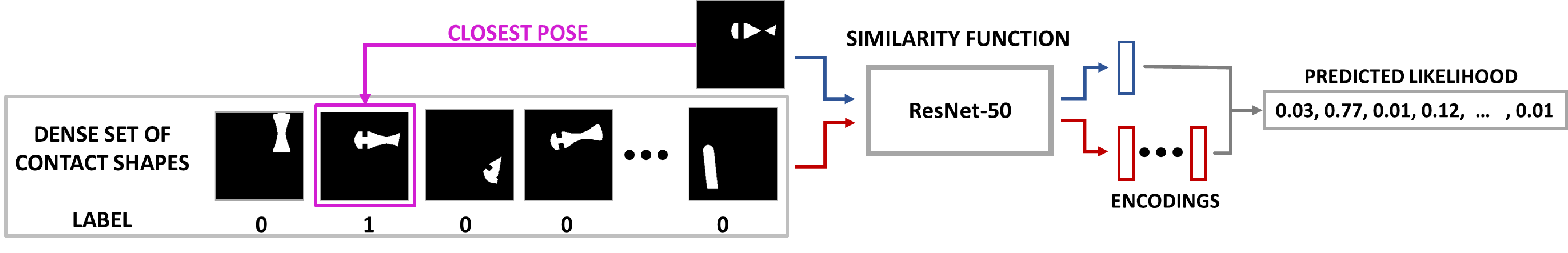}
\centering
\caption{\textbf{Similarity function.} We build a similarity function that learns to encode contact shapes into a low dimensional space and predicts, given a new contact shape, the likelihood of being the closest match of each contact shape in the pre-computed set. By learning an encoder for the contact shapes, we can compare them very efficiently.
} \label{fig:encoder}
\end{figure*} 

\subsection{Multi-contact pose estimation}\label{sec:multi_contact}

In this section, we show how to extend \methodd to multi-contact settings where we simultaneously need to reason over several tactile observations.  
In~\secref{sec:append_multi} of the Appendix we show that the likelihood of an object pose from the grid, $x$, given the estimated contact shapes, $C_{1,..., N}$, from N sensors is proportional to 
\begin{equation}
P(x | C_1, ... , C_N) \propto P(x | C_1) \cdot ... \cdot P(x | C_N) \cdot P(x) \label{eq:multi-contact}
\end{equation}
where $P(x)$ is a prior on the object pose, and the embedding network has been trained using uniformly-sampled poses. 
$P(x | C_i)$ is the likelihood that pose $x$ produces the contact shape $C_i$ on sensor $i$.
%
%
This allows combining \methodd with additional pose constraints such as the ones coming from kinematics, previous tactile observations, or other perceptions systems. 

The terms $P(x | C_i)$ come directly from computing the similarity function between $C_i$ and the contact shape from the grid of sensor $i$ that is closest to the contact pose $x$.
As a result, the computational cost of considering multiple contacts scales linearly with the number of sensors. In this paper, we consider the case of grasping with a parallel jaw gripper. We include information from two tactile images and the opening of the gripper during the grasp. We also consider the case of filtering the parallel jaw grasp distribution with a prior on the object pose.

\section{Results}
\label{sec:results}
\subsection{Real data collection}\label{sec:data_collection}

While \methodd is trained purely with simulated data, in this section we describe how it can provide accurate pose estimation when evaluated on real tactile data. 
To that aim, we design a system that collects tactile observations on accurately-controlled poses. 
Below we describe the tactile sensor, the robot platform, and the objects used to perform the experiments.

\myparagraph{Tactile sensor.} 
We consider the tactile sensor GelSlim 3.0 ~\citep{taylor2021} which provides high-resolution tactile readings in the form of RGB images. 
The sensor consists of a membrane that deforms when contacted and a camera that records the deformation. 
The sensor publishes tactile observations through ROS as 470x470 compressed images at a frequency of 90Hz (see top left portion of Figure \ref{fig:motivation}). 

\myparagraph{Robot platform.} 
We collect labelled datasets of tactile observations of grasps on 20 objects mounted in known positions and orientations in the world. Each dataset contains pairs of RGB tactile images, and the corresponding ground-truth object pose relative to the gripper.
%
The robotic system we use to collect the labelled datasets consists of a dual arm ABB Yumi with two WSG-32 grippers and GelSlim 3.0 tactile sensing fingers. We provide further details on the collection of labelled real data in \secref{sec:append_collect} of the Appendix.

We collect labelled observations of feasible grasps on each object. We compute the contacts in each dataset by considering the \textit{stable poses} of each object. Stable poses are the poses that an object is most likely to fall in when dropped onto a table. We then determine a set of feasible grasp approach directions for each stable pose. We define the grasp approach direction as the axis of the grasp during a parallel jaw grasp (Figure~\ref{fig:grid dimensions}). For some objects, we also include additional grasp approach directions that have interesting tactile features.
%
For each grasp approach direction, we collect a set of equispaced real observations, that correspond to a given grasp approach direction at various x,y locations relative to the object. In total, these observations correspond to the set of contacts that we are likely to encounter when picking up each object from a table.

\begin{figure*}[t]
\centering
	\includegraphics[width=0.9\linewidth]{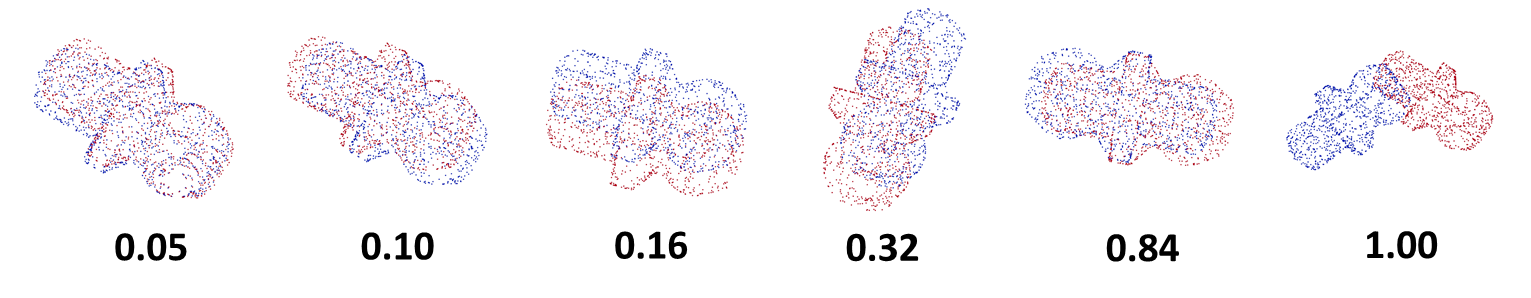}
\centering
\caption{\textbf{Normalized pose errors}, i.e., pose errors w.r.t. to the average random error, for the object \textit{grease}. The closest distance in the grid for this object is 0.03, similar to the first case. The median pose error when using parallel jaw contact information~(0.10) corresponds to an error like the second case , and the average random error~(1) would match with the last case. Finally, the example with 0.84 normalized error depicts a non-unique contact shape, i.e., the two object poses result in very similar contact shapes at the center that are not possible to distinguish without additional information.
} \label{fig:pose_error} 
\end{figure*}

\myparagraph{Objects.}
We test \methodd on 20 objects, derived from CAD models of objects available on McMaster. During the selection of objects, we aimed at covering object features that differently impact tactile localization. The features we considered include:~\looseness=-1
\begin{enumerate}
\item \textit{Contact non-uniqueness}: Contact non-uniqueness is the degree to which a contact is ambiguous, i.e., the contact in one pose resembles the contacts from other object poses (that are not equivalent under any symmetry). Non-unique contacts are more difficult to localize. An example of a non-unique contact on the object {\fontfamily{lmtt}\selectfont grease} is shown in Figure~\ref{fig:pose_error} (second from right).~\looseness=-1
\item \textit{Symmetry}: Object symmetries are sets of transformations under which the object pose is indistinguishable. Symmetry is a desirable property because it reduces the size of the grid.
\item \textit{Object size}: Larger objects are generally more challenging to localize with tactile sensing alone, since a single touch corresponds to a more local view of the object.~\looseness=-1
\item \textit{Contact size}: Large flat regions of contact with the sensor are generally more difficult to localize than smaller or textured ones. This is because most tactile sensors are less likely to produce crisp imprints for large flat contacts.
\end{enumerate}
We share the CAD models of each of the objects we use on the project website.
%

\subsection{ Real pose estimation results}
\label{sec:real_results}
We evaluate the accuracy of \methodd at estimating object poses from real tactile images. 
We note that \methodd, even for large grids (100K elements), can easily run at 50Hz allowing real-time estimation of pose distributions.
For each object, we evaluate \methodd on 90 to 400 (varies depending on object size and number of grasp approach directions) pairs of real tactile images and object poses per object using the approach described in~\ref{sec:data_collection}. 
%

Given the ground truth object pose and our best estimate (the object pose with highest likelihood), we measure the resulting \textit{pose error} (or distance between the two poses) by sampling a pointcloud of 10K points from the object 3D model and averaging the distance between these points when the object is at either of the two poses. 
This distance is sometimes called ADD (average 3D distance) but, for simplicity, we just refer to it as the pose error.
To compare errors across shapes and object sizes, we also compute the \textit{normalized pose error} which divides the original pose error by the average error obtained from predicting a random contact pose.
Figure~\ref{fig:pose_error} shows examples of different normalized pose errors.

For each object, we evaluate the pose error for three ablations of \methodd:
\begin{enumerate}
    \item \textit{Single Contact}: Estimate object pose from a single tactile image.
    \item \textit{Parallel Jaw}: Estimate object pose from a pair of tactile images, collected during a parallel jaw grasp on the object. The estimate also factors in the opening of the gripper during the grasp by computing the joint probability distribution of each object pose given both the two contacts and the gripper opening. In practice, this corresponds to multiplying the pose distribution obtained from the two contacts with a normal distribution centered around the gripper opening that each object pose would produce. The standard deviation of the normal distribution is 3mm which accounts for the expected error between the actual opening and the one reported by the gripper.
    \item \textit{Parallel Jaw + Prior}: Filter the distribution from the parallel jaw estimate to remove any poses that are more than a given pose distance from ground truth. This approximates cases in which an object pose is known within a margin of error, and is relevant when tactile localization is used to refine a coarse estimate of object pose from another sensing modality, like vision. We evaluate prior distances of 10mm and 5mm.
\end{enumerate}
The median error for each ablation for each object is shown in Table \ref{table:main_all_results}.

\begin{table*}[h]\caption{\textbf{Median error of the most likely pose.} We take the error corresponding to the most likely pose for each grasp on an object, and report the median. The \textcolor{blue}{blue} line next to each object is 20mm long, to show relative object scale. We consider the error when using a single contact, parallel jaw contacts, and parallel jaw contacts + a pose prior (prior distances of 10mm and 5mm). We report the median error in mm, and as a normalized error (in parenthesis). To normalize the distributions filtered with a prior, we use the expected random error from the filtered distribution. Objects that can be localized accurately (more than twice as good as random) with a single contact are colored in \textcolor{Green}{green}, objects that can be localized accurately after inclusion of parallel jaw information are colored in \textcolor{blue}{blue}, and objects that can be localized accurately after inclusion of parallel jaw information and a 10mm prior on the object pose are colored in \textcolor{violet}{purple}. Median values referenced in the text are bolded.}\label{table:main_all_results}\renewcommand{\arraystretch}{1.5}\begin{center}\begin{tabular}{|c|c|c c c c|}\bottomrule\multicolumn{2}{|c}{}&\multicolumn{2}{|c}{Tactile Only} & \multicolumn{2}{c|}{Pose Prior}\\\cline{3-6}\multicolumn{2}{|c|}{}&\makecell{\vspace{-10pt}\\\textbf{Single Contact}\\[1pt]mm (norm)} & \makecell{\vspace{-10pt}\\\textbf{Parallel Jaw}\\[1pt]mm (norm)} & \makecell{\vspace{-10pt}\\\textbf{10mm Prior}\\[1pt]mm (norm)} & \makecell{\vspace{-10pt}\\\textbf{5mm Prior}\\[1pt]mm (norm)}\\[1pt]\cline{1-6}Snap Ring &\makecell{\begin{minipage}{14mm}\vspace{3pt}\centering\includegraphics[height=8mm]{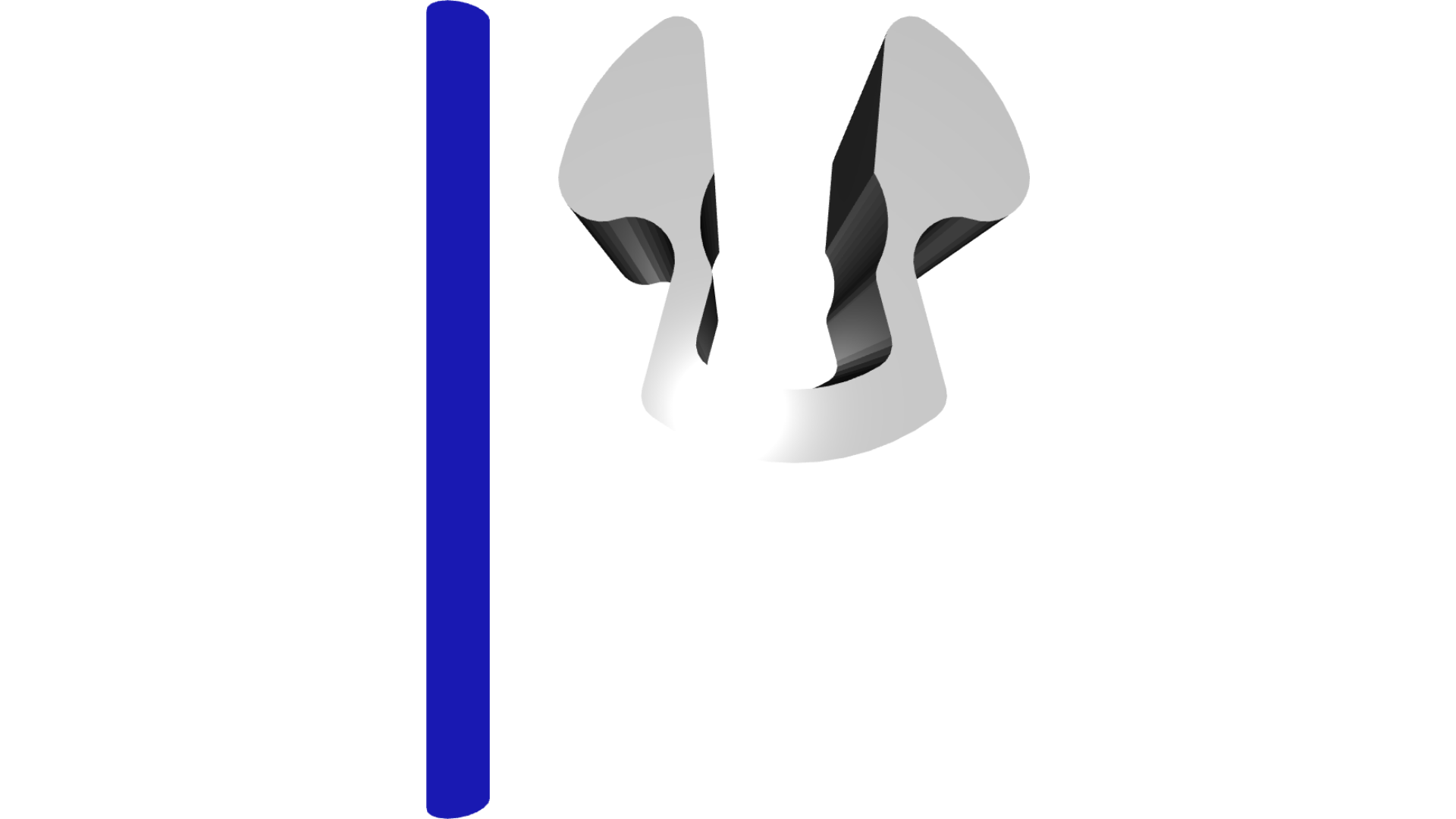}\vspace{3pt}\\\end{minipage}} & \cellcolor{singlecontact} \textbf{1.5 (0.10)}& 1.4 (0.10) & 1.4 (0.17) & 1.4 (0.38)\\Grease &\makecell{\begin{minipage}{14mm}\vspace{3pt}\centering\includegraphics[height=8mm]{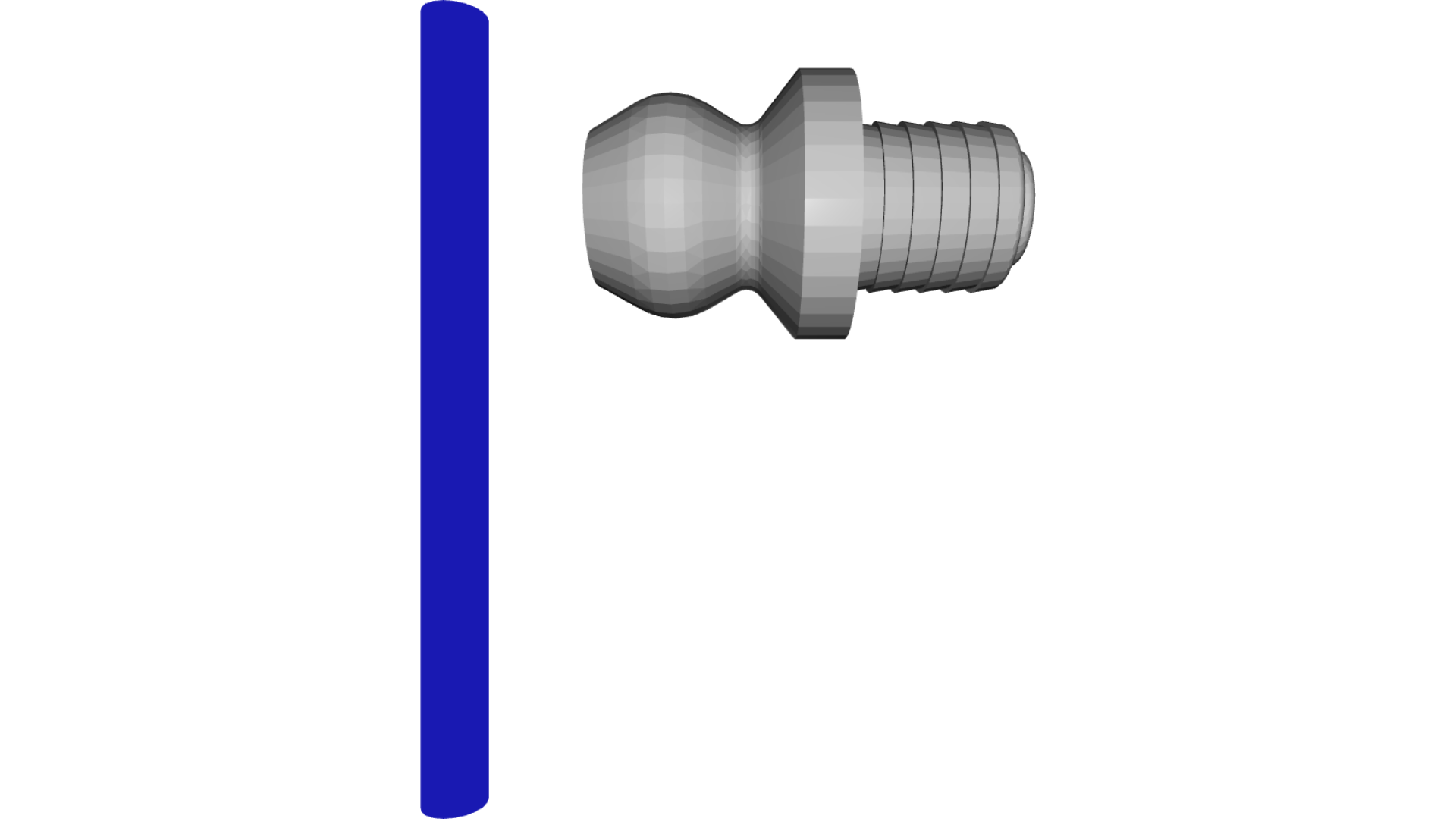}\vspace{3pt}\\\end{minipage}} & \cellcolor{singlecontact}1.3 (0.12) & 1.2 (0.10) & 1.2 (0.15) & 0.9 (0.26)\\Slotted Shim &\makecell{\begin{minipage}{14mm}\vspace{3pt}\centering\includegraphics[height=8mm]{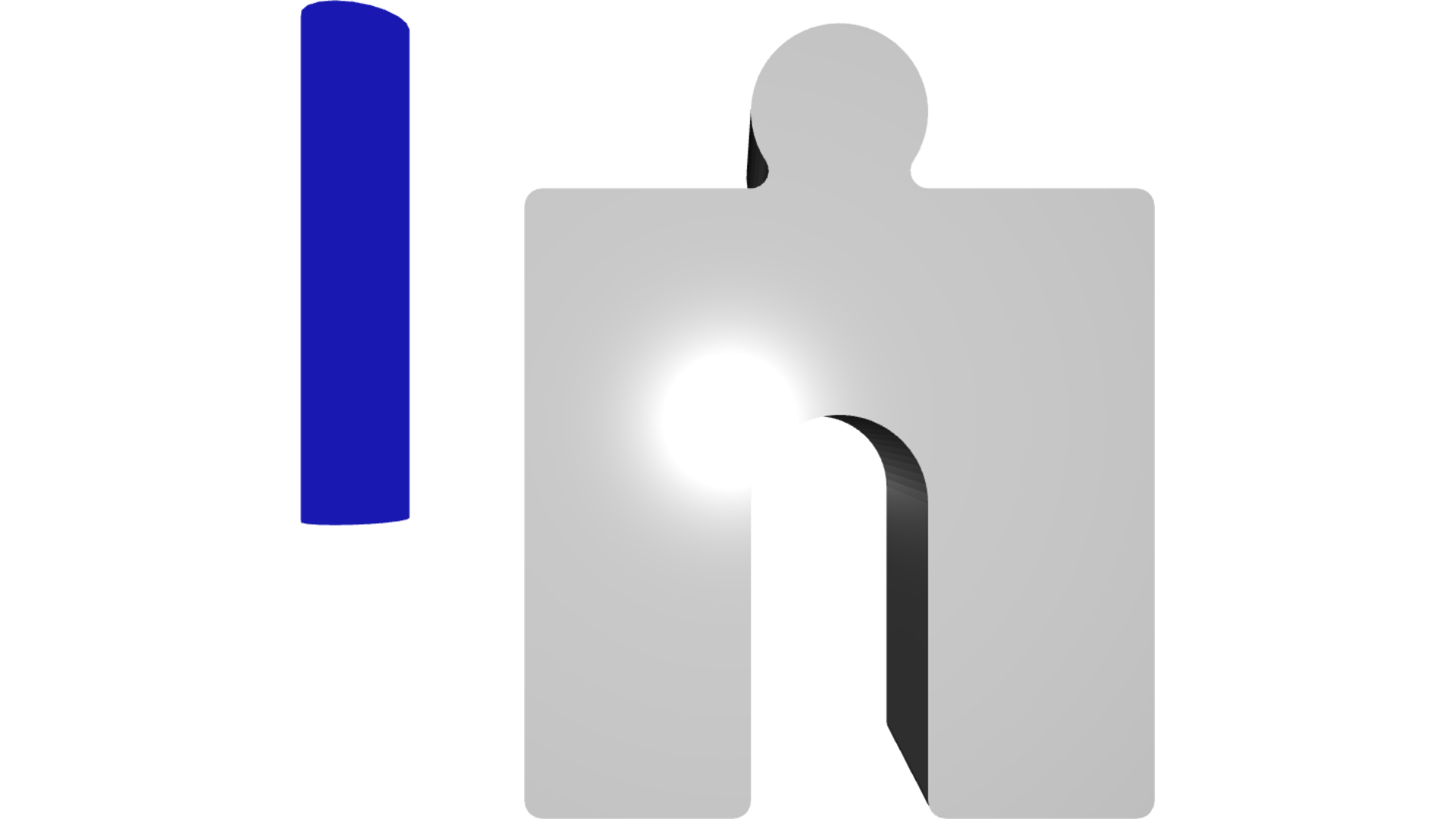}\vspace{3pt}\\\end{minipage}} & \cellcolor{singlecontact}4.0 (0.15) & 3.0 (0.12) & 2.4 (0.30) & 2.3 (0.57)\\Round Clip &\makecell{\begin{minipage}{14mm}\vspace{3pt}\centering\includegraphics[height=8mm]{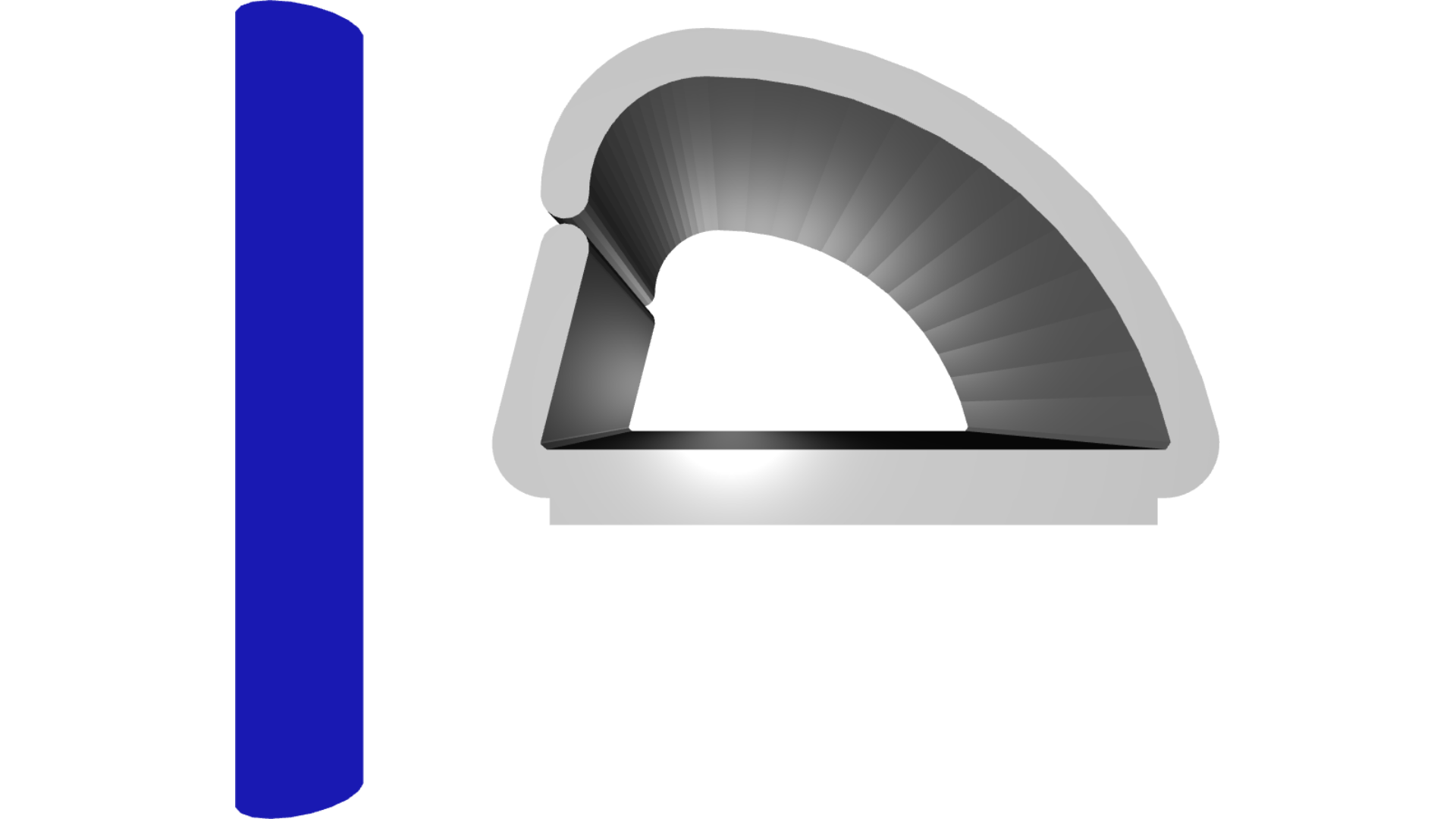}\vspace{3pt}\\\end{minipage}} & \cellcolor{singlecontact}3.4 (0.17) & 11.5 (0.58) & 2.0 (0.24) & 1.9 (0.50)\\Hanger &\makecell{\begin{minipage}{14mm}\vspace{3pt}\centering\includegraphics[height=8mm]{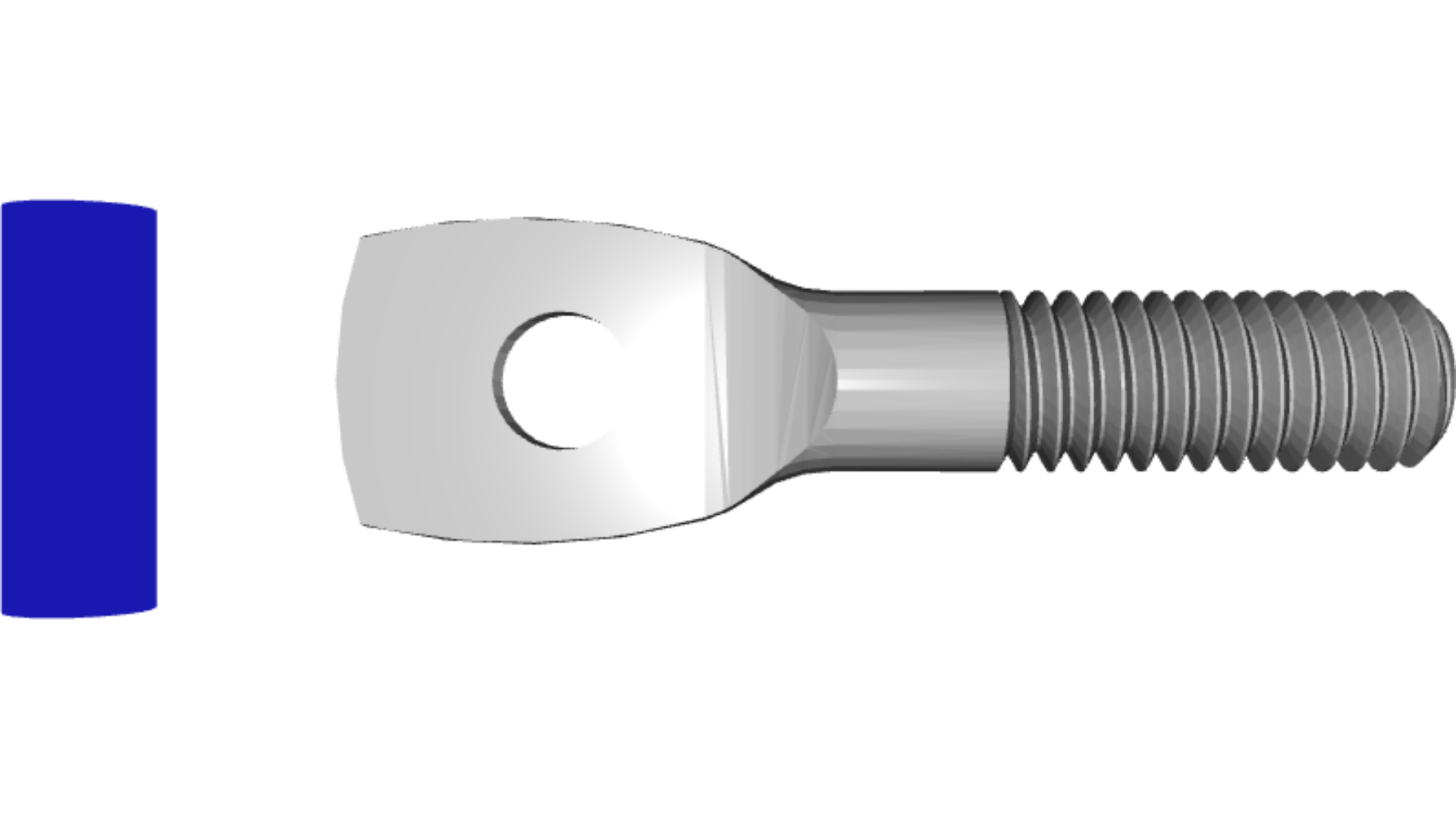}\vspace{3pt}\\\end{minipage}} & \cellcolor{singlecontact}6.6 (0.19) & 2.6 (0.07) & 2.4 (0.30) & 2.4 (0.57)\\Pin &\makecell{\begin{minipage}{14mm}\vspace{3pt}\centering\includegraphics[height=8mm]{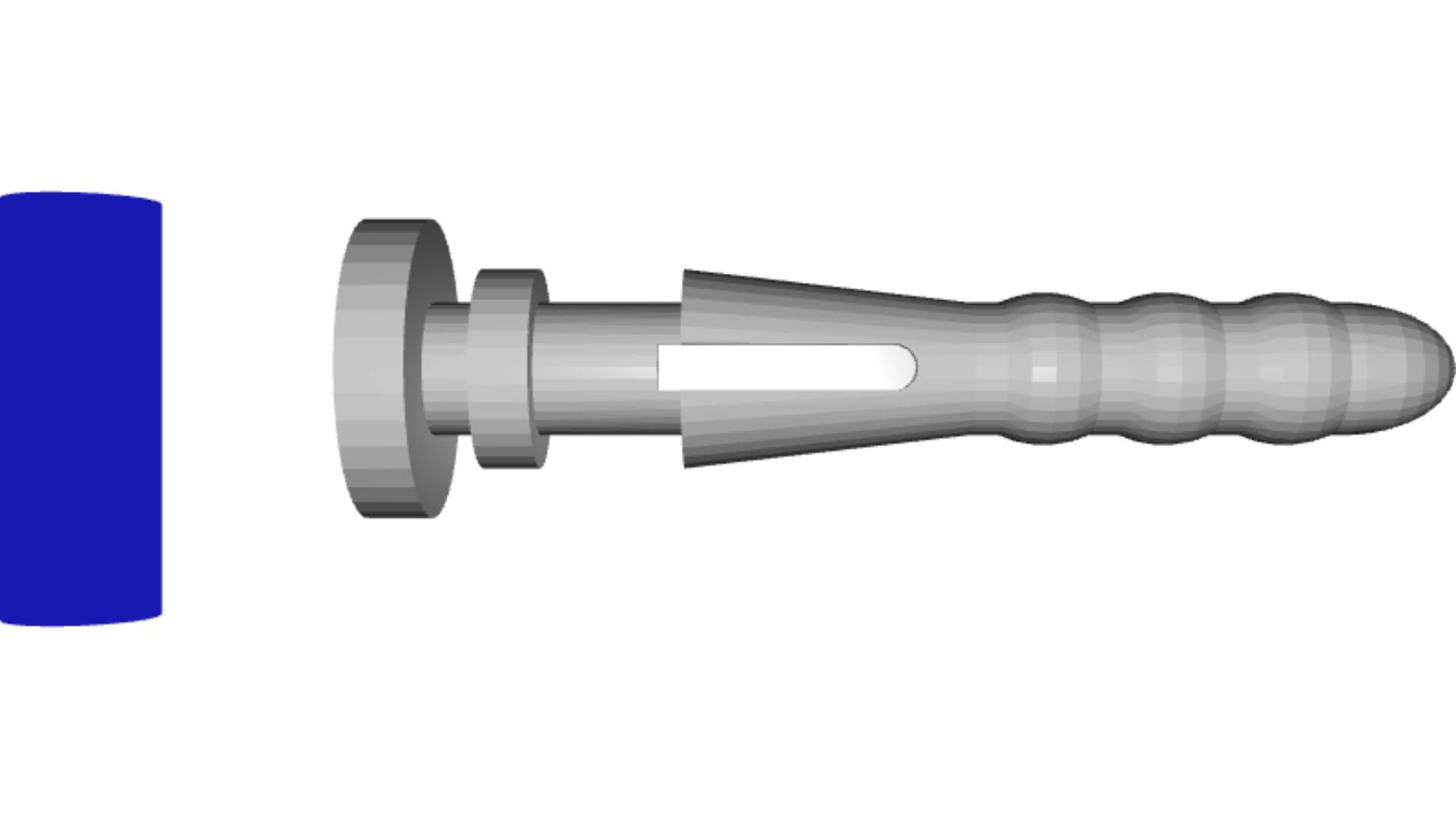}\vspace{3pt}\\\end{minipage}} & \cellcolor{singlecontact}6.5 (0.20) & 5.6 (0.17) & 4.7 (0.66) & 3.6 (1.02)\\Big Head &\makecell{\begin{minipage}{14mm}\vspace{3pt}\centering\includegraphics[height=8mm]{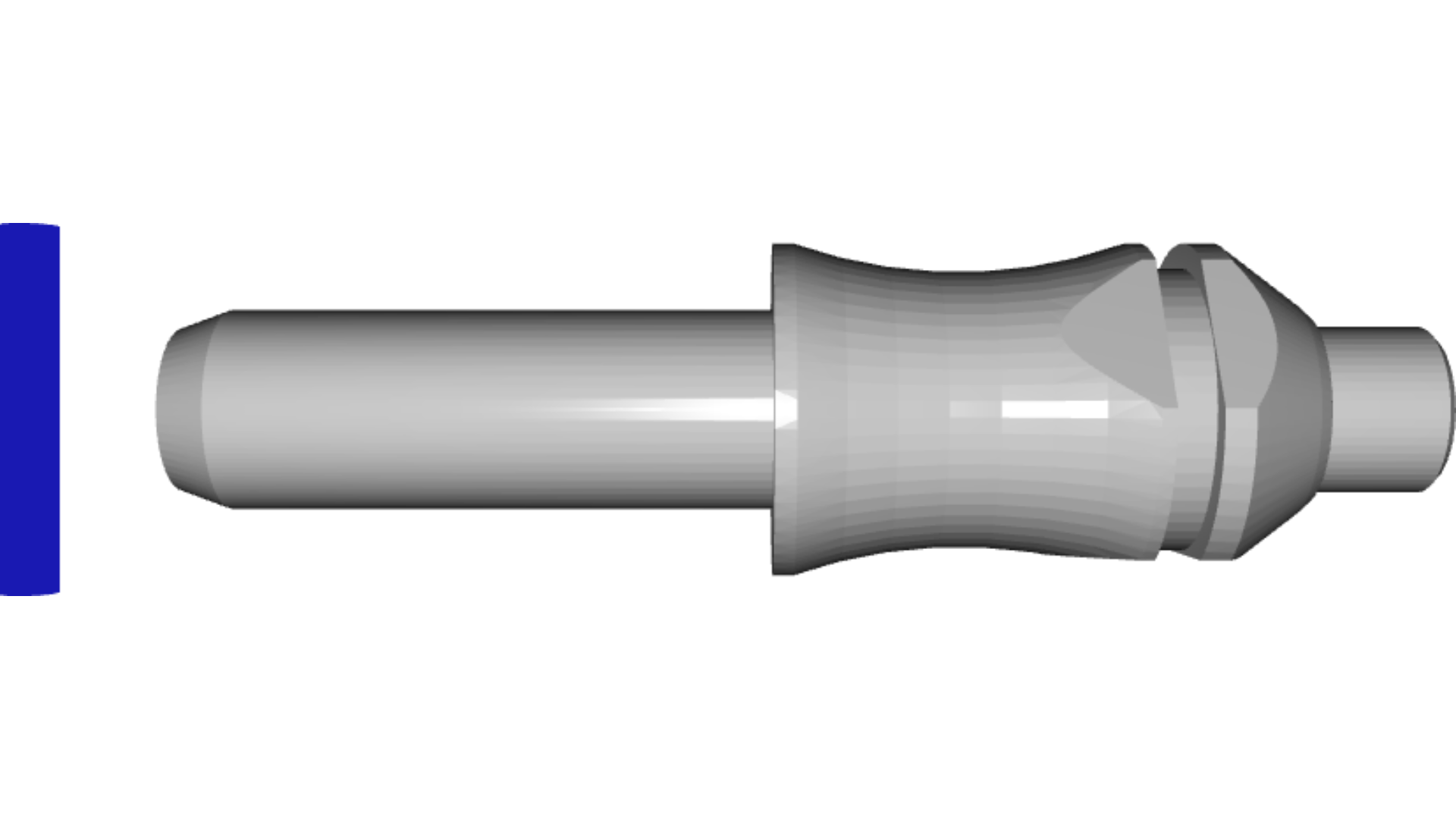}\vspace{3pt}\\\end{minipage}} & \cellcolor{singlecontact}7.8 (0.20) & 6.1 (0.16) & 4.9 (0.72) & 3.9 (1.19)\\Holder &\makecell{\begin{minipage}{14mm}\vspace{3pt}\centering\includegraphics[height=8mm]{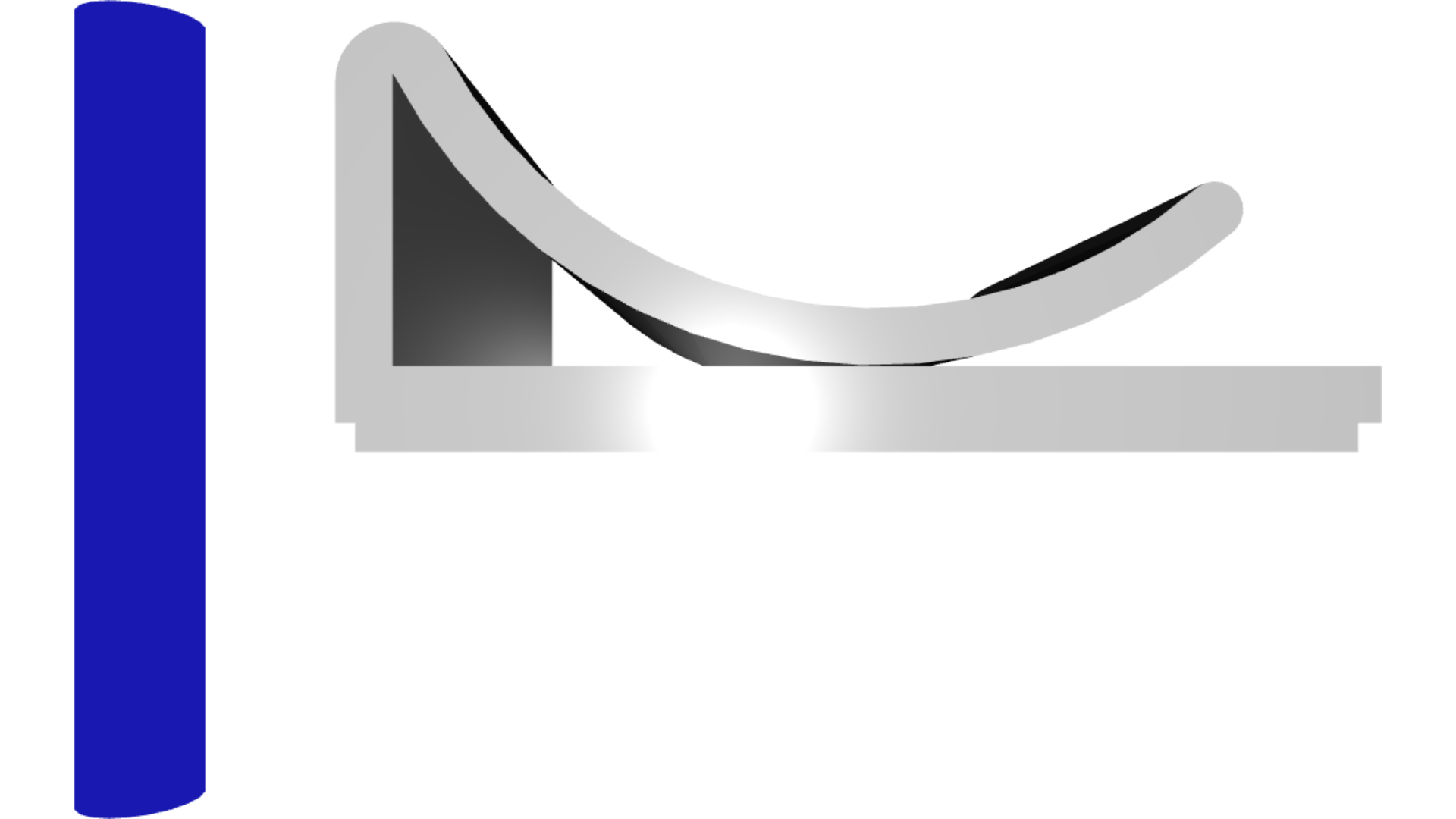}\vspace{3pt}\\\end{minipage}} & \cellcolor{singlecontact}5.8 (0.26) & 2.2 (0.10) & 1.8 (0.23) & 1.8 (0.45)\\Round Couple &\makecell{\begin{minipage}{14mm}\vspace{3pt}\centering\includegraphics[height=8mm]{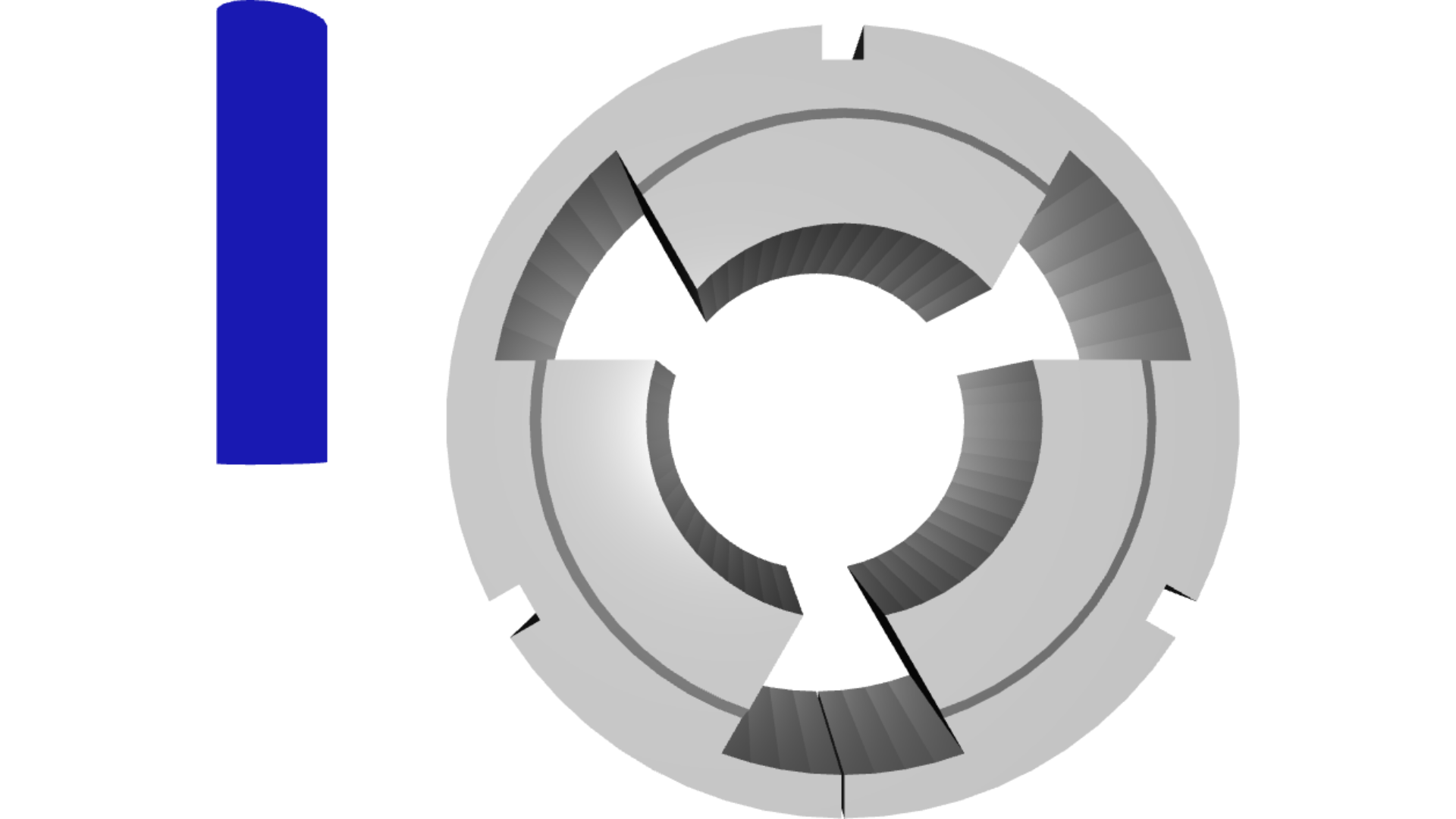}\vspace{3pt}\\\end{minipage}} & 13.6 (0.53)& \cellcolor{paralleljaw}10.9 (0.43) & 6.0 (0.75) & 3.5 (0.91)\\Hydraulic &\makecell{\begin{minipage}{14mm}\vspace{3pt}\centering\includegraphics[height=8mm]{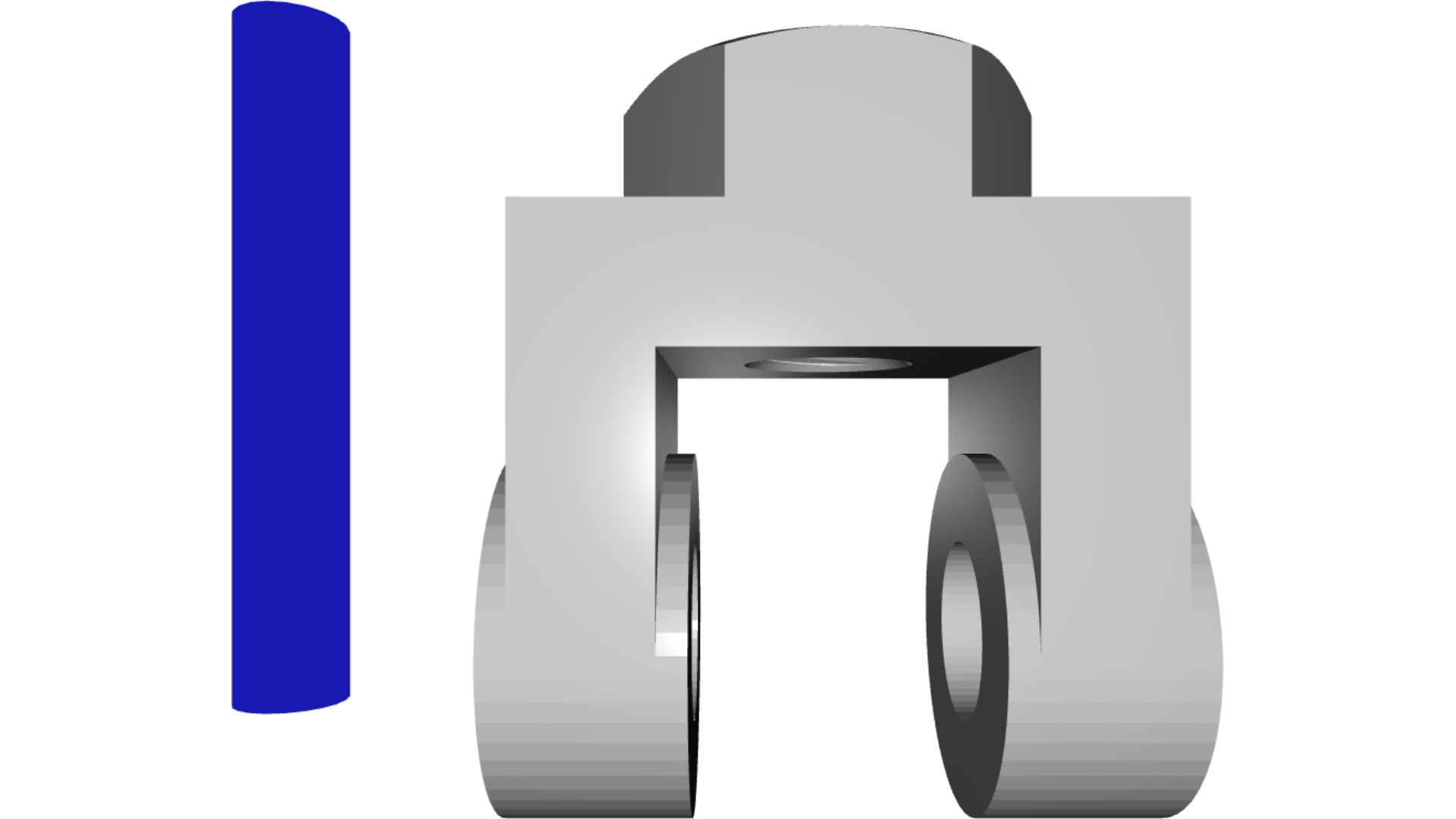}\vspace{3pt}\\\end{minipage}} & 14.0 (0.67)& \cellcolor{paralleljaw} \textbf{4.9 (0.23)}& 2.5 (0.31) & 2.0 (0.49)\\Long Grease &\makecell{\begin{minipage}{14mm}\vspace{3pt}\centering\includegraphics[height=8mm]{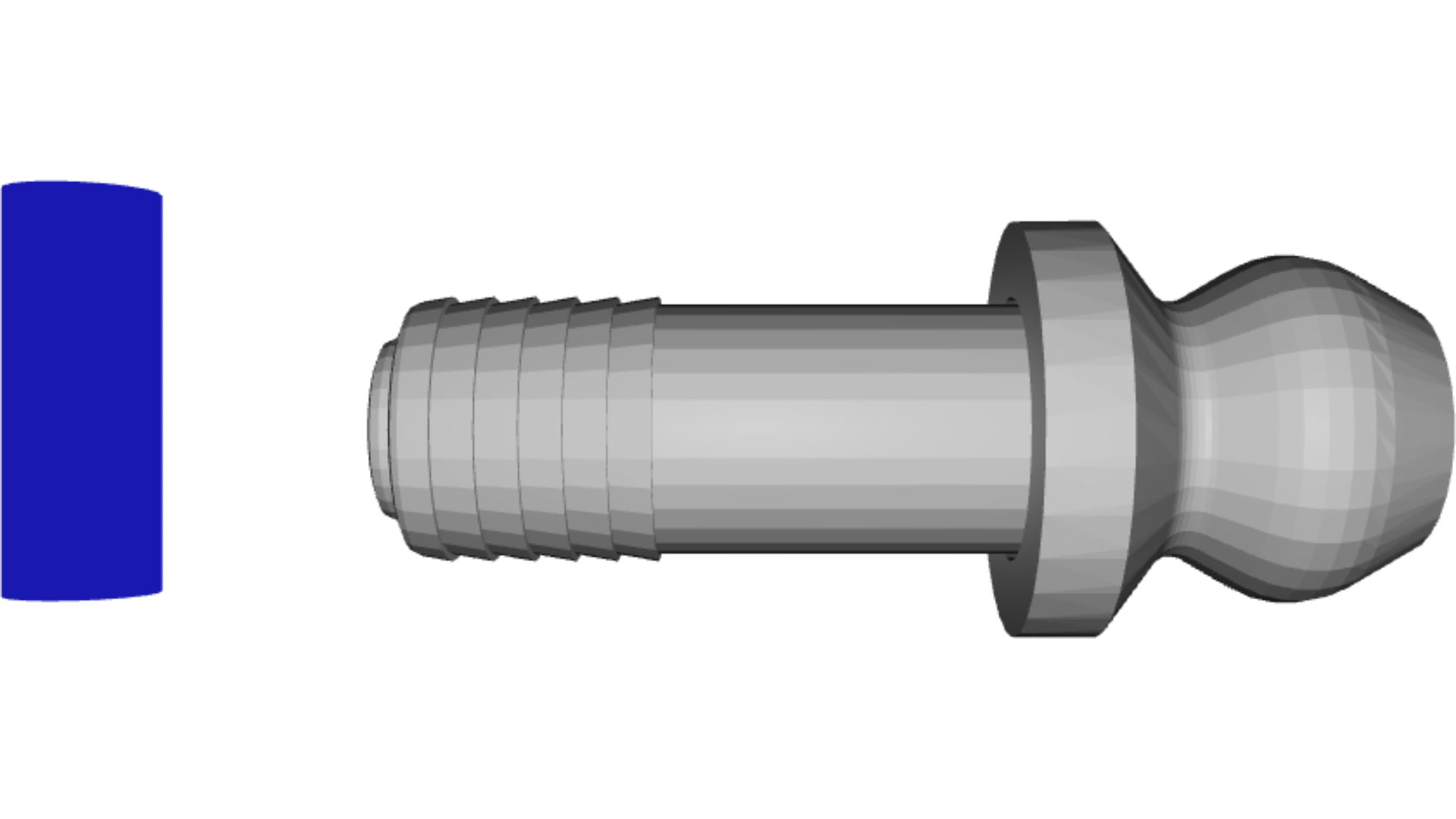}\vspace{3pt}\\\end{minipage}} & 26.6 (0.76)& \cellcolor{paralleljaw} \textbf{3.3 (0.09)}& 2.3 (0.33) & 2.3 (0.61)\\Stud &\makecell{\begin{minipage}{14mm}\vspace{3pt}\centering\includegraphics[height=8mm]{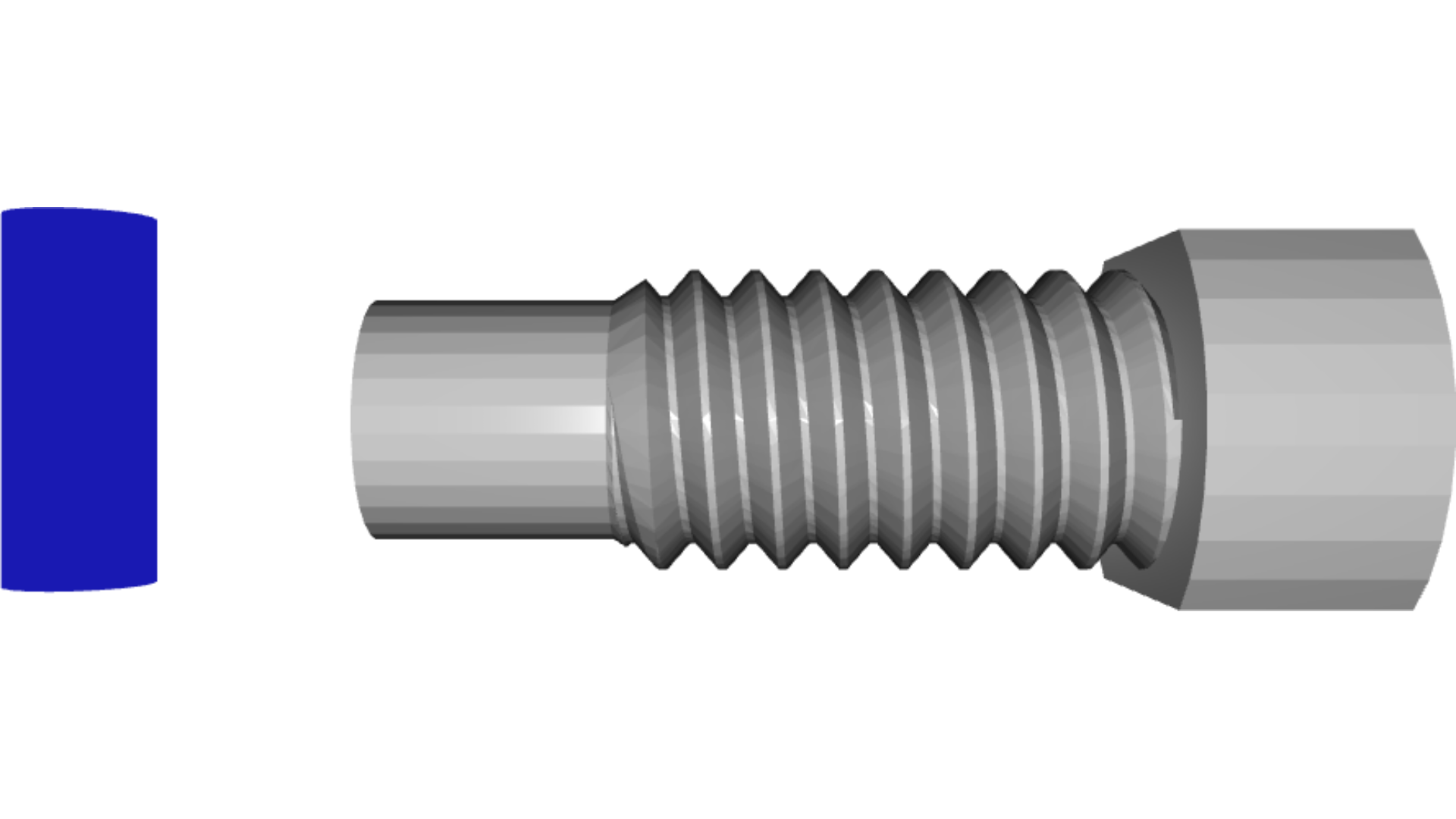}\vspace{3pt}\\\end{minipage}} & 33.7 (0.85)& \cellcolor{paralleljaw}13.4 (0.34) & 4.8 (0.57) & 3.4 (0.82)\\Cotter &\makecell{\begin{minipage}{14mm}\vspace{3pt}\centering\includegraphics[height=8mm]{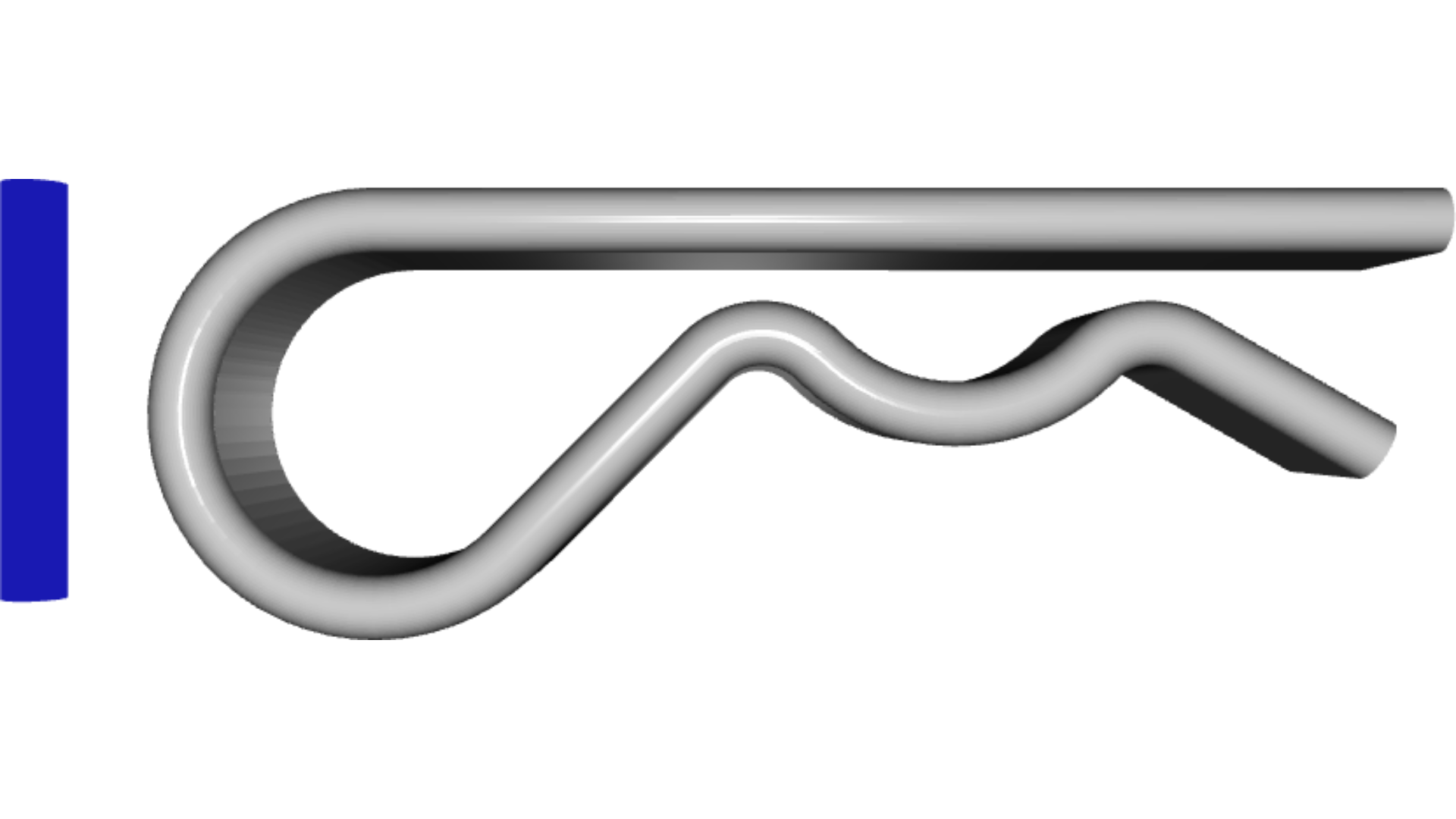}\vspace{3pt}\\\end{minipage}} & 19.0 (0.49) & 19.6 (0.51) &\cellcolor{prior} \textbf{2.9 (0.38)}& 2.7 (0.70)\\Cable Clip &\makecell{\begin{minipage}{14mm}\vspace{3pt}\centering\includegraphics[height=8mm]{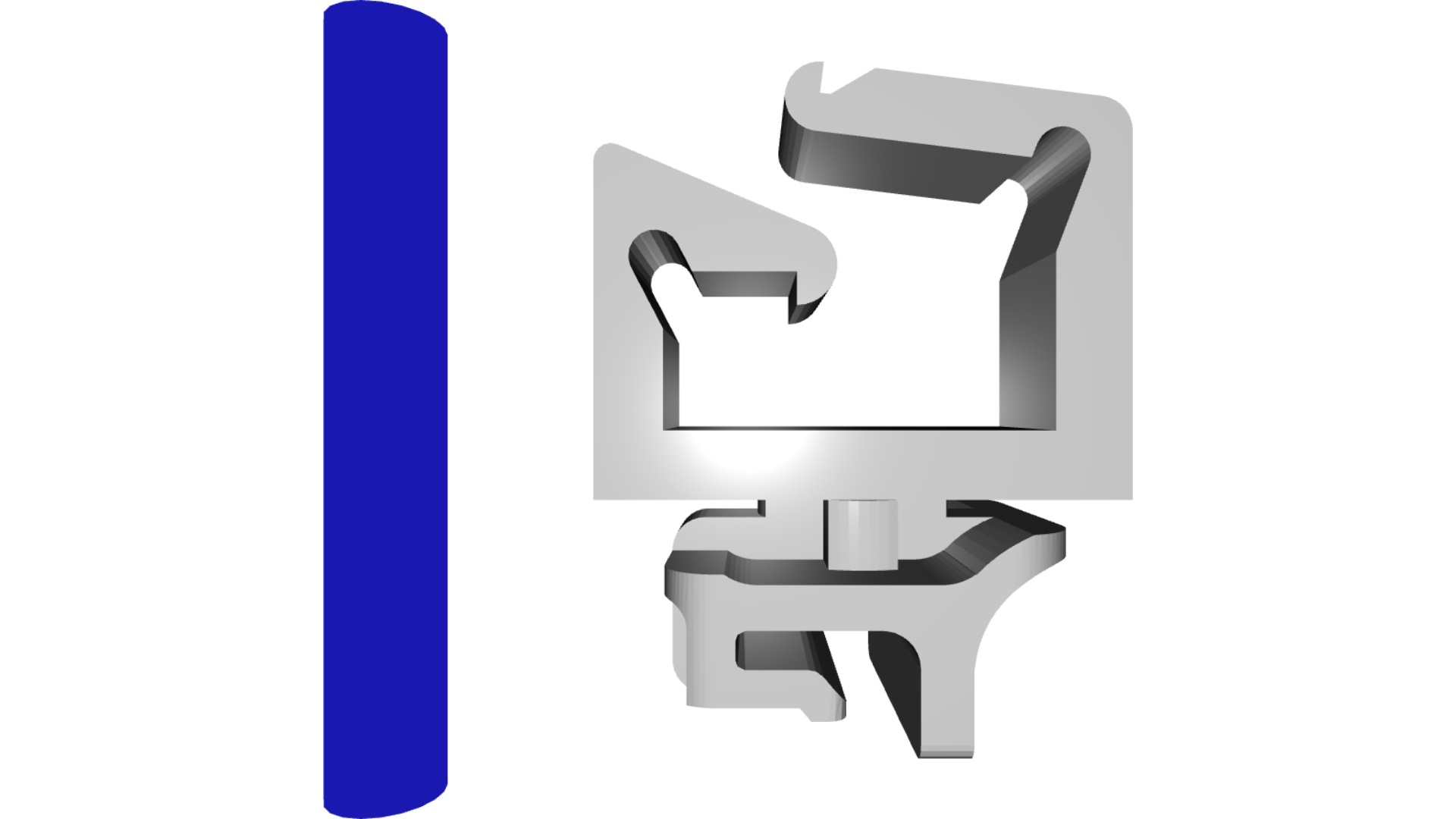}\vspace{3pt}\\\end{minipage}} & 10.2 (0.59) & 11.7 (0.67) &\cellcolor{prior}2.5 (0.30) & 1.9 (0.45)\\Hook &\makecell{\begin{minipage}{14mm}\vspace{3pt}\centering\includegraphics[height=8mm]{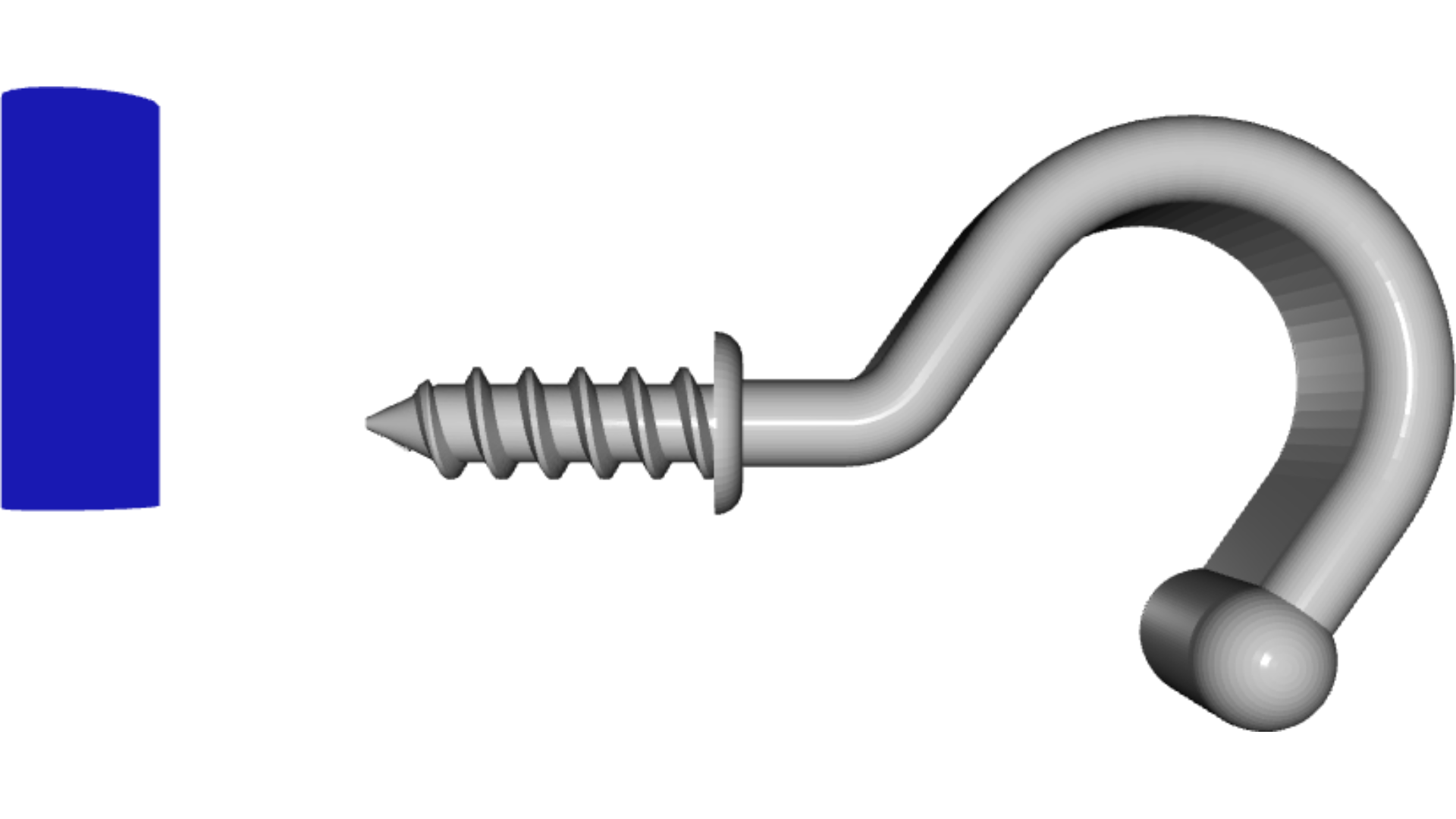}\vspace{3pt}\\\end{minipage}} & 24.5 (0.77) & 27.2 (0.85) &\cellcolor{prior}3.0 (0.38) & 2.4 (0.63)\\Couple &\makecell{\begin{minipage}{14mm}\vspace{3pt}\centering\includegraphics[height=8mm]{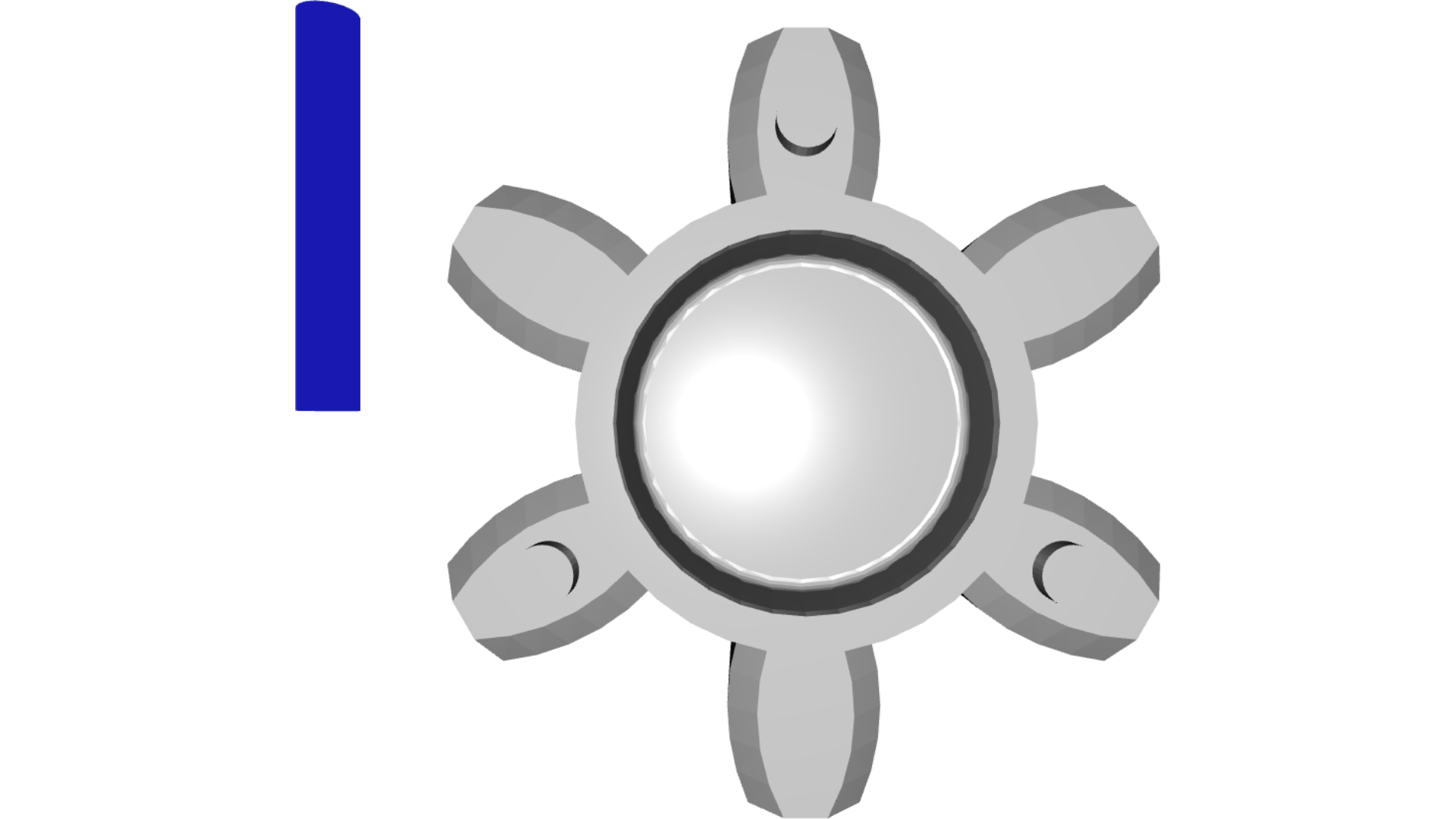}\vspace{3pt}\\\end{minipage}} & 20.7 (0.79) & 19.9 (0.76) &\cellcolor{prior}3.5 (0.42) & 2.6 (0.66)\\Hose &\makecell{\begin{minipage}{14mm}\vspace{3pt}\centering\includegraphics[height=8mm]{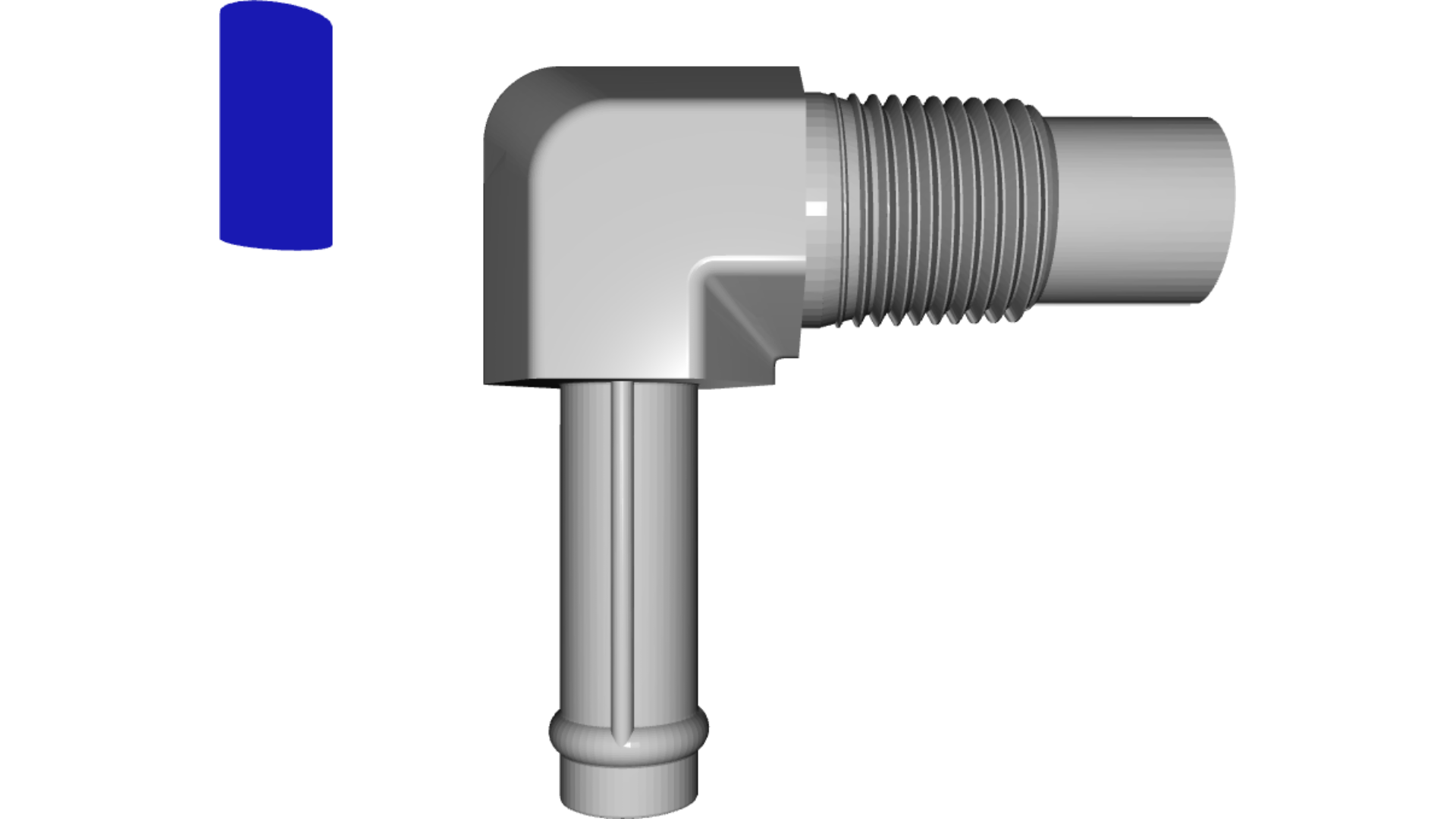}\vspace{3pt}\\\end{minipage}} & 39.0 (0.62) & 41.6 (0.66) & 7.8 (1.00) & 4.2 (1.03)\\Pencil &\makecell{\begin{minipage}{14mm}\vspace{3pt}\centering\includegraphics[height=8mm]{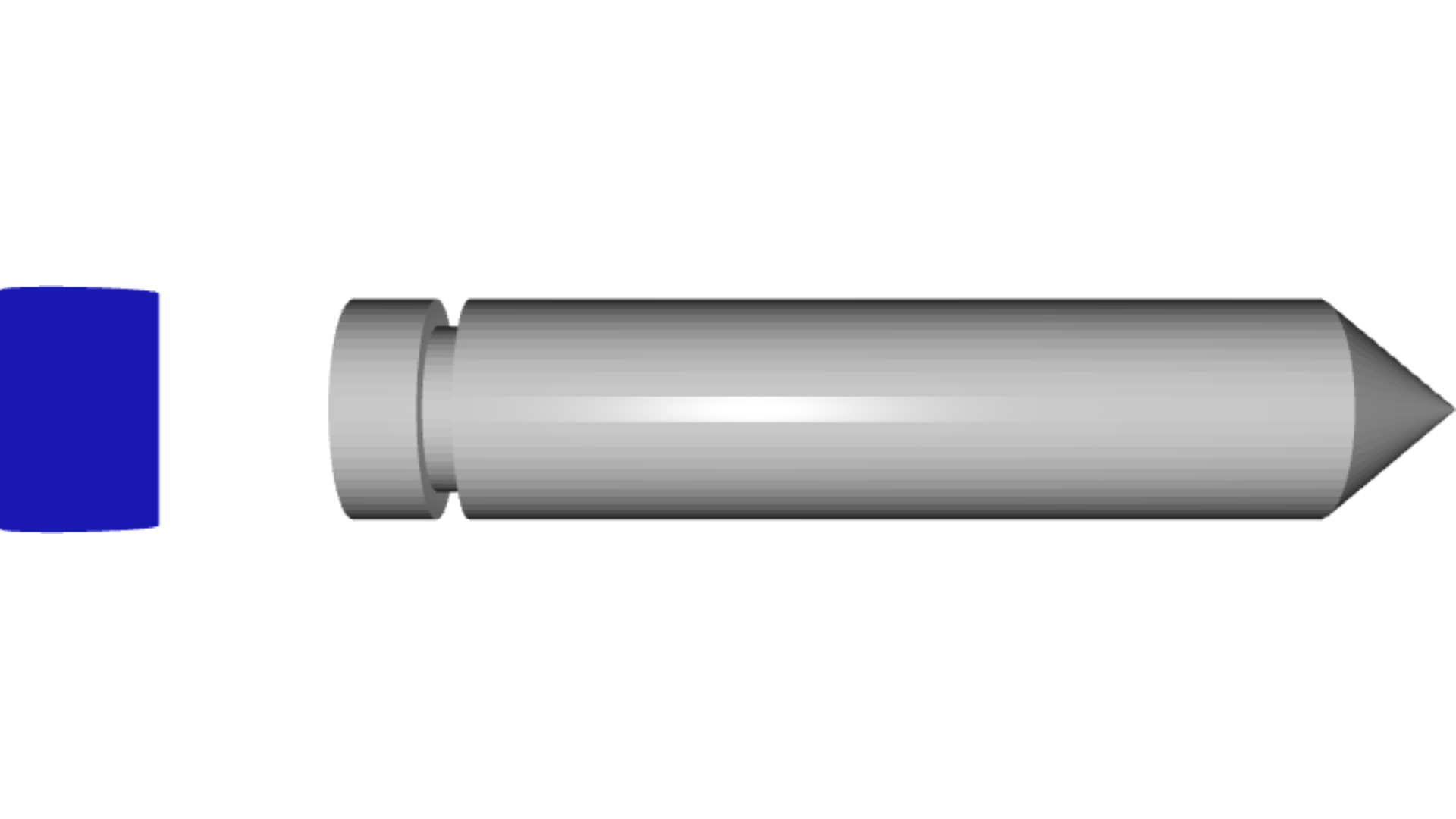}\vspace{3pt}\\\end{minipage}} & 38.1 (0.69) & 41.6 (0.76) & 5.0 (0.62) & 3.9 (1.05)\\Round Hose &\makecell{\begin{minipage}{14mm}\vspace{3pt}\centering\includegraphics[height=8mm]{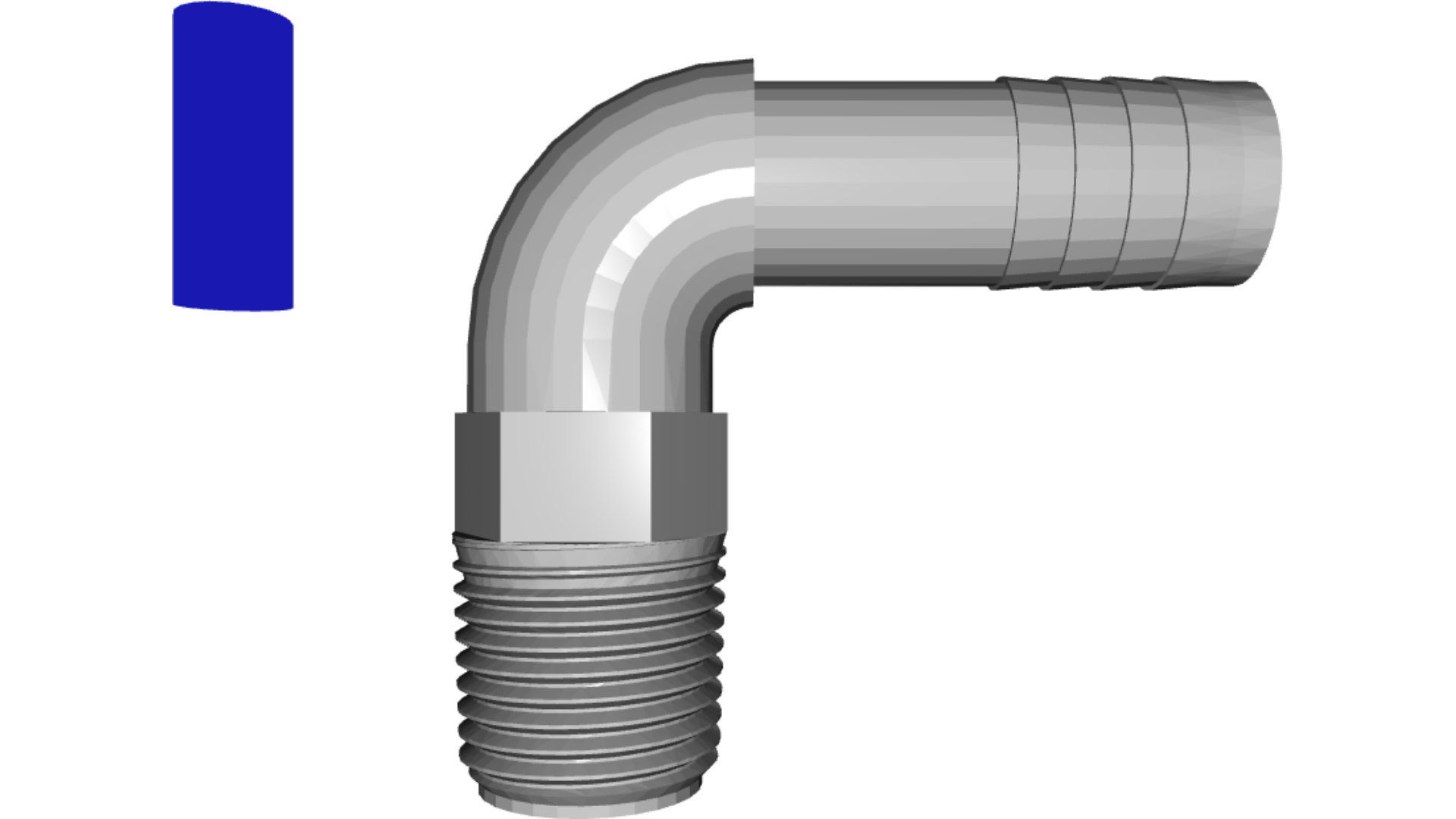}\vspace{3pt}\\\end{minipage}} & 37.6 (0.70) & 37.3 (0.69) & 5.5 (0.71) & 4.2 (1.05)\\Long Pencil &\makecell{\begin{minipage}{14mm}\vspace{3pt}\centering\includegraphics[height=8mm]{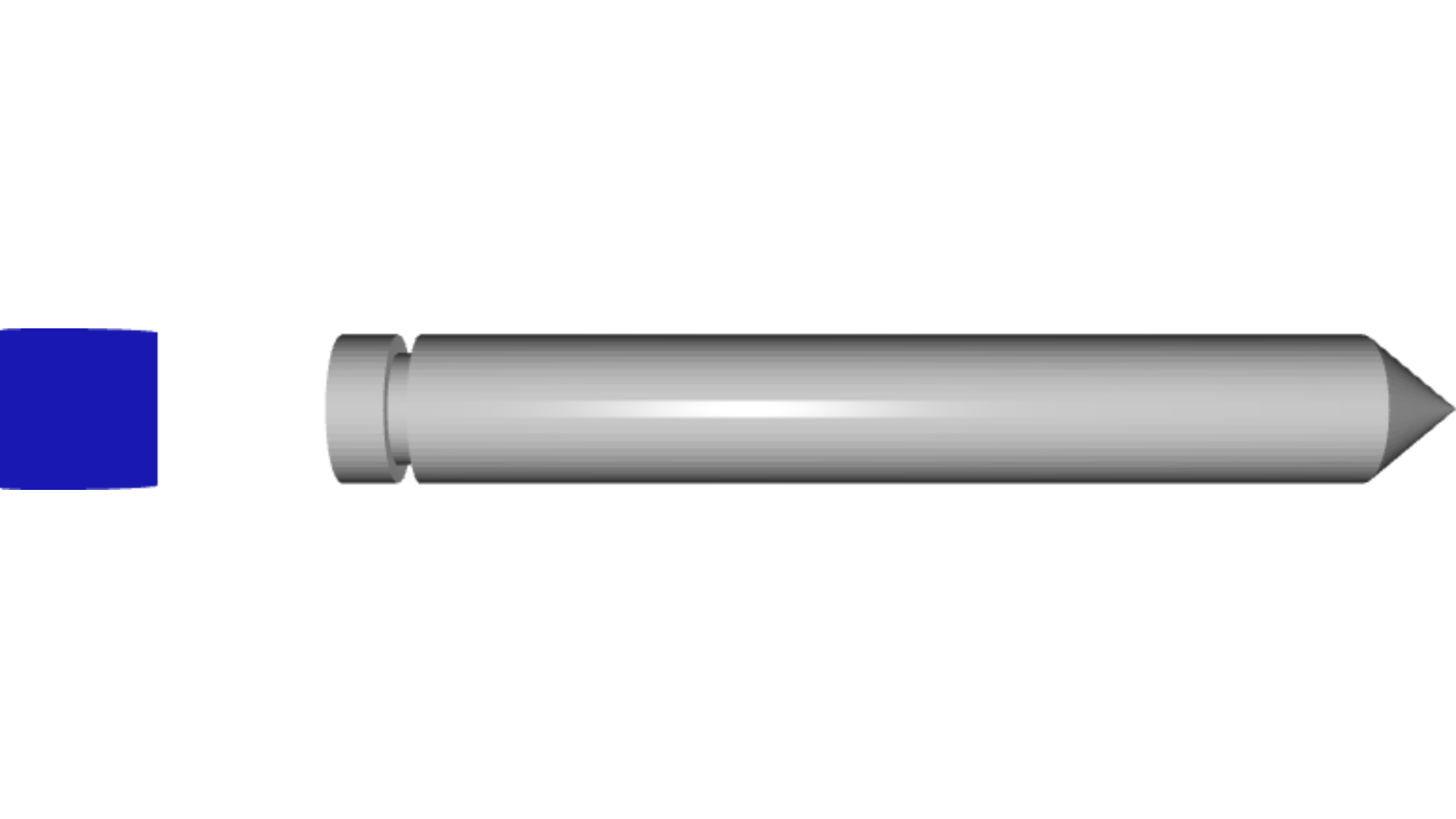}\vspace{3pt}\\\end{minipage}} & 77.5 (0.96) & 78.5 (0.97) & 6.9 (0.85) & 4.3 (1.01)\\\hline\end{tabular}\end{center}\end{table*}

\begin{figure*}[h]
    \centering
    \includegraphics[width=0.93\linewidth]{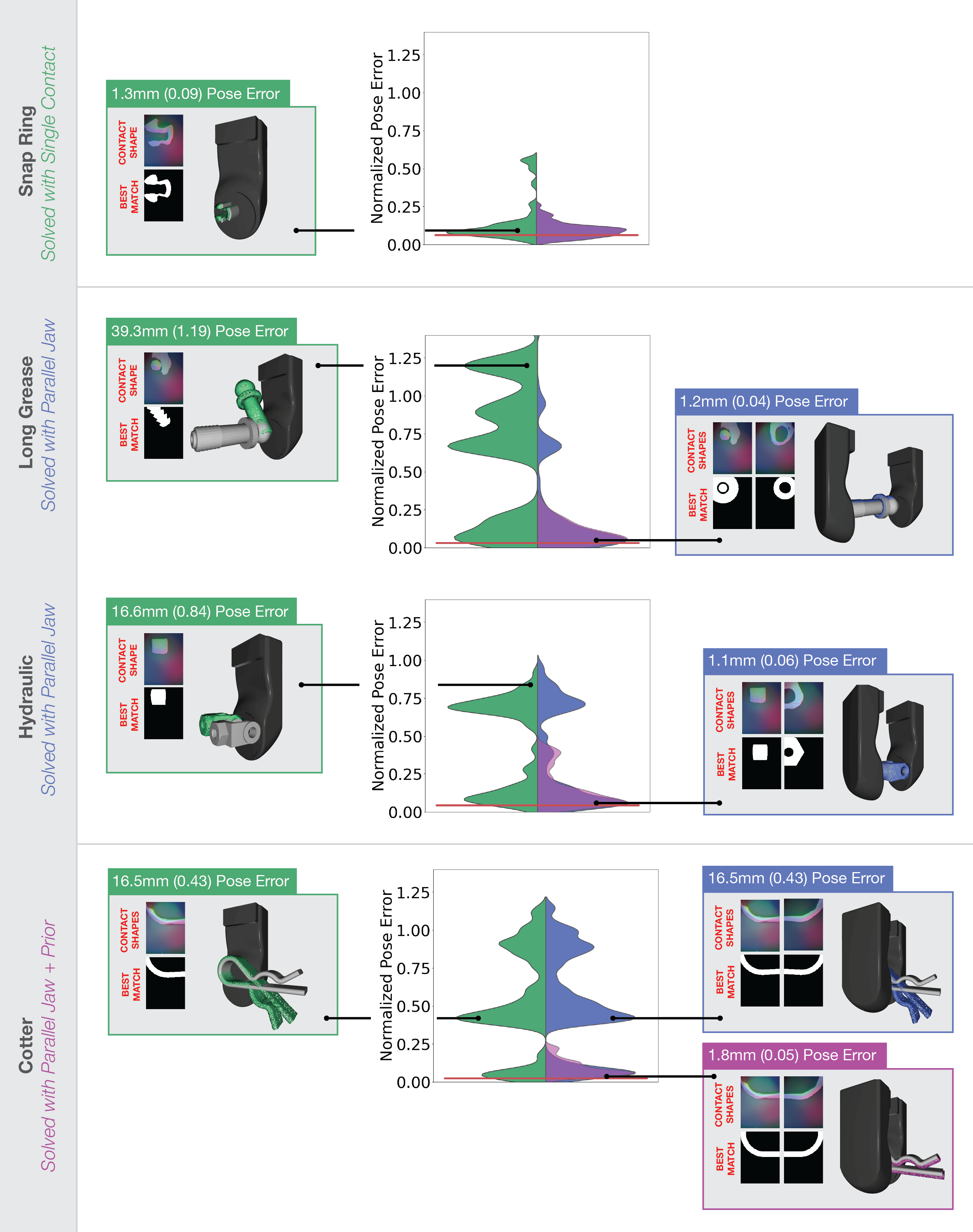}
    \caption{Error distributions for {\fontfamily{lmtt}\selectfont snap ring} (top row), {\fontfamily{lmtt}\selectfont long grease} (second row), {\fontfamily{lmtt}\selectfont hydraulic} (third row), and {\fontfamily{lmtt}\selectfont cotter} (bottom row). Each violin plot contains three distributions. At left, the single contact error distribution is visualized in \textcolor{Green}{green}. At right, the parallel jaw contact distribution is visualized in \textcolor{blue}{blue}, while the parallel jaw distribution filtered with a 10mm prior is overlaid in transparent \textcolor{violet}{purple}. The median closest grid error is visualized as a \textcolor{red}{red} line. This corresponds to the best possible median performance. The median random error corresponds to a normalized pose error of 1. 
    To the left and right of each violin plot, we show the localization errors for a sample grasp on each object. In the \textcolor{Green}{green} boxes, the true object pose (grey mesh) is visualized with the best match when using a single contact (\textcolor{Green}{green} pointcloud). The \textcolor{blue}{blue} boxes and \textcolor{violet}{purple} box show the same true object pose (grey mesh) with the best match when using parallel jaw contacts (\textcolor{blue}{blue} pointcloud), and parallel jaw contacts + a 10mm pose prior (\textcolor{violet}{purple} pointcloud), respectively.
    A single contact is enough to, on average, localize {\fontfamily{lmtt}\selectfont snap ring} accurately using \methodd (top row). 
    Parallel jaw contacts are enough to localize {\fontfamily{lmtt}\selectfont long grease} and {\fontfamily{lmtt}\selectfont hydraulic} accurately (second and third rows), even though a single contact is not. 
    Parallel jaw contacts and a coarse (10mm) pose prior are enough to localize {\fontfamily{lmtt}\selectfont cotter} accurately (bottom row), even though single contact and parallel jaw contact information are not enough.}
    \label{fig:main_figure}
\end{figure*}

We also aggregate the pose errors corresponding to the best estimate for each grasp, then visualize the distribution of errors as a violin plot in Figure \ref{fig:main_figure}, for selected objects. The medians reported in Table \ref{table:main_all_results} correspond to the medians of the distributions visualized in the violin plots.

For each of the selected objects, we show three different error distributions, corresponding to the best estimates when using: a single contact (\textcolor{Green}{green} distribution), parallel jaw contacts (\textcolor{blue}{blue} distribution), and parallel jaw contacts plus a 10mm pose prior (\textcolor{violet}{purple} distribution).
%
%
To facilitate comparison between the objects, we plot normalized pose errors. Note that a normalized pose error of 1 corresponds to the expected error from selecting a pose at random from the grid.
The \textcolor{red}{red} line measures the median pose error between the ground truth pose and its closest pose in the object's grid of simulated contacts (a.k.a. the \textit{closest error}). 
This sets a lower bound on the median performance for any given method.~\looseness=-1

\myparagraph{Single Contact.} We first analyze the performance of \methodd with a single contact. 
%
Of the 20 objects we evaluate, 9 have median localization error with a single contact that is at least twice as good as random. These 9 objects have single contact normalized errors less than 0.5 in Table \ref{table:main_all_results}.

The objects that tend to perform best with a single contact are small objects with unique tactile features. 
For these objects, a single tactile imprint is often discriminative enough to fully determine the object pose. 
Consider the error distribution for {\fontfamily{lmtt}\selectfont snap ring}, visualized in top row of Figure \ref{fig:main_figure} as an example. 
Even with a single contact (\textcolor{Green}{green} distribution), the error distribution has a clear primary mode which is centered near the closest error from the grid (\textcolor{red}{red} line). 
When incorporating additional constraints from a parallel jaw grasp, the distribution (\textcolor{blue}{blue}) tightens around the closest error from the grid (\textcolor{red}{red} line), and secondary, higher error modes disappear. 
The median normalized error is approximately 0.10 for both the single contact and parallel jaw contact cases (Table \ref{table:main_all_results}). In other words, choosing the most likely pose using \methodd results in, on average, approximately ten times less error than selecting a pose at random from the grid.

A sample contact that is localized  within 1.3mm (0.09 normalized error) of the true pose using a single contact is visualized at left of the violin plot in Figure~\ref{fig:main_figure}. This provides a sense of the scale of the errors.

\myparagraph{Parallel Jaw.} We next consider objects that perform significantly better with parallel jaw contacts. 
Of the 20 objects we evaluate, 11 have a median localization error with parallel jaw contacts that is more than twice as good as random. Of these 11 objects, 5 perform much better with parallel jaw contacts than with a single contact. The 5 objects satisfying both criteria are those in Table \ref{table:main_all_results} where the parallel jaw normalized error is less than 0.5 and the single contact normalized error is more than double the parallel jaw contact normalized error. 
Two such objects are {\fontfamily{lmtt}\selectfont long grease} and {\fontfamily{lmtt}\selectfont hydraulic}.~\looseness=-1

The objects that perform significantly better with parallel jaw contacts compared with a single contact tend to be larger. Contacts with the object may appear as large, featureless patches from the local perspective of the tactile sensor. In general, large, flat contacts are more challenging to reconstruct from tactile images because they do not deform the gel membrane of the tactile sensor as much. Including the constraints of a second contact and the gripper opening provides higher robustness to noisy contact reconstructions. 
Furthermore, objects that vary significantly in width depending on where they are grasped benefit from parallel jaw contacts. In these cases, the additional constraint of the gripper opening during grasp can disambiguate between otherwise non-unique contacts. Recall that a contact is non-unique if its look similar to contacts where the object is in a different pose. In these cases, the contact is not a unique indicator of the object pose. 

As an example, the error distribution for {\fontfamily{lmtt}\selectfont long grease} is visualized as a violin plot in the second row of Figure~\ref{fig:main_figure}. 
Considering parallel jaw information shifts the mode of the distribution much closer to the closest grid match (\textcolor{red}{red} line), and significantly reduces the prevalence of higher error modes. 
%
%
The medians of the error distributions for the single contact and parallel cases are significantly different; with a single contact, the median pose error is 26.6mm (0.76 normalized error) while with parallel jaw contacts, the median pose error is 3.3mm (0.10 normalized error). The random error, in comparison, is 35.2mm. With a single contact, \methodd only performs about 1.3 times better than random. When including parallel jaw contacts, however, \methodd is more 10 times better than random.
The second row of Figure~\ref{fig:main_figure} (left of the violin plot) shows a sample contact on {\fontfamily{lmtt}\selectfont long grease}, and the corresponding best match when using only a single contact (\textcolor{Green}{green} pointcloud). The best match using a single contact is 39.3mm (1.19 normalized error) away from the true pose. The same contact, and the corresponding best match when using parallel jaw contacts (\textcolor{blue}{blue} pointcloud) is visualized to the right of the violin plot. The best match using parallel jaw contacts is only 1.2mm (0.04 normalized error) away from the true pose. In this case, the additional information from the second contact and gripper opening is enough to resolve the ambiguity and substantially improve the localization.

Parallel jaw contacts can also be an important tool for disambiguating ambiguity inherent to the object geometry.
As an example, consider a grasp on the object {\fontfamily{lmtt}\selectfont hydraulic}, visualized at left of the violin plot in the third row of Figure~\ref{fig:main_figure}. The true object pose is visualized as a grey mesh, whereas the best match using a single contact is visualized as a \textcolor{Green}{green} pointcloud. The localization error with a single contact, in this case, is 16.6mm (0.84 normalized error). When considering only contacts from one of the fingers, the contact shapes corresponding to the true object pose and the best match are indistinguishable. Considering parallel jaw contacts, on the other hand, resolves the ambiguity.~\looseness=-1

The same contact, and the corresponding best match using parallel jaw contacts, is visualized at the right of the violin plot in the third row of the same figure. The localization (\textcolor{blue}{blue} pointcloud) using parallel jaw contacts is much better because the contact on the second finger provides critical information about the object's pose. The localization improves from 16.6mm (0.84 normalized) error with a single contact, to 1.1mm (0.06 normalized) error with parallel jaw contacts.

All grasps on {\fontfamily{lmtt}\selectfont hydraulic} in contact with this non-unique feature are subject to the same problem when considering information from only a single contact, and therefore the non-uniqueness impacts the overall localization quality; the median localization error for {\fontfamily{lmtt}\selectfont hydraulic} with a single contact is 14mm (0.67 normalized error), while with parallel jaw contact is 4.9mm (0.23 normalized error). The violin plot in the third row of Figure~\ref{fig:main_figure} shows that the distribution of errors when using a single contact (\textcolor{Green}{green}) is bimodal, with a portion of the second, high error, mode corresponding to flipped localizations, like the one shown at left of the violin plot. When considering the distribution of best match errors using parallel jaw contacts (\textcolor{blue}{blue} distribution), we see that the density of the high error mode corresponding to the flipped localization decreases, and more probability mass shifts into the low error mode centered around the closest error.

In the case of {\fontfamily{lmtt}\selectfont hydraulic}, it is informative to break out the results by \textit{grasp approach direction}. The primary reason that the localization error with parallel jaw contacts is, on average, nearly 3 times better than with a single contact is that including parallel jaw contacts resolves the aforementioned non-uniqueness, which only impacts one of the two grasp approach directions. For the grasp approach direction that contains the non-unique feature, including parallel jaw contacts reduces the localization error by nearly 8 times. For the other grasp approach direction, including parallel jaw contacts reduces the localization error by 1.6 times.~\looseness=-1

%

Only one object, {\fontfamily{lmtt}\selectfont round clip}, has significantly higher median error in the parallel jaw case (see Appendix for details) compared with the single contact case.

\myparagraph{Parallel Jaw + Prior.} Finally, we consider the performance of \methodd with parallel jaw contacts plus a prior on the object pose. Pose priors, in practice, can be obtained using additional sensing modalities (e.g. vision), kinematics, or previous estimates of the object's pose.
Because tactile sensing is inherently local, an object can have non-unique contacts which won't be possible to fully disambiguate even with parallel jaw contacts. In such cases, a prior on the object's pose can help resolve these ambiguities.
To test the effect of a prior on \methodd, we take the prediction distribution over possible object poses, and filter any pose that is more than a given distance from ground truth. Note that this implementation of a pose prior requires knowledge of the ground truth object pose, but it is a useful proxy for cases in which the object pose is known roughly, but not accurately.
We evaluate localization accuracy for prior distances of 10mm and 5mm. Results for both prior distances for each of the 20 objects are listed in Table \ref{table:main_all_results}.

The objects that benefit most from incorporating a prior on the object pose are those with \textit{discrete non-uniqueness}. An object with discrete non-uniqueness has features that are unique (i.e. create discriminate tactile imprints) relative to most other possible contacts on the object, with a discrete number of exceptions. This type of discrete non-uniqueness can be an inherent feature of the object, or an artifact of noisy and incomplete contact shapes.
Because there is a discrete number of possible object poses that are likely to produce such a contact shape, we expect the distribution over object poses to have a discrete number of modes where the majority of probability mass is concentrated.
In these cases, incorporating a pose prior can truncate the distribution in such a way that only one of the modes is left.  In other words, because there is a discrete number of possible poses that are likely to produce a given contact, if we eliminate the higher error options using a pose prior, we are likely to get a near perfect match. Therefore, for some objects, even a coarse prior on the object pose results in precise localization when combined with tactile information.

We evaluate which objects benefit most from incorporating a pose prior by comparing the pose error of the best match after filtering the pose distribution, with the expected error from selecting a pose at random from the filtered distribution. The normalized error we report in Table \ref{table:main_all_results} for the 10mm and 5mm prior ablations uses the expected random error from the filtered distribution. 
For the remainder of this section, we consider just the case of a 10mm pose prior. For 12 of the 20 objects we evaluate, using \methodd on top of a 10mm pose prior results in performance more than twice as good as selecting a pose at random from the filtered distribution. For 5 of the 12 objects, incorporating a pose prior plays an important role in driving down the localization error. With just parallel jaw contact information, the median localization error for these 5 objects is not significantly better than selecting a pose at random from the grid (where here we take \textit{significantly better} to be more than twice as good as random). However, after filtering the distribution with a 10mm prior on the object pose, \methodd selects the best pose with more than twice as much accuracy as selecting a pose at random from the filtered distribution. The objects that benefit most from incorporation of a pose prior are those in Table \ref{table:main_all_results} with parallel jaw normalized error greater than 0.5, but 10mm prior normalized errors less than 0.5. An example of an object for which the incorporation of a 10mm prior leads to significantly better localization is {\fontfamily{lmtt}\selectfont cotter}.

By examining the multi-modal error distributions with a single contact, or parallel jaw contact (violin plot in the bottom section of Figure \ref{fig:main_figure}), we can notice that the object {\fontfamily{lmtt}\selectfont cotter} is likely to benefit from using pose prior. Its error distributions have a discrete number of modes (three) where the majority of probability mass is concentrated. 
Note that the symmetry of {\fontfamily{lmtt}\selectfont cotter} is such that parallel jaw contacts do not provide much new information relative to a single contact, so the distributions of error for a single contact and parallel jaw contacts are very similar. The median error is 19.0mm (0.49 normalized error) when using a single contact, and 19.6mm (0.51 normalized error) when using parallel jaw contacts.
Incorporation of a 10mm prior on the object pose eliminates the two higher error modes, and shifts the median of the distribution toward the closest error. The median error becomes 2.9mm (0.38 normalized error), which is nearly seven times lower than using parallel jaw contact information alone. Recall that after incorporating a pose prior, the median error is normalized by the expected error from selecting a pose at random from the filtered distribution rather than from all contact poses.

We show an example of a grasp that benefits from a pose prior at left of the violin plots in the bottom row of Figure \ref{fig:main_figure}. With single or parallel jaw contacts, the localization error is 16.5mm (0.43 normalized). The contacts when {\fontfamily{lmtt}\selectfont cotter} is in the given pose (grey mesh) are informative, but not completely unique relative to other possible contacts on the object. Incorporation of a coarse 10mm prior on the object pose, though, resolves the ambiguity, and the localization improves from 16.5mm (0.43 normalized) to 1.8mm (0.23 normalized) error for this grasp. The best match after filtering the parallel jaw distribution is visualized as a \textcolor{violet}{purple} pointcloud on the right side of the violin plot in the bottom row of Figure~\ref{fig:main_figure}.

\begin{figure}[!ht]
    \centering
    \includegraphics[width=\linewidth]{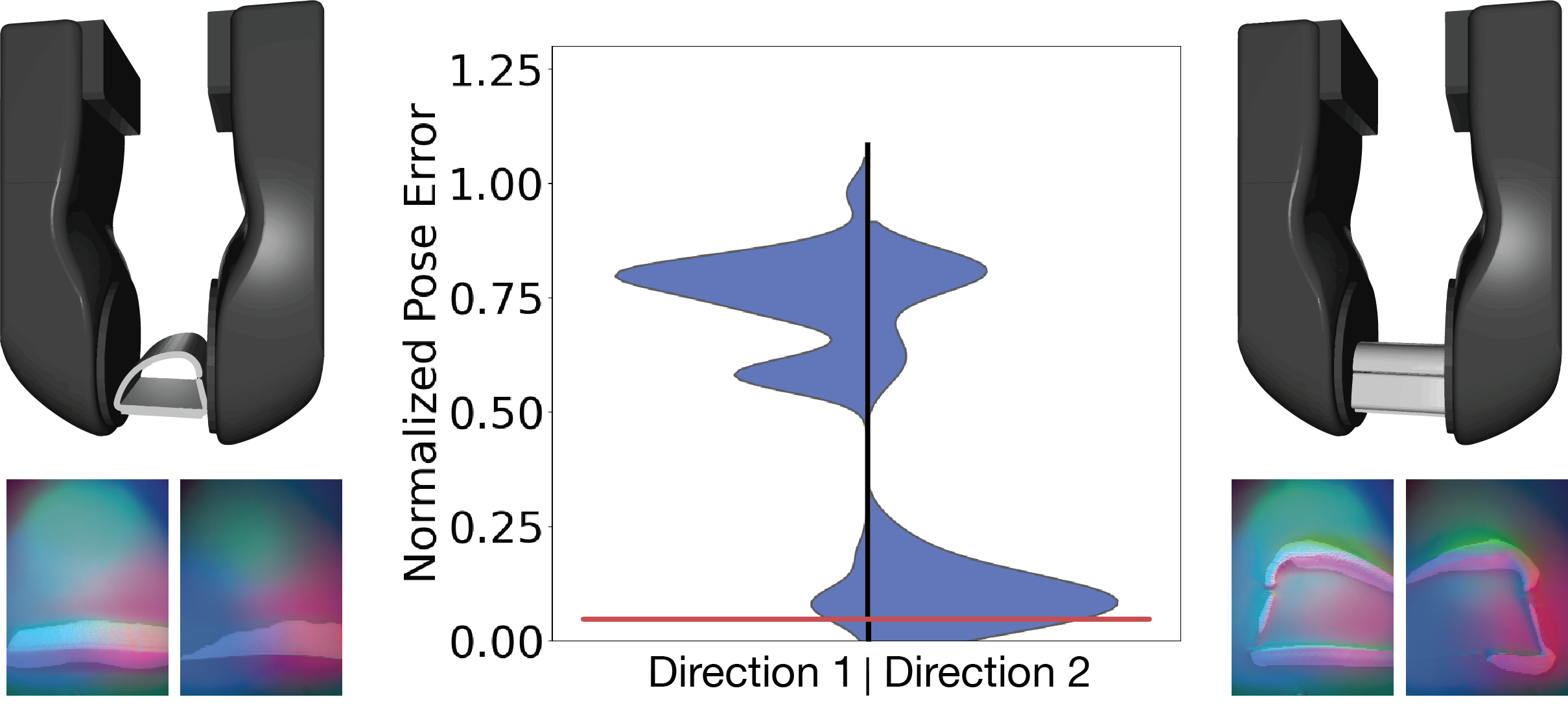}
    \caption{Error distributions for the first (left) and second (right) grasp approach directions on {\fontfamily{lmtt}\selectfont \textbf{round clip}} using parallel jaw contacts. The \textcolor{red}{red} line represents the median closest grid error. The first grasp approach direction has more non-unique contacts (sample contacts visualized at left of the plot), and therefore has higher median error. Contacts in the second grasp approach direction, on the other hand, are much more unique (sample contacts visualized at right of the plot).}
    \label{fig:cable_g_figure}
\end{figure}

\myparagraph{Comparing Grasp Approach Directions.} \label{sec: grasp_appoach} We evaluate multiple grasp approach directions for 13 of the 20 objects. For these objects, we compare the median localization error using parallel jaw contacts of each grasp approach directions. In practice, knowing which sets of grasps lead to lower localization errors could be leveraged in a grasp planning framework to select more informative grasps. There are 6 objects for which one direction has half as much error or less as the other(s): {\fontfamily{lmtt}\selectfont long grease}, {\fontfamily{lmtt}\selectfont grease}, {\fontfamily{lmtt}\selectfont hanger}, {\fontfamily{lmtt}\selectfont hydraulic}, {\fontfamily{lmtt}\selectfont round clip} and {\fontfamily{lmtt}\selectfont round couple}. Full results are shown in Table~\ref{table:grasp_approach_directions} in Appendix Section~\ref{sec:direction_appendix}.
As example, we compare the two grasp approach directions for {\fontfamily{lmtt}\selectfont round clip} in Figure~\ref{fig:cable_g_figure}. The first grasp approach direction, visualized in the left half of the figure, has non-unique contacts. As a result, the median normalized error is 15.2mm (0.85 normalized) when using parallel jaw contacts. The second grasp approach direction, visualized in the right half of the figure, has much more unique contacts; the median normalized error is 2.2mm (0.11 normalized), and the primary mode of the error distribution is concentrated near the closest grid error. The second grasp approach direction consists of grasps that, on average, lead to about eight times lower localization errors.

\myparagraph{Comparing Individual Grasps.} Some objects are difficult to localize even in the presence of a pose prior. For these objects, \methodd does not reduce the median error much beyond the prior on the object pose. These objects are characterized by large, continuous, regions of non-unique contacts, and their error distributions tend to be broader, rather than having a discrete number of modes. 
\begin{figure*}[ht]
    \begin{subfigure}[t]{0.5\linewidth}
        \includegraphics[width=\linewidth]{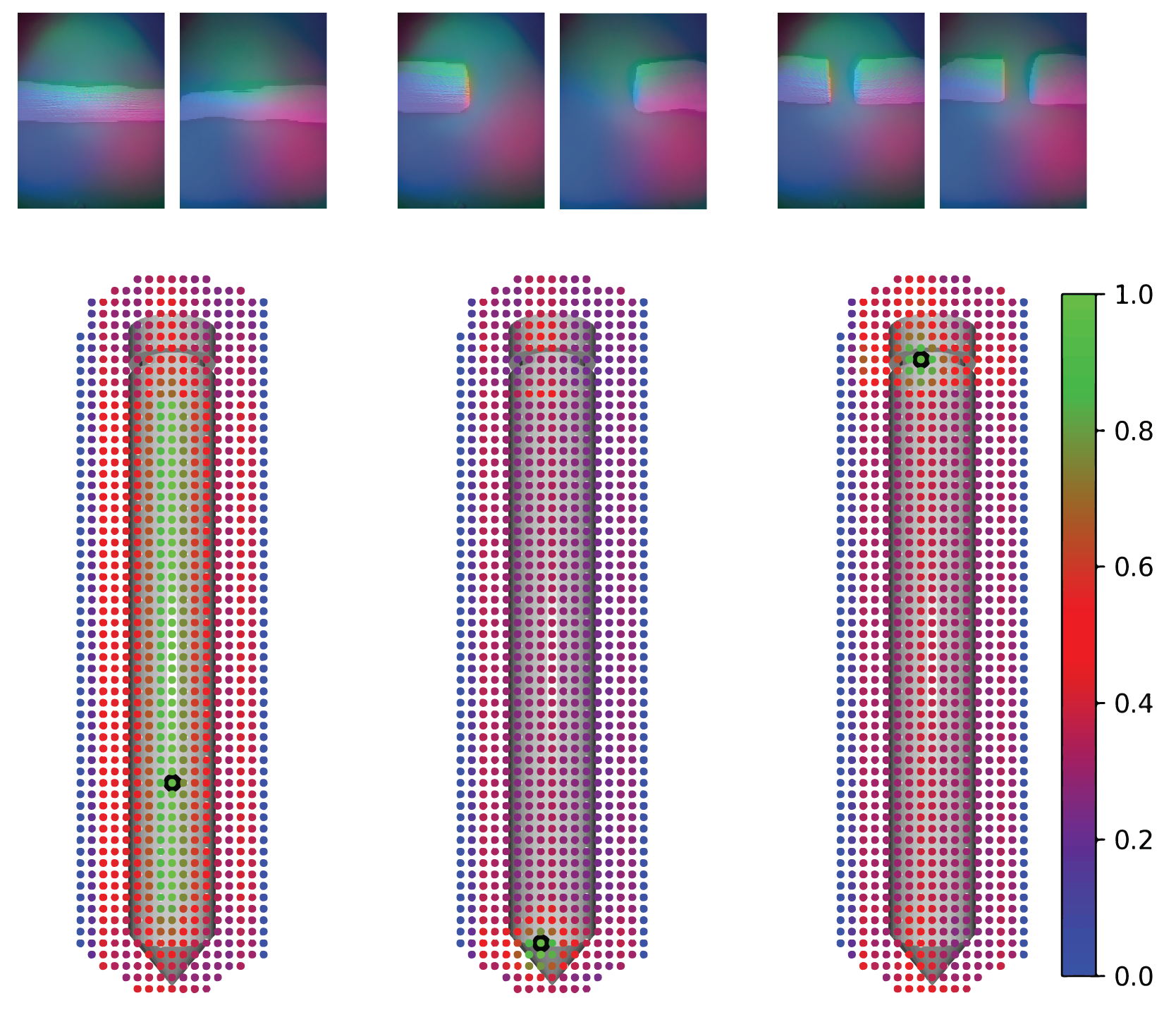}
        \caption{Real RGB tactile images from the sensors with contact masks overlaid for three grasps on {\fontfamily{lmtt}\selectfont long pencil} (top). Distributions over possible contacts on the object, using parallel jaw information from the real grasp (bottom).}
        \label{fig:long_pencil_figure}
    \end{subfigure}
    \quad
    \begin{subfigure}[t]{0.5\linewidth}
        \includegraphics[width=\linewidth]{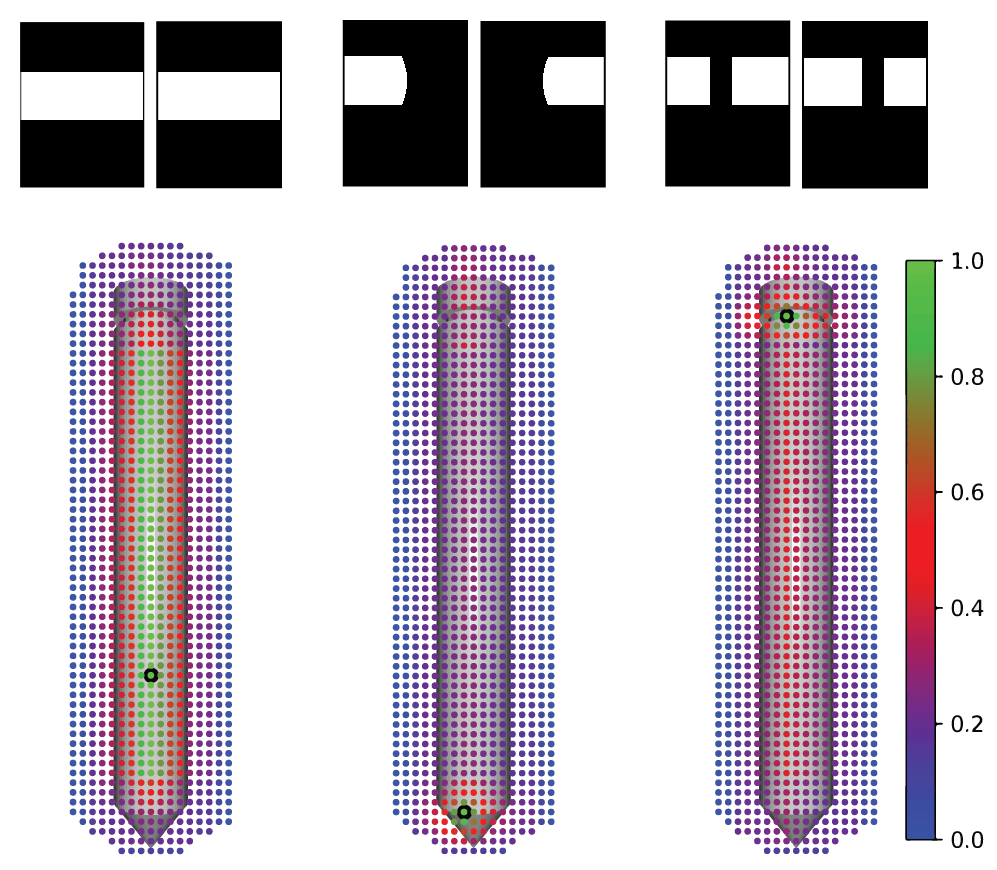}
        \caption{Simulated contact shapes for three grasps on {\fontfamily{lmtt}\selectfont long pencil} (top). Distributions over possible contacts on the object, using parallel jaw information from the simulated grasp  (bottom).}
        \label{fig:long_pencil_figure_sim}
    \end{subfigure}
    \caption{Output pose distributions for three grasps on {\fontfamily{lmtt}\selectfont long pencil}, using real (\ref{fig:long_pencil_figure}) and simulated (\ref{fig:long_pencil_figure_sim}) parallel jaw contacts. Each dot represents a possible grasp center location when the object is oriented parallel to the table, and its color represents the likelihood that a given grasp generated the tactile observation. The true grasp location is shown as a black dot. The left grasp in both sub-figures is on a non-unique region of the object; many grasps along the length of the object have high probability (\textcolor{Green}{green} dots). On the other hand, the grasps at center and right of both sub-figures are on unique regions of the object. Only grasps immediately surrounding the true grasp (black dot) have high probability (\textcolor{Green}{green} dots). The distributions from the real and simulated contact masks are qualitatively similar.}
\end{figure*}

{\fontfamily{lmtt}\selectfont Long pencil} (Figure~\ref{fig:long_pencil_figure}), for example, suffers from having large regions of non-unique contacts. Many of the contact poses along the length of the object result in the same information, and thus on average the match is chosen essentially at random from a large set of possible contacts. Note that this is not a limitation of \methodd for tactile localization, but rather results from the object geometry and the local nature of tactile feedback; most contacts on {\fontfamily{lmtt}\selectfont long pencil} are not sufficient to uniquely determine the object pose.~\looseness=-1

Even for objects with a large amount of non-uniqueness on average, there can be regions of the object which create distinctive and unique tactile imprints. Contacts near either end of {\fontfamily{lmtt}\selectfont long pencil}, for example, are easier to localize accurately. Grasping on these regions leads to better localization. Depending on the downstream application, it might be beneficial to plan for grasps on an object that are expected to produce more unique tactile imprints, and thus lower localization errors. The existence of good grasps is something we can detect and exploit to avoid large regions of non-uniqueness, and achieve low localization errors. 
The distribution over contact poses for selected contacts on unique and non-unique regions of {\fontfamily{lmtt}\selectfont long pencil} is shown in Figure \ref{fig:long_pencil_figure}. The top row of the figure shows three possible grasps on the object. The bottom row shows the output of \methodd - a distribution over possible contact poses. Each dot overlaid on {\fontfamily{lmtt}\selectfont long pencil} represents a possible grasp location, and the color represents the likelihood that a given grasp generated the input tactile observation. The \textcolor{Green}{green} dots are grasps with highest likelihood, while the \textcolor{blue}{blue} dots are grasps with lowest likelihood. The black dot represents the true grasp location. Note that for ease of visualization, we show only points that represent center locations of the grasp when {\fontfamily{lmtt}\selectfont long pencil} is oriented horizontally. This is a small subset (935) of the total number of contacts (65k) \methodd reasons over.
For contacts near the middle of the object (left of Figure \ref{fig:long_pencil_figure}, for example), many grasps along the length of the object have high probability (\textcolor{Green}{green} dots). For instance, the left contact in Figure~\ref{fig:long_pencil_figure} on its own is not enough to uniquely determine the object pose. In fact, the pose with the highest likelihood results in 77.9mm (0.99 normalized) error for this case. This aligns with our intuitive understanding that the best match is chosen essentially at random from a large set of likely contacts (\textcolor{Green}{green} dots) along the middle of the object. Cases like this highlight the importance of outputting meaningful distributions over possible contact poses; the true contact pose has high likelihood, so \methodd could be used in combination with information from other sensing modalities, kinematics, or previous tactile estimates to converge on a unique estimate of the true pose.

For contacts near the tips of the pencil (center and right of Figure~\ref{fig:long_pencil_figure}, for example), only grasps immediately surrounding the correct tip have high probability. The cases illustrated result in 6mm (0.09 normalized) for the center contact, and  7mm (0.09 normalized) for the right.

In Figure~\ref{fig:long_pencil_figure_sim}, we show distributions over contact poses for simulated versions of the same three contacts on {\fontfamily{lmtt}\selectfont long pencil}. The output distributions when using simulated contacts are qualitatively similar as when using real contacts. Simulated contact shapes can therefore be used to detect regions of non-uniqueness before ever encountering the object. This feature could be used in a grasp planning framework to avoid regions of non-unique contacts and promote accurate, unique pose estimation from a single grasp.~\looseness=-1

\subsection{ Comparison with Scanned Object Models}
\label{sec:scans}

\begin{table}[h]\caption{Pose error and normalized pose error (in parenthesis) for \methodd on manufacturer's CAD models versus scanned object models. We compare results with parallel jaw contacts + 10mm pose prior.}\label{table:scan_results}\renewcommand{\arraystretch}{1.5}\begin{center}
\resizebox{\columnwidth}{!}{
\begin{tabular}{|c| c c c c|}\bottomrule\multicolumn{1}{|c|}{} & \multicolumn{2}{c}{\makecell{\textbf{Scanned Model}\\mm (norm)}} & \multicolumn{2}{c|}{\makecell{\textbf{Manufacturer's Model}\\mm (norm)}}\\\cline{1-5}Snap Ring &\makecell{\begin{minipage}{14mm}\vspace{3pt}\centering\includegraphics[height=8mm]{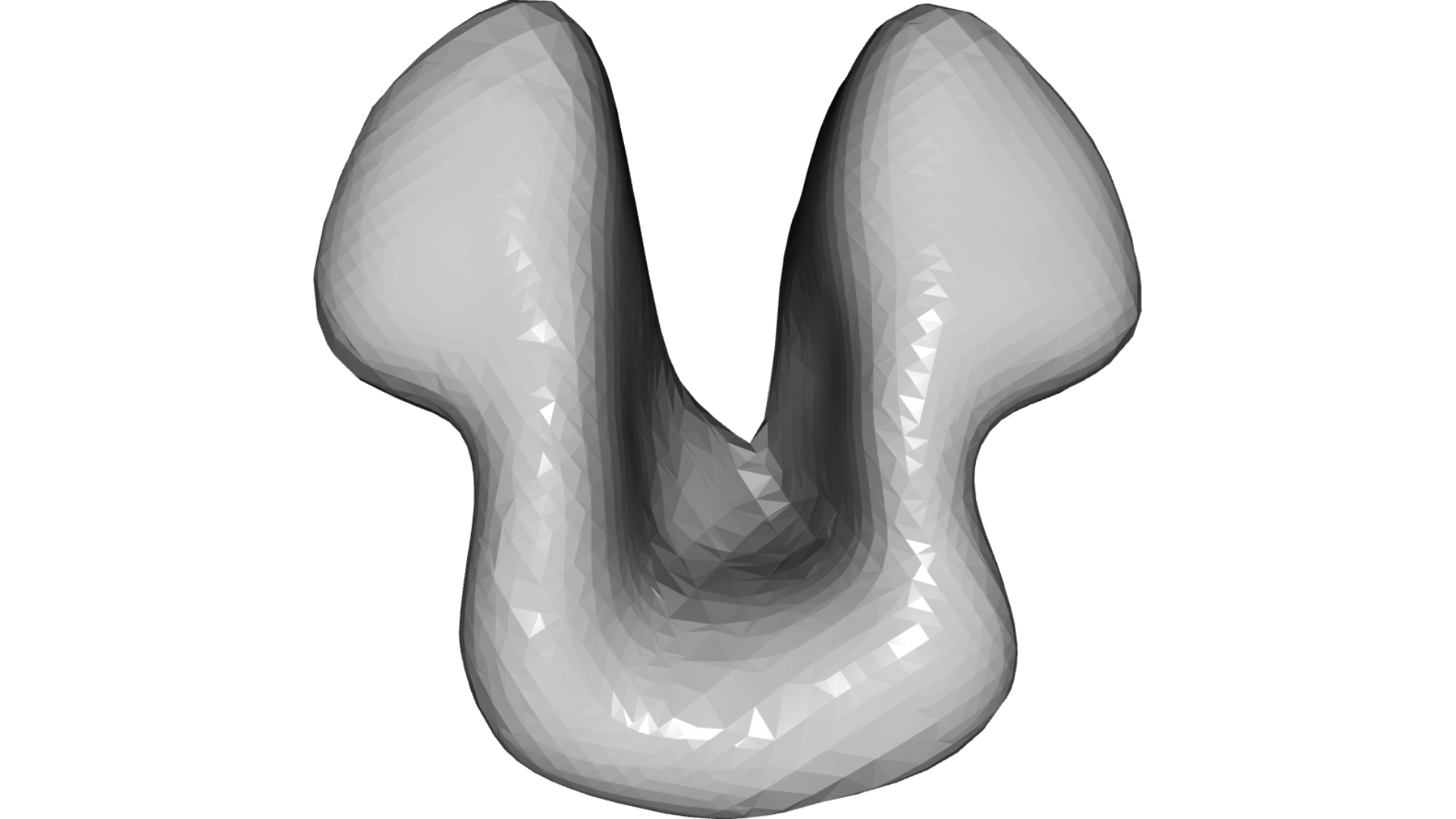}\vspace{3pt}\end{minipage}} & 1.6 (0.19) &\makecell{\begin{minipage}{14mm}\vspace{3pt}\centering\includegraphics[height=8mm]{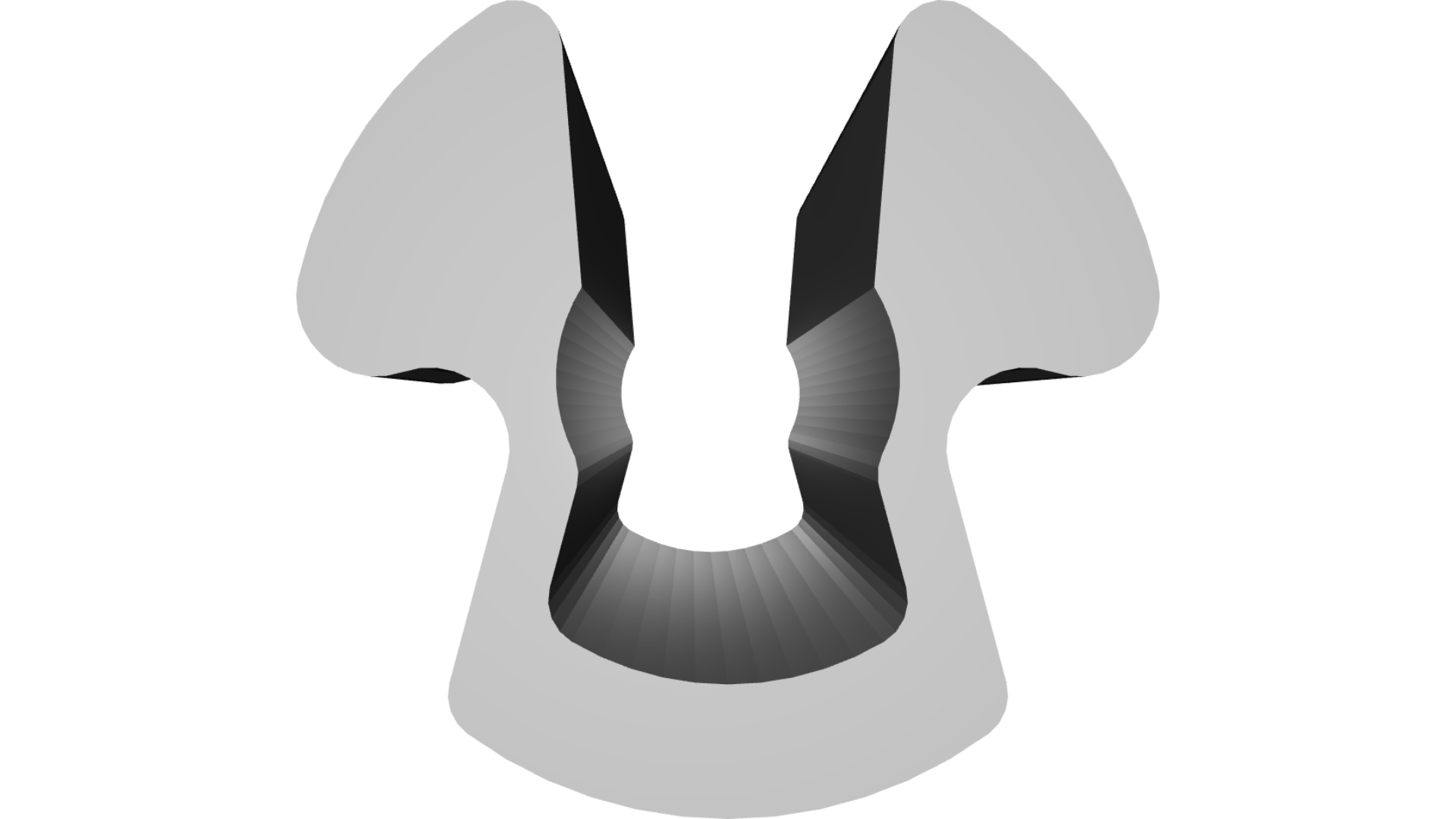}\vspace{3pt}\end{minipage}} &1.4 (0.17)\\Hanger &\makecell{\begin{minipage}{14mm}\vspace{3pt}\centering\includegraphics[height=8mm]{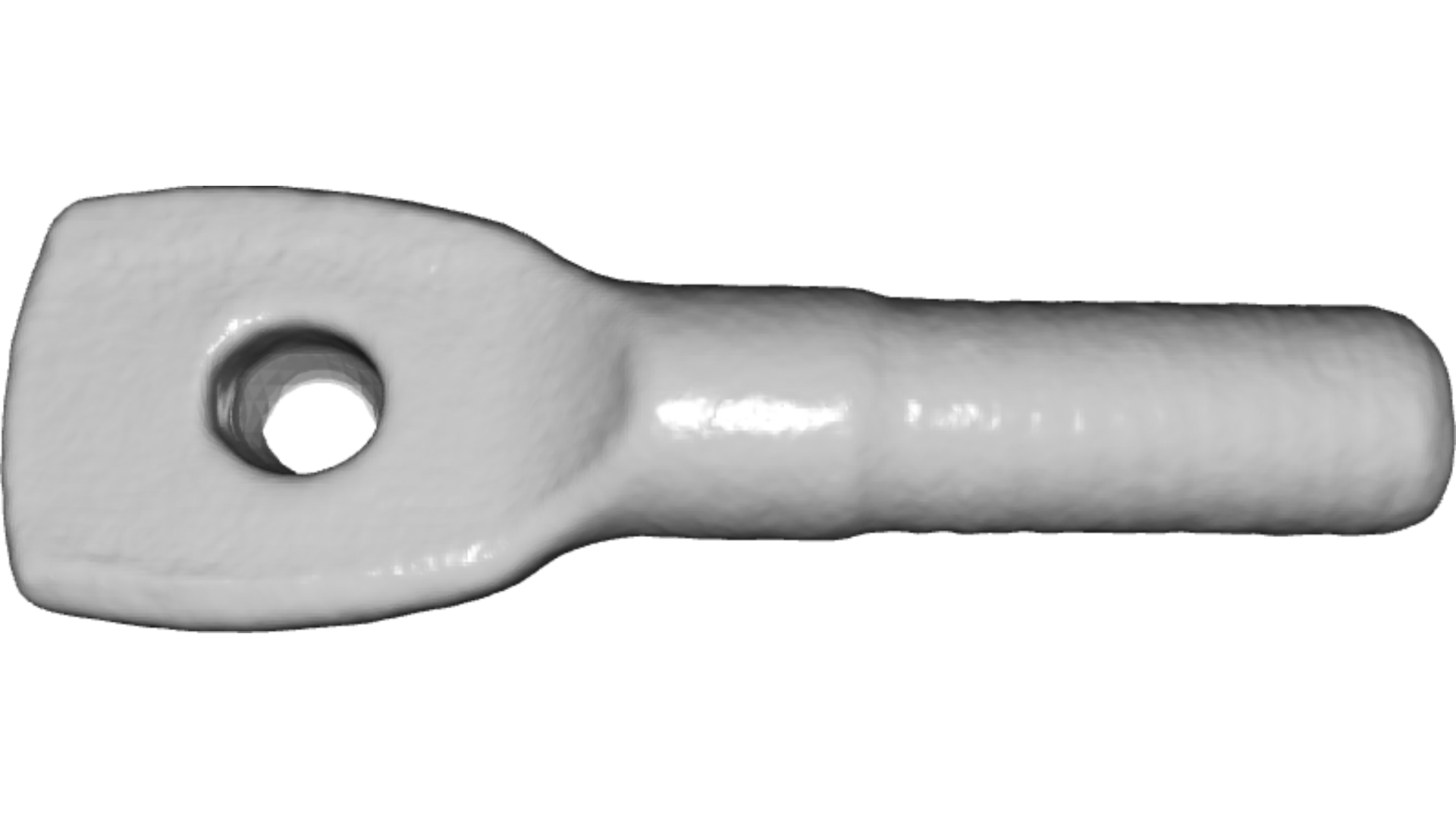}\vspace{3pt}\end{minipage}} & 3.0 (0.39) &\makecell{\begin{minipage}{14mm}\vspace{3pt}\centering\includegraphics[height=8mm]{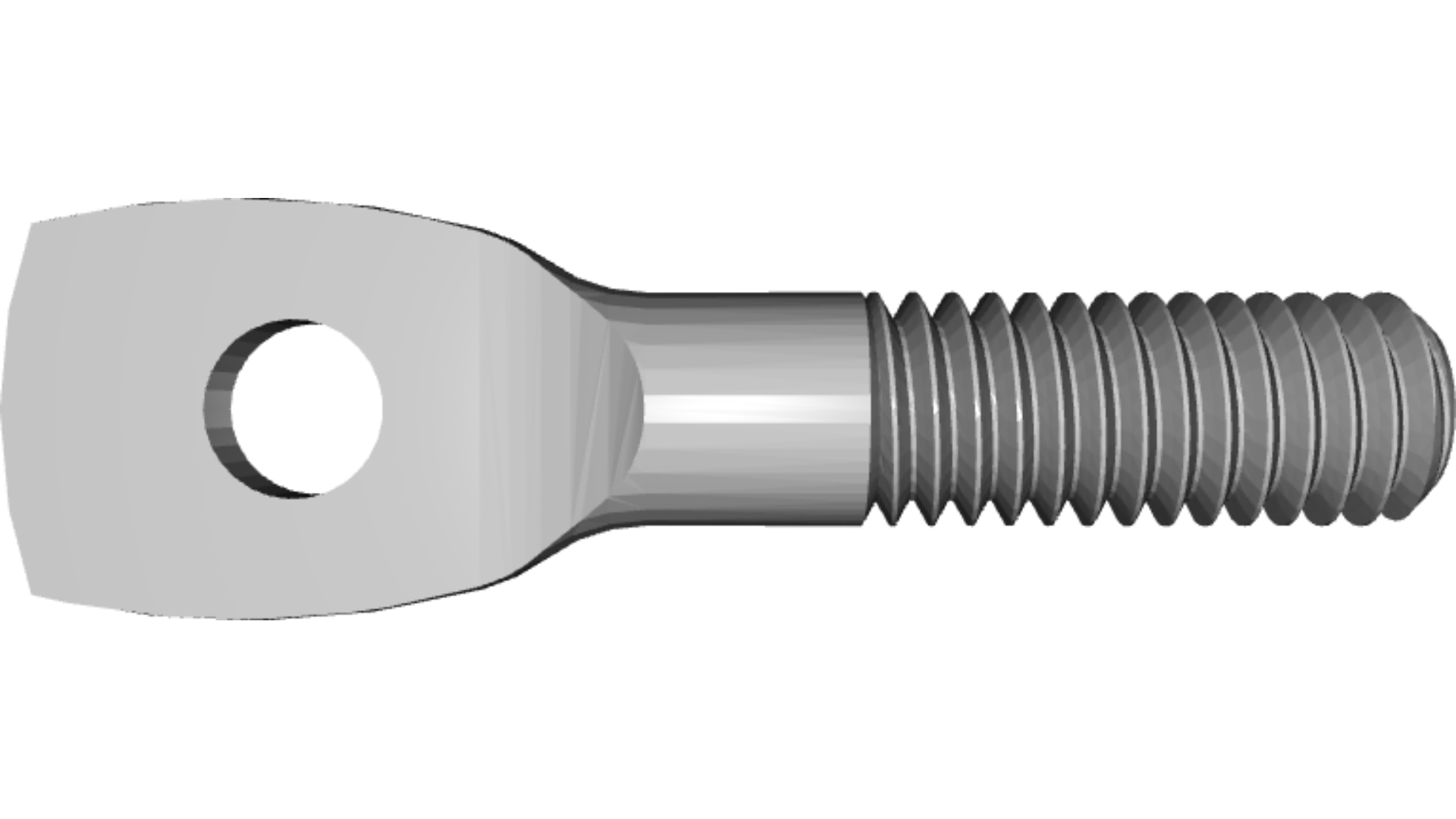}\vspace{3pt}\end{minipage}} &2.4 (0.30)\\Long Grease &\makecell{\begin{minipage}{14mm}\vspace{3pt}\centering\includegraphics[height=8mm]{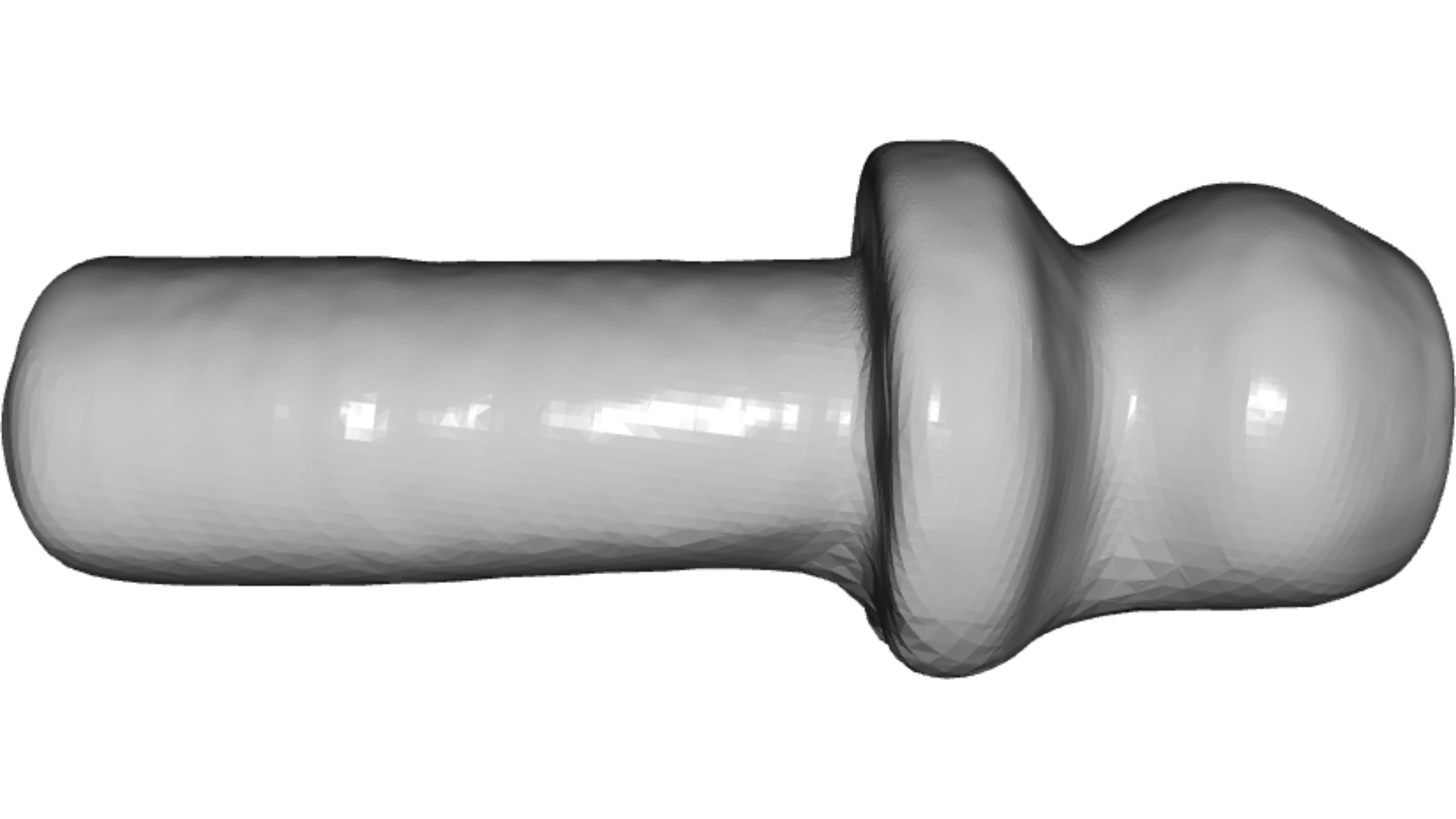}\vspace{3pt}\end{minipage}} & 2.5 (0.33) &\makecell{\begin{minipage}{14mm}\vspace{3pt}\centering\includegraphics[height=8mm]{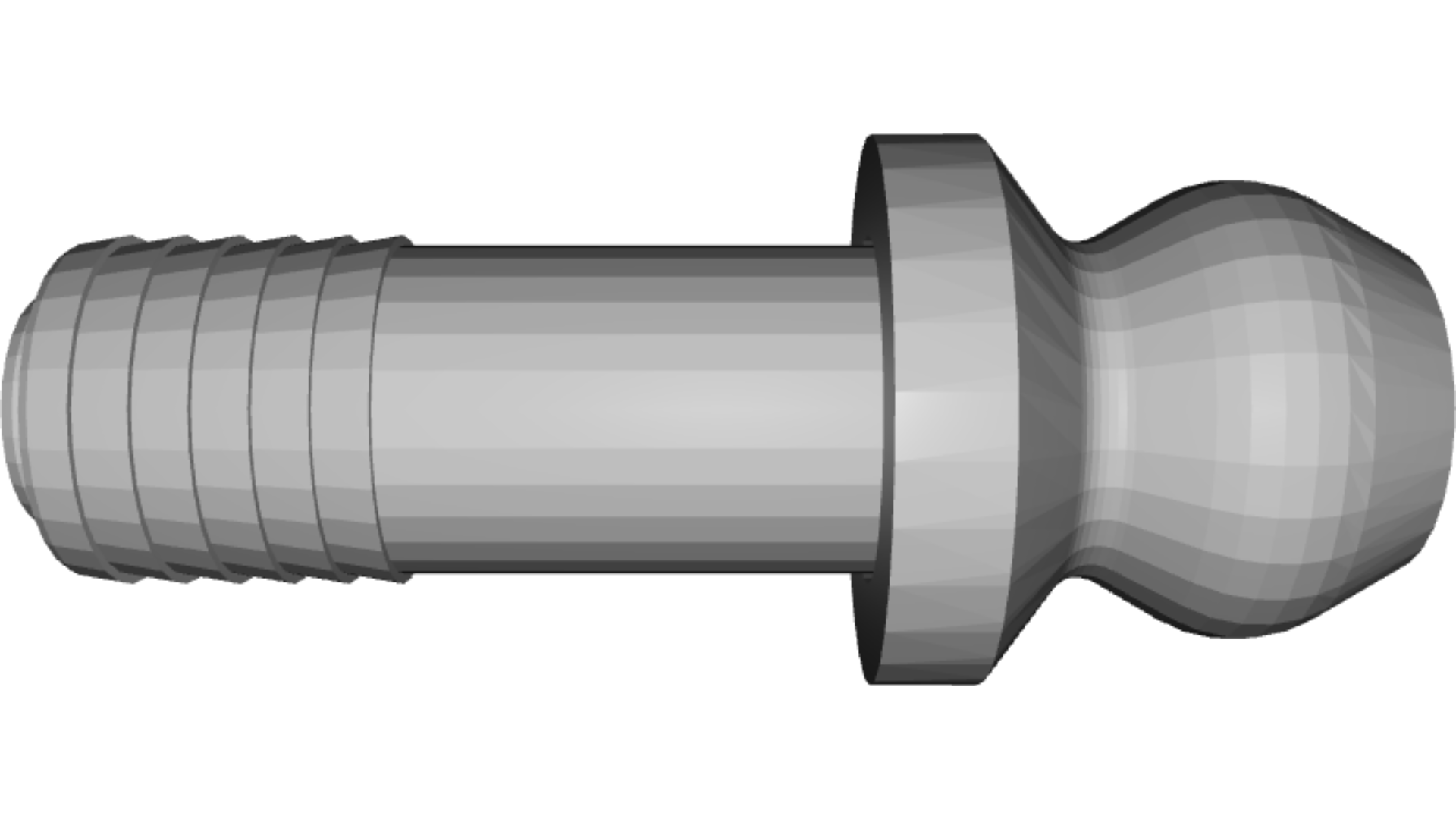}\vspace{3pt}\end{minipage}} &2.3 (0.33)\\Cotter &\makecell{\begin{minipage}{14mm}\vspace{3pt}\centering\includegraphics[height=8mm]{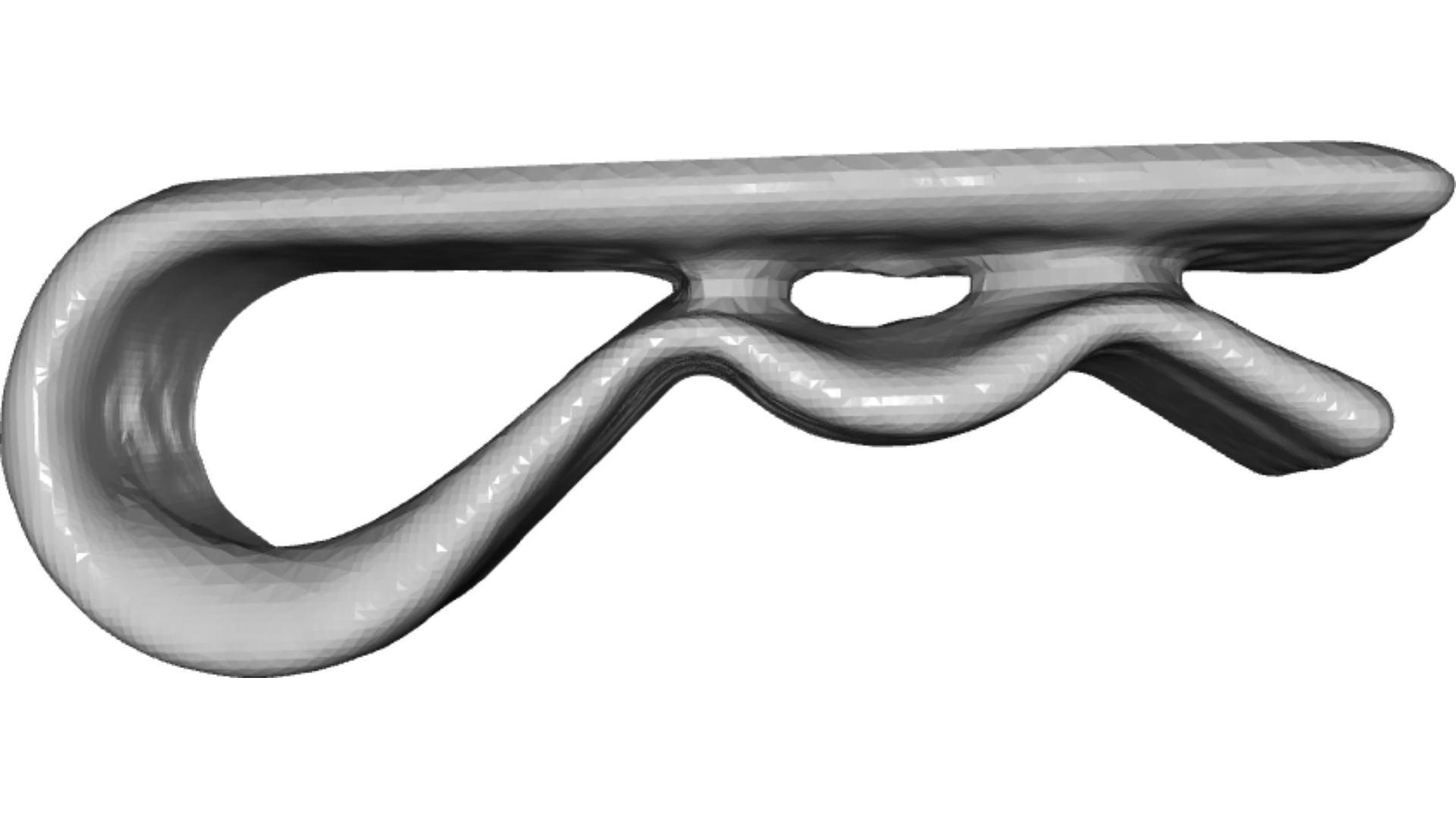}\vspace{3pt}\end{minipage}} & 4.0 (0.51) &\makecell{\begin{minipage}{14mm}\vspace{3pt}\centering\includegraphics[height=8mm]{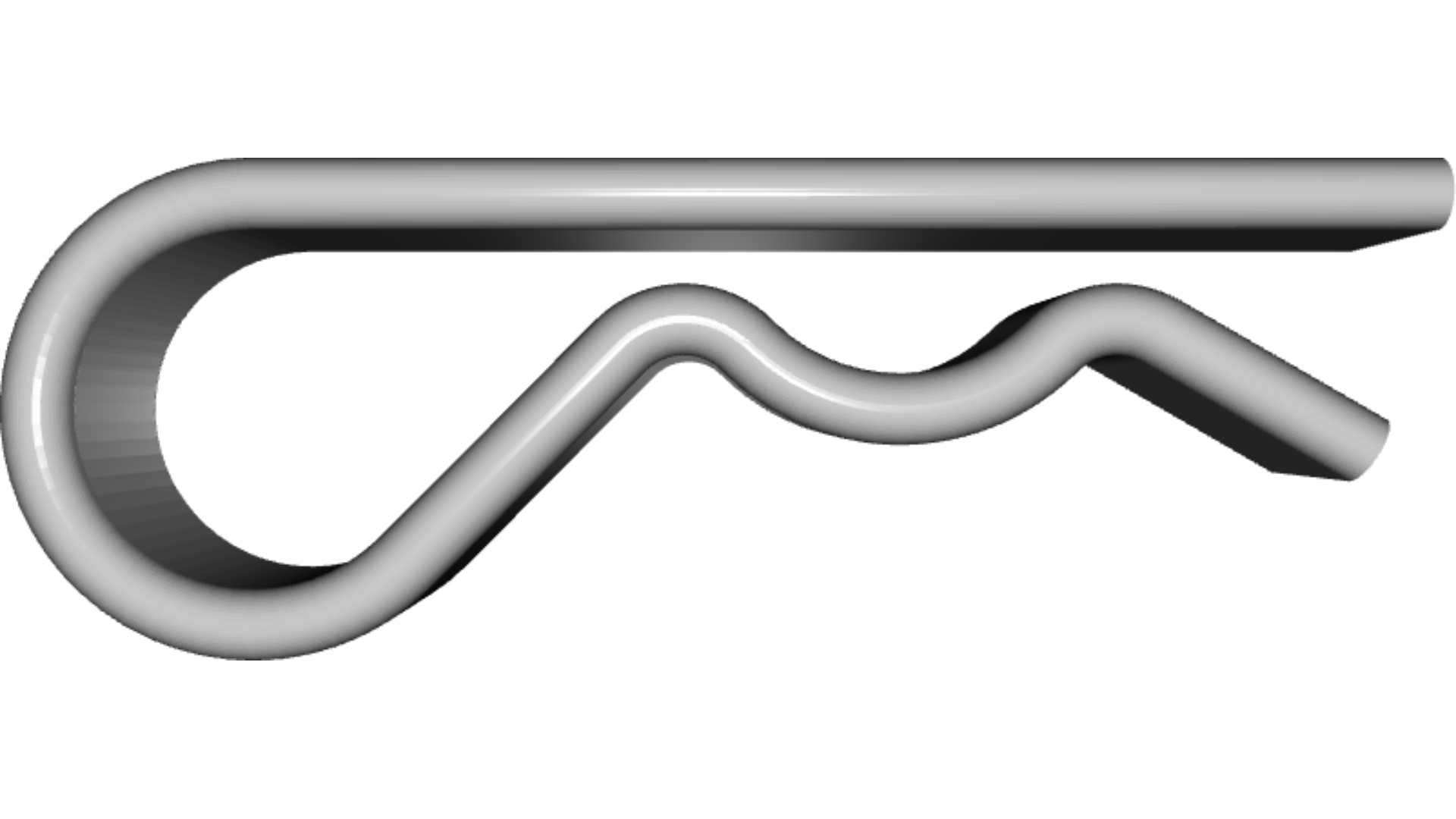}\vspace{3pt}\end{minipage}} &2.9 (0.38)\\Big Head &\makecell{\begin{minipage}{14mm}\vspace{3pt}\centering\includegraphics[height=8mm]{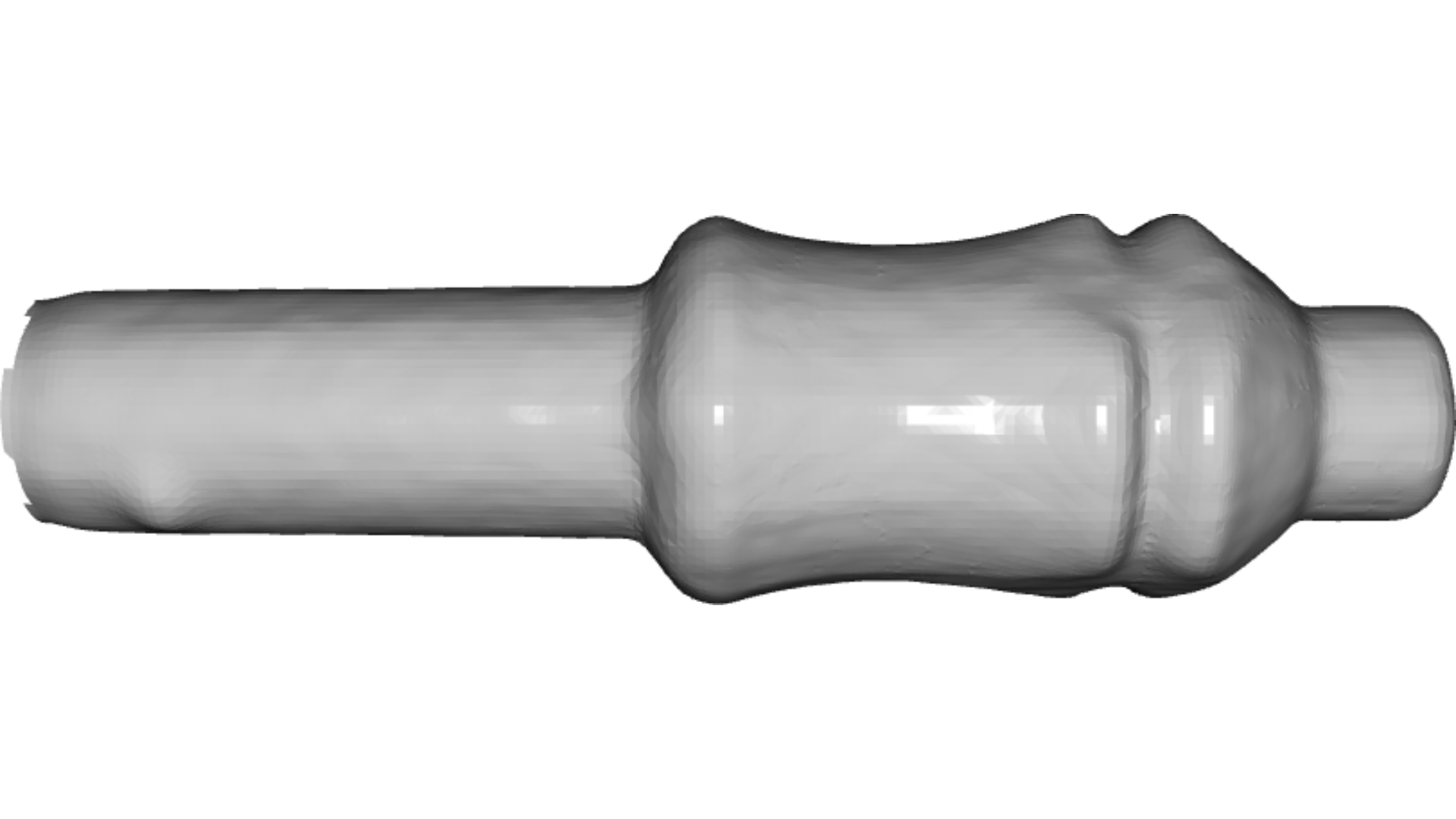}\vspace{3pt}\end{minipage}} & 6.4 (0.88) &\makecell{\begin{minipage}{14mm}\vspace{3pt}\centering\includegraphics[height=8mm]{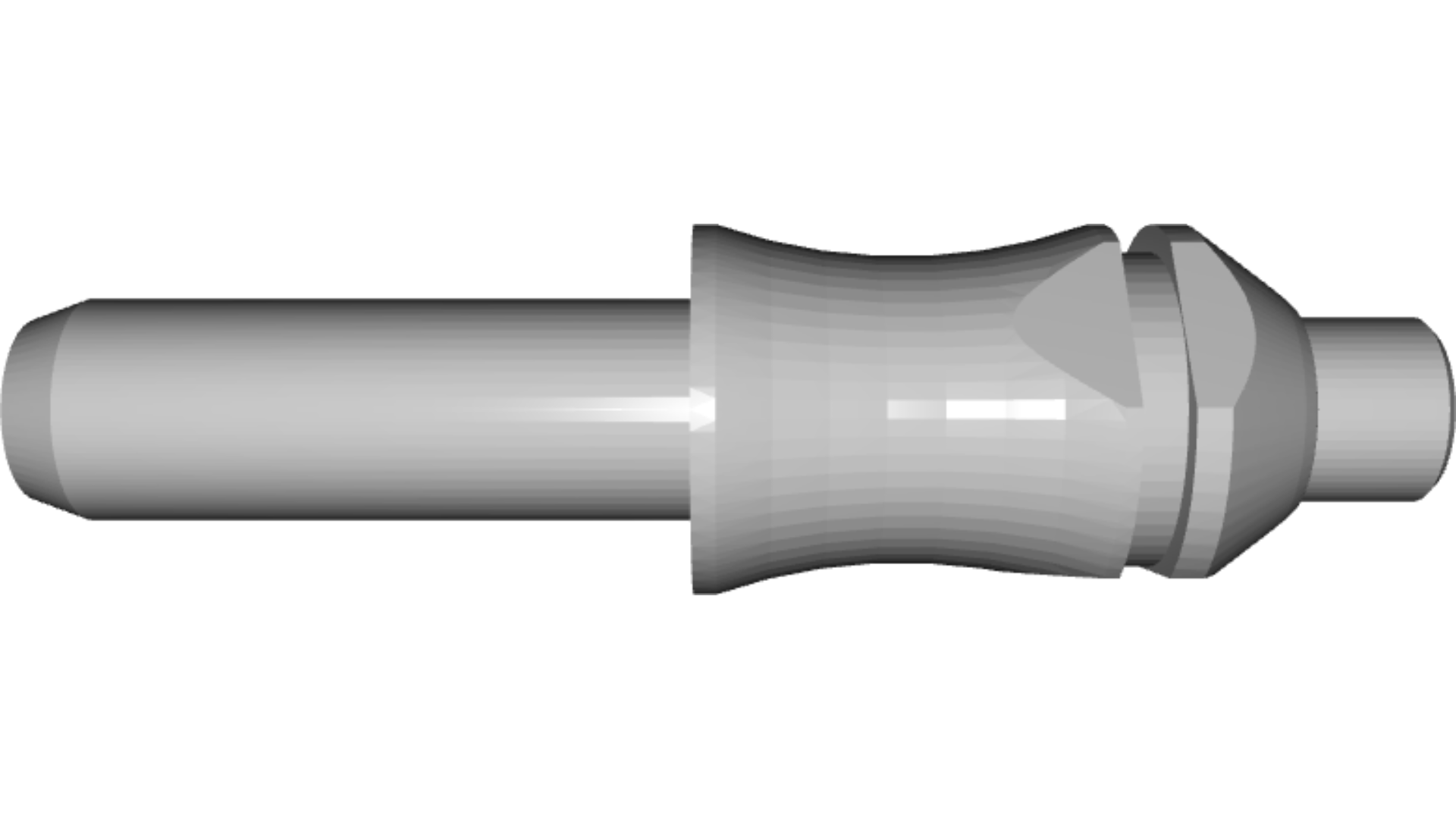}\vspace{3pt}\end{minipage}} &4.9 (0.72)\\\hline\end{tabular}
}
\end{center}\end{table}

\methodd relies on apriori knowledge of the object geometry, in the form of a 3D CAD model. In Section \ref{sec:real_results}, we evaluate \methodd on manufacturer's CAD models. Here, we consider the case where the object model is reconstructed using a 3D scanner. In particular, we use a SOL 3D Scanner by Scan Dimension, to generate the scanned object models.
%

We simulate a grid of possible object contacts and contact poses using the scanned object model. At test time, we compare contacts observed on the real object (same real datasets used in Section~\ref{sec:real_results}) against this imperfect set of simulated contacts.

We evaluate the accuracy of \methodd when
using scanned object models for 5 of the 20 objects, {\fontfamily{lmtt}\selectfont long grease}, {\fontfamily{lmtt}\selectfont snap ring}, {\fontfamily{lmtt}\selectfont big head}, {\fontfamily{lmtt}\selectfont cotter}, and {\fontfamily{lmtt}\selectfont hanger}. We qualitatively compare the scanned and manufacturer's CAD models for each of the 5 objects in Table~\ref{table:scan_results}. The left column shows the result of the 3D scan, and the right column shows the true object model.~\looseness=-1

When evaluating on the scanned object models, we relax the parameter that controls our confidence in the measurement of the gripper opening to reflect the fact that we expect the scanned object models to provide noisy information about the object geometry.
The intuition for this decision is the following: when the object model is known, the gripper opening is a reliable signal of object pose. Therefore, we only want to assign high likelihood to those contact poses that match the observed gripper opening closely. When the object model is reconstructed with a scanner, we cannot be as confident that the gripper opening we measure corresponds closely to the opening of the true nearest contact pose in the grid. This is because the gripper opening computed using the scanned model will be noisy. Therefore, we may want to assign high likelihood to a contact pose, even if the observed gripper opening does not match the opening of the contact pose as closely.

Table~\ref{table:scan_results} compares the localization accuracy on scanned models versus manufacturer's CAD models for the 5 objects, when using parallel jaw contact information with a 10mm pose prior. Although globally the geometry of the object looks similar between the scanned and manufacturer's CAD models, there are clear differences at the local level. This ablation of \methodd allows us to evaluate whether we are still able to achieve significant pose refinement, even when the local geometries are noisy.

We find that the error is comparable to that when using a manufacturer's CAD model for all 5 objects. The normalized error is between 1.01 times ({\fontfamily{lmtt}\selectfont long grease}) and 1.33 times ({\fontfamily{lmtt}\selectfont cotter}) higher when using a scanned model compared with the manufacturer's CAD model. {\fontfamily{lmtt}\selectfont Cotter} is most impacted by using a scanned object model because the contact non-uniqueness becomes less discrete. When the object model is known exactly, many of the contacts look unique with a discrete number of exceptions. Therefore, even a coarse prior on the object pose leads to precise localization. When the object model is noisy, the discrete nature of the non-uniqueness is impacted for a fraction of the contacts, and the median localization error increases.

For {\fontfamily{lmtt}\selectfont snap ring}, {\fontfamily{lmtt}\selectfont hanger}, {\fontfamily{lmtt}\selectfont long grease}, and {\fontfamily{lmtt}\selectfont cotter} \methodd is able to improve on the 10mm prior by a significant amount. For the first three objects, including parallel jaw information on top of the 10mm prior results in localization errors twice as low as selecting a pose at random from the filtered distribution (in Table \ref{table:scan_results}, the normalized pose errors for the scanned objects are less than 0.5). For {\fontfamily{lmtt}\selectfont cotter}, the localization error is nearly twice as low as selecting a pose at random from the filtered distirbution (in Table \ref{table:scan_results}, the normalized pose error is 0.51). For these four objects, \methodd is able to refine the object pose within a small amount of error, even when local geometry is noisy. 
For {\fontfamily{lmtt}\selectfont big head}, the amount of error is comparable (1.22 times higher) to that when using a manufacturer's CAD model. Because {\fontfamily{lmtt}\selectfont big head} has a symmetry-breaking feature around its principle axis, many contacts are continuously non-unique, and therefore the localization error does not improve much when incorporating a coarse prior, even when the object model is known.

\subsection{ Comparison with Baselines}
\label{sec:baselines}
Next, we compare \methodd to pose estimation from tactile images with three baseline methods:

\begin{enumerate}
    \item \textit{Pixel}: We perform direct pixel comparison between the observed contact mask, and each of the contact shapes in the grid.
    \begin{enumerate}
        \item \textit{Single Contact}: We select the grid shape that has the most pixels in common with the observed contact mask as the best match. We take the object pose to be that corresponding to the best match in the grid.
        \item \textit{Parallel Jaw}: We take into account both tactile images and the gripper opening during a parallel jaw grasp. To do so, we compare the pair of observed contact masks, and the observed gripper opening, with triplets of contact masks and gripper openings from the grid. We sum the pixel error from each of the contacts, and the normalized error between the observed opening and the grid opening, to score each of the parallel jaw triplets in the grid. We take the object pose to be the contact pose corresponding to the triplet with the lowest score.
    \end{enumerate}
    \item \textit{Classification}: We formulate the task of pose prediction as a standard classification problem between the elements of the grid. We train a convolutional neural network (CNN) based on ResNet-50 to predict a distribution over the grid elements from a given contact mask. Recall that in \method, the neural network is an encoder that maps contact shapes to vectors and we obtain a distribution over object poses by comparing the encoding from an observed contact shape to all the encodings from the contact shapes in the grid. In contrast, the neural network in the {\fontfamily{lmtt}\selectfont classification} baseline learns to predict the distributions over object poses directly. We train this baseline by using the same data generation as in \method. The inputs are simulated contacts that we obtained from slightly perturbing a pose from the grid and rendering its corresponding contact. For each of these new contacts, we compute its closest grid element and use as training label a vector with length equal to the number of elements as the grid, where all its values are zero except the entry corresponding to the index of the closest element, which has value one. This vector represents a probability distribution that measures the likelihood of each element of the grid to be the closest to the new contact. The loss function is the cross-entropy loss between the predicted likelihood and the classification label. At test time, from an observed tactile image in the form of a contact mask we obtain a distribution over grid elements. 
    \item \textit{Pose}: We train a CNN based on ResNet-50 to regress a nine-element representation of the object pose based on an observed tactile image. The training data consists of contact shapes as inputs, and the contact poses that generated such contacts as labels. Each label is a pose represented by three translational elements, and six rotational elements that can be mapped into a rotation matrix. In comparison with quaternions, our 6D representation of rotations is continuous (meaning that similar orientations are close together in 6D representation), and therefore better suited to regression. We construct the last column of the rotation matrix from the 6D representation by applying Gram-Schmidt as a post-process on the first two columns, and taking the third column as the cross product of the first two~\citep{continuous_rotations}. The loss function is the mean squared error between the output and true nine-element pose. This method for pose prediction does not rely on matching to elements on the grid, but rather predicts the pose of the object directly using supervised regression.
    %
    \end{enumerate}
    Both \methodd and {\fontfamily{lmtt}\selectfont pixel} are compatible with using parallel jaw information, while {\fontfamily{lmtt}\selectfont classification} and {\fontfamily{lmtt}\selectfont pose} only work with single contacts.
    
    We evaluate the performance of \methodd compared to baselines for 5 of the 20 objects: {\fontfamily{lmtt}\selectfont long grease}, {\fontfamily{lmtt}\selectfont snap ring}, {\fontfamily{lmtt}\selectfont big head}, {\fontfamily{lmtt}\selectfont cotter}, and {\fontfamily{lmtt}\selectfont hanger}. We select these 5 objects to cover a range of difficulty in terms of grid size and degrees of symmetry and non-uniqueness.
    
\myparagraph{Baseline comparison with simulated data.} 
\label{sec:baseline_sim}

\begin{table*}[h]\caption{Pose error and normalized pose error (in parenthesis) for \methodd versus baseline methods for a range of grid sizes, using 100 randomly simulated contacts per grid. Columns labelled SC are evaluated with a single contact, while columns labelled PJ use parallel jaw information.}\label{table:sim_baseline_results}\renewcommand{\arraystretch}{1.5}\begin{center}\begin{tabular}{|l l|c c c c c c|}\bottomrule\multicolumn{2}{|c|}{}&\multicolumn{2}{c}{\textbf{\method}} & \multicolumn{2}{c}{\textbf{Pixel}} & \textbf{Classification} & \textbf{Pose}\\\multicolumn{2}{|c|}{}&\makecell{SC\\mm (norm)} & \makecell{PJ\\mm (norm)} & \makecell{SC\\mm (norm)} & \makecell{PJ\\mm (norm)} & \makecell{SC\\mm (norm)} & \makecell{SC\\mm (norm)}\\\hline\multicolumn{1}{|l}{\textbf{Long Grease}} & \multicolumn{1}{l}{\begin{minipage}{9mm}\vspace{2pt}\centering\includegraphics[height=5mm]{figures/scaled_grasp_figures/long_grease.png}\vspace{2pt}\\\end{minipage}} & \multicolumn{6}{c|}{}\\\hline \multicolumn{2}{|l|}{Mini One Face} & 2.3 (0.09) & 2.3 (0.09) & 2.7 (0.11) & 2.4 (0.09) & 2.1 (0.08) & 16.1 (0.57)\\\multicolumn{2}{|l|}{Bigger Mini One Face} & 18.1 (0.51) & 11.5 (0.31) & 27.1 (0.76) & 16.6 (0.45) & 34.8 (0.99) & 25.2 (0.71)\\\multicolumn{2}{|l|}{One Face} & 1.5 (0.04) & 1.3 (0.04) & 12.0 (0.35) & 4.8 (0.14) & 32.7 (0.96) & 7.5 (0.22)\\\multicolumn{2}{|l|}{Full} & 1.3 (0.04) & 1.3 (0.04) & 5.0 (0.14) & 2.2 (0.06) & 33.8 (0.92) & 21.6 (0.59)\\\hline\multicolumn{1}{|l}{\textbf{Snap Ring}} & \multicolumn{1}{l}{\begin{minipage}{9mm}\vspace{2pt}\centering\includegraphics[height=5mm]{figures/scaled_grasp_figures/snap_ring.png}\vspace{2pt}\\\end{minipage}} & \multicolumn{6}{c|}{}\\\hline \multicolumn{2}{|l|}{Mini One Face} & 2.0 (0.16) & 2.0 (0.16) & 2.1 (0.14) & 2.1 (0.15) & 1.2 (0.11) & 8.5 (0.76)\\\multicolumn{2}{|l|}{Bigger Mini One Face} & 3.1 (0.20) & 3.1 (0.22) & 6.9 (0.44) & 6.9 (0.45) & 2.6 (0.16) & 10.1 (0.63)\\\multicolumn{2}{|l|}{Full} & 1.0 (0.06) & 1.0 (0.07) & 1.2 (0.08) & 1.2 (0.08) & 7.2 (0.49) & 4.9 (0.33)\\\hline\multicolumn{1}{|l}{\textbf{Big Head}} & \multicolumn{1}{l}{\begin{minipage}{9mm}\vspace{2pt}\centering\includegraphics[height=5mm]{figures/scaled_grasp_figures/big_head.png}\vspace{2pt}\\\end{minipage}} & \multicolumn{6}{c|}{}\\\hline \multicolumn{2}{|l|}{Mini One Face} & 2.5 (0.09) & 2.3 (0.09) & 3.5 (0.13) & 3.2 (0.12) & 13.4 (0.50) & 21.5 (0.81)\\\multicolumn{2}{|l|}{Bigger Mini One Face} & 37.5 (0.83) & 14.6 (0.34) & 37.9 (0.89) & 31.1 (0.77) & 43.4 (1.04) & 30.5 (0.73)\\\multicolumn{2}{|l|}{One Face} & 1.7 (0.05) & 1.6 (0.05) & 4.8 (0.15) & 4.7 (0.14) & 31.4 (0.97) & 15.6 (0.48)\\\multicolumn{2}{|l|}{Full} & 1.6 (0.04) & 1.6 (0.04) & 9.9 (0.23) & 9.9 (0.23) & 39.6 (0.97) & 13.8 (0.34)\\\hline\multicolumn{1}{|l}{\textbf{Cotter}} & \multicolumn{1}{l}{\begin{minipage}{9mm}\vspace{2pt}\centering\includegraphics[height=5mm]{figures/scaled_grasp_figures/cotter.png}\vspace{2pt}\\\end{minipage}} & \multicolumn{6}{c|}{}\\\hline \multicolumn{2}{|l|}{Mini One Face} & 2.5 (0.08) & 2.5 (0.10) & 15.5 (0.52) & 17.7 (0.60) & 2.1 (0.07) & 19.6 (0.71)\\\multicolumn{2}{|l|}{Bigger Mini One Face} & 8.3 (0.21) & 8.8 (0.21) & 34.0 (0.87) & 34.0 (0.87) & 15.2 (0.39) & 22.2 (0.56)\\\multicolumn{2}{|l|}{One Face} & 1.3 (0.03) & 1.4 (0.03) & 5.8 (0.15) & 6.2 (0.16) & 35.2 (0.90) & 8.5 (0.22)\\\multicolumn{2}{|l|}{Full} & 13.8 (0.34) & 8.9 (0.22) & 17.1 (0.44) & 18.2 (0.42) & 34.9 (0.88) & 18.2 (0.46)\\\hline\multicolumn{1}{|l}{\textbf{Hanger}} & \multicolumn{1}{l}{\begin{minipage}{9mm}\vspace{2pt}\centering\includegraphics[height=5mm]{figures/scaled_grasp_figures/hanger.png}\vspace{2pt}\\\end{minipage}} & \multicolumn{6}{c|}{}\\\hline \multicolumn{2}{|l|}{Mini One Face} & 2.3 (0.08) & 2.2 (0.09) & 2.5 (0.09) & 2.6 (0.10) & 2.8 (0.12) & 9.8 (0.42)\\\multicolumn{2}{|l|}{Bigger Mini One Face} & 9.1 (0.24) & 5.1 (0.14) & 25.2 (0.71) & 25.1 (0.71) & 21.1 (0.60) & 19.6 (0.56)\\\multicolumn{2}{|l|}{One Face} & 1.5 (0.04) & 1.4 (0.04) & 4.0 (0.11) & 3.3 (0.09) & 30.9 (0.88) & 9.0 (0.26)\\\multicolumn{2}{|l|}{Full} & 2.6 (0.08) & 1.9 (0.06) & 6.6 (0.19) & 4.2 (0.12) & 34.2 (0.95) & 6.4 (0.18)\\\hline\end{tabular}\end{center}\end{table*}

    \begin{figure*}[h]
        \begin{subfigure}[t]{0.5\linewidth}
            \includegraphics[width=\linewidth]{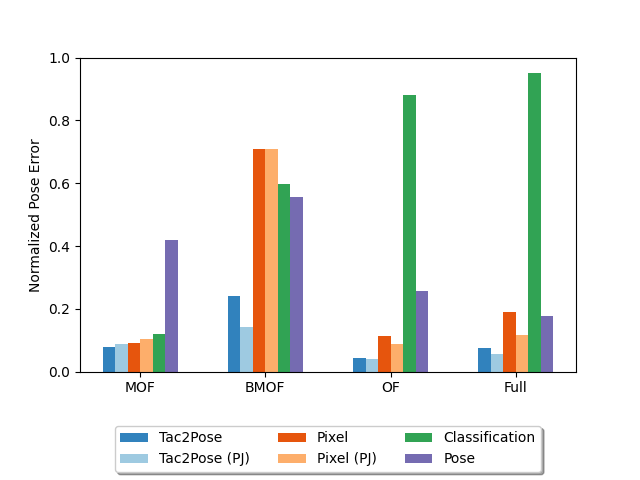}
            \caption{Results for 100 randomly sampled \textit{simulated} contacts across four {\fontfamily{lmtt}\selectfont hanger} grid sizes. Mini One Face (MOF) is the smallest and simplest grid, while Full is the largest and most complex.}
            \label{fig:baseline_b}
        \end{subfigure}
        \quad
        \begin{subfigure}[t]{0.5\linewidth}
            \includegraphics[width=\linewidth]{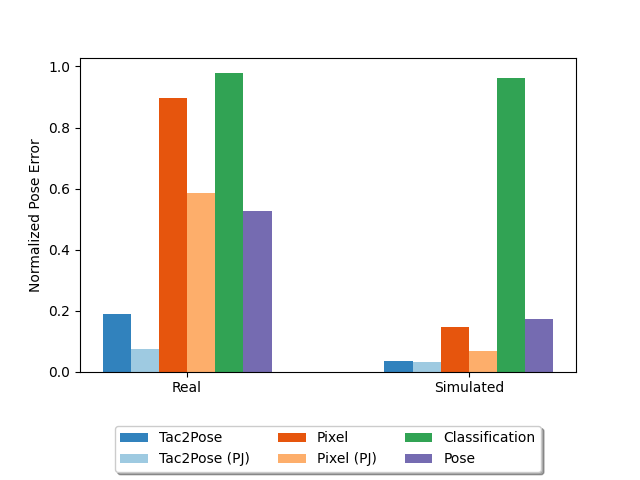}
            \caption{Results on the {\fontfamily{lmtt}\selectfont hanger} dataset (Section~\ref{sec:real_results}) with over 100 contacts, using the full-sized grid for all methods. We evaluate all methods on the same \textit{real} (left) and \textit{simulated} (right) contact shapes.}
            \label{fig:baseline_a}
        \end{subfigure}
        \caption{Normalized pose error for the object {\fontfamily{lmtt}\selectfont hanger} using tactile matching (\method) and three baseline methods. We compare the performance of each method on simulated contacts from different grid sizes in (\ref{fig:baseline_b}), and on real vs. simulated versions of the contacts in the {\fontfamily{lmtt}\selectfont hanger} dataset (Section~\ref{sec:real_results}) in (\ref{fig:baseline_a}).}
        \label{fig:baseline_figure}
    \end{figure*}
    We first evaluate each method on sets of 100 simulated contacts per object. We sample 100 contact poses from the object's grid, add noise to each pose, and render the contacts corresponding to the new poses.
    
    For each of the 5 baseline objects, we create three smaller, simpler, grids in addition to the original. The reduced grids serve to test each method's ability to scale to different problem sizes and types of complexity.
    
    Recall that the original grids are determined by 4 spatial coordinates (grasp approach direction, x, y, $\theta$) as illustrated in Figure~\ref{fig:grid dimensions}. They have angular resolution of 6 degrees, and x, y translational resolution of 2.5mm. The size of the original grids for the baseline objects ranges from 8.1k to 91.5k poses. The reduced grids are modified as follows:
    \begin{enumerate}
        \item \textit{Mini One Face}: We include only one grasp approach direction, and one angle. We consider translations in x, y with 5mm resolution. This grid is much smaller than the original (ranging from 29 to 124 poses, depending on the object), and is effectively 2D (x, y).
        \item \textit{Bigger Mini One Face}: We include only one grasp approach direction, and 10 angles (angular resolution of 36 degrees). We consider translations in x, y with 5mm resolution. This grid is smaller than the original, and ranges from 331 to 1137 poses, depending on the object. It is effectively 3D (x, y, $\theta$).
        \item \textit{One Face}: We include only one grasp approach direction, with full angular (6 degrees) and translational (2.5mm) resolution. This grid ranges in size from 4.8k to 26.8k poses, depending on the object. It is effectively 3D (x, y, $\theta$). For {\fontfamily{lmtt}\selectfont snap ring}, which only has one grasp approach direction even in the full sized grid, this grid is equivalent to the original and is therefore omitted.
    \end{enumerate}

    Comparing results between the reduced grids can provide insight into each method's sensitivity to problem size, resolution, and complexity. For example, comparing {\fontfamily{lmtt}\selectfont mini one face} and {\fontfamily{lmtt}\selectfont one face} can indicate the sensitivity of each method to grid size and resolution. Comparing {\fontfamily{lmtt}\selectfont one face} and the original grid (called {\fontfamily{lmtt}\selectfont full} from here on out), which have the same resolution but cover different faces, provides insight into how each method handles complexity in the form of 3D rotations. {\fontfamily{lmtt}\selectfont Mini one face} and {\fontfamily{lmtt}\selectfont bigger mini one face} have the same translational resolution, but {\fontfamily{lmtt}\selectfont bigger mini one face} varies the angle in the plane of the grasp. Comparing results on these two grids provides insight into how incorporating rotations in the plane of the grasp impacts the accuracy of each method.
    
    We first conduct experiments using only simulated data. The full set of results on the 5 baseline objects are listed in Table~\ref{table:sim_baseline_results}. We consider the results for {\fontfamily{lmtt}\selectfont hanger}, visualized in Figure \ref{fig:baseline_b}, for simulated contacts from {\fontfamily{lmtt}\selectfont mini one face}, {\fontfamily{lmtt}\selectfont bigger mini one face}, {\fontfamily{lmtt}\selectfont one face}, and {\fontfamily{lmtt}\selectfont full}. The trends we note for this object are representative of the trends for the remaining objects.

    The performance of {\fontfamily{lmtt}\selectfont classification} (\textcolor{Green}{green} bar) is inversely correlated with grid size; its performance is comparable to \methodd only for {\fontfamily{lmtt}\selectfont mini one face} (MOF in Figure \ref{fig:baseline_b}), which has 124 contacts. The normalized error is 0.12, compared with 0.08 for the single contact case of \method. This indicates that although {\fontfamily{lmtt}\selectfont classification} could be effective for very small grid sizes, the grids quickly become too large. Even with {\fontfamily{lmtt}\selectfont bigger mini one face}, which contains 794 transformations, {\fontfamily{lmtt}\selectfont classification} has a normalized error of 0.6, making it more than six times worse than \methodd (single contact), which has 0.09 normalized error.
    
    The {\fontfamily{lmtt}\selectfont classification} baseline struggles with larger grid sizes because it needs to provide a classification over all elements in the grid, which often contains thousands of elements (the {\fontfamily{lmtt}\selectfont full} grid for {\fontfamily{lmtt}\selectfont hanger}, for example, has more than 90k contact poses).
    In comparison with the {\fontfamily{lmtt}\selectfont classification} baseline, \methodd scales better to large grid sizes because it learns to generate an embedding space based on the distance between contact poses. The encoder in \methodd learns to push distant contacts away in embedding space, and therefore when comparing a new encoding to all the encodings in the grid, this direct comparison scales better than {\fontfamily{lmtt}\selectfont classification} when the set of contacts in the grid is large. Instead, in {\fontfamily{lmtt}\selectfont classification} the learned NN needs to encapsulate all the information relevant about all poses solely in the NN weights. Another way to think about the difference is to consider \methodd as describing each contact in the grid with 1000 parameters (the encoder), whereas with {\fontfamily{lmtt}\selectfont classification}, each contact in the grid is described by a single parameter (0 or 1).
    
    The {\fontfamily{lmtt}\selectfont pose} baseline (\textcolor{violet}{purple} bar), on the other hand, is most comparable to \methodd for the larger grids with high resolution. For {\fontfamily{lmtt}\selectfont one face} (OF in Figure~\ref{fig:baseline_b}), which covers one object face with the same translational and rotational resolution as the {\fontfamily{lmtt}\selectfont full} grid, the normalized pose error using {\fontfamily{lmtt}\selectfont pose} is 0.26, compared with 0.04 for the single contact version of \method. For the {\fontfamily{lmtt}\selectfont full} grid, the normalized error for {\fontfamily{lmtt}\selectfont pose} and the single contact version of \methodd is 0.18 and 0.08, respectively. In the absence of contact shape noise (simulated contacts), \methodd is about 2.3 times better on the {\fontfamily{lmtt}\selectfont full} grid.
    Recall that the {\fontfamily{lmtt}\selectfont pose} baseline does not match observed contacts to contact poses on the grid, but instead regresses the corresponding contact pose directly. 
    %
    In pose, the learned NN needs to encapsulate all information relevant about all poses solely in the NN weights, as it doesn't have access to embeddings it can compare against.
    
    The {\fontfamily{lmtt}\selectfont pixel} baseline, for both single contact and parallel jaw contact cases, performs most comparably to \methodd for the {\fontfamily{lmtt}\selectfont mini one face} grid. The {\fontfamily{lmtt}\selectfont mini one face} grid is small and simple enough (one angle, one face) that all methods (excluding {\fontfamily{lmtt}\selectfont pose}) perform comparably. For {\fontfamily{lmtt}\selectfont one face} and {\fontfamily{lmtt}\selectfont full}, \methodd outperforms {\fontfamily{lmtt}\selectfont pixel} by about 2-3 times for both single contact and parallel jaw cases, which is still relatively comparable. This implies that {\fontfamily{lmtt}\selectfont pixel} is most successful for high resolution grids. 
    The performance of the {\fontfamily{lmtt}\selectfont pixel} baseline method degrades most when the grid is low resolution, particularly when the grid contains object rotations. {\fontfamily{lmtt}\selectfont Bigger mini one face} (shown as BMOF in the figure) has 0.71 median normalized error, compared with 0.24 for \methodd (single contact). With parallel jaw contacts, the {\fontfamily{lmtt}\selectfont pixel} baseline has 0.71 median normalized error, compared with 0.14 for\method. \methodd outperforms {\fontfamily{lmtt}\selectfont pixel} by nearly 3 times in the single contact case, and more than 5 times in the parallel jaw case. Furthermore, {\fontfamily{lmtt}\selectfont pixel} (both single contact and parallel jaw contact cases) is the least effective method when the grid is low resolution but complex.

    When the grid resolution is low, the nearest grid match will be farther (in pixel distance) from an observed contact than when the grid resolution is high. This is particularly harmful for {\fontfamily{lmtt}\selectfont pixel} when rotations are introduced, because a rotated version of a remote contact may, by chance, have more pixel overlap with the observed contact than the true closest match. This is much less likely to occur for high resolution grids, in which there exists a very close match on the grid to any observed contact. When the grid is simple (does not contain planar or 3D rotations) {\fontfamily{lmtt}\selectfont pixel} can be successful even when the grid resolution is low (as is the case for {\fontfamily{lmtt}\selectfont mini one face}) because the best match on the grid is still likely to have the smallest pixel distance. If the grid is sufficiently complex, it must also be high resolution for {\fontfamily{lmtt}\selectfont pixel} to be relatively successful (as is the case for {\fontfamily{lmtt}\selectfont one face} and {\fontfamily{lmtt}\selectfont full}).
    It is worth noting that a drawback of {\fontfamily{lmtt}\selectfont pixel}, particularly with high resolution grids, is the execution time. In order to choose the best match, an observed contact is compared with contact shapes corresponding to every contact pose in the grid. Execution time is doubled when considering parallel jaw information, because two contacts are compared for each contact pose. {\fontfamily{lmtt}\selectfont Pixel} is therefore the slowest method we evaluated by a significant margin. In practice, matching a contact to a high resolution grid in real-time using {\fontfamily{lmtt}\selectfont pixel} is likely infeasible. Instead, \methodd only compares low-dimensional embeddings resulting in at least an order of magnitude speed-up.
    
    Finally, \methodd performs consistently for both single and parallel jaw simulated contacts for all grid sizes for {\fontfamily{lmtt}\selectfont hanger}. \methodd also outperforms all baselines for all grid sizes. The parallel jaw case of \methodd slightly outperforms the single contact case for all {\fontfamily{lmtt}\selectfont hanger} grid sizes, except {\fontfamily{lmtt}\selectfont mini one face}, in which the single contact case outperforms parallel jaw by 0.08 versus 0.09 normalized error. \methodd (both single contact and parallel jaw) also performs slightly better for high resolution grids ({\fontfamily{lmtt}\selectfont one face} and {\fontfamily{lmtt}\selectfont full}). 
    
    \myparagraph{Baseline comparison with real data.}
    \label{sec:baseline_real}
    
    \begin{table*}[h]\caption{Pose error and normalized pose error (in parenthesis) for \methodd versus baseline methods, on \textit{real} datasets. Columns labelled SC are evaluated with a single contact, while columns labelled PJ use parallel jaw information.}\label{table:baseline_results}\renewcommand{\arraystretch}{1.5}\begin{center}\begin{tabular}{|c|c|c c c c c c|}\bottomrule\multicolumn{2}{|c|}{}&\multicolumn{2}{c}{\textbf{\method}} & \multicolumn{2}{c}{\textbf{Pixel}} & \textbf{Classification} & \textbf{Pose}\\\multicolumn{2}{|c|}{}& \makecell{SC\\mm (norm)} & \makecell{PJ\\mm (norm)} & \makecell{SC\\mm (norm)} & \makecell{PJ\\mm (norm)} & \makecell{SC\\mm (norm)} & \makecell{SC\\mm (norm)}\\\cline{1-8}Long Grease &\makecell{\begin{minipage}{14mm}\vspace{3pt}\centering\includegraphics[height=8mm]{figures/scaled_grasp_figures/long_grease.png}\vspace{3pt}\\\end{minipage}} & 26.6 (0.76) & 3.3 (0.09) & 32.8 (0.93) & 6.0 (0.17) & 33.3 (0.95) & 25.3 (0.72)\\Snap Ring &\makecell{\begin{minipage}{14mm}\vspace{3pt}\centering\includegraphics[height=8mm]{figures/scaled_grasp_figures/snap_ring.png}\vspace{3pt}\\\end{minipage}} & 1.5 (0.10) & 1.4 (0.10) & 5.6 (0.39) & 2.2 (0.15) & 6.0 (0.42) & 5.9 (0.41)\\Big Head &\makecell{\begin{minipage}{14mm}\vspace{3pt}\centering\includegraphics[height=8mm]{figures/scaled_grasp_figures/big_head.png}\vspace{3pt}\\\end{minipage}} & 7.8 (0.20) & 6.1 (0.16) & 27.6 (0.70) & 11.7 (0.30) & 35.0 (0.89) & 33.8 (0.86)\\Cotter &\makecell{\begin{minipage}{14mm}\vspace{3pt}\centering\includegraphics[height=8mm]{figures/scaled_grasp_figures/cotter.png}\vspace{3pt}\\\end{minipage}} & 19.0 (0.49) & 19.6 (0.51) & 31.5 (0.81) & 36.7 (0.95) & 35.8 (0.93) & 38.1 (0.99)\\Hanger &\makecell{\begin{minipage}{14mm}\vspace{3pt}\centering\includegraphics[height=8mm]{figures/scaled_grasp_figures/hanger.png}\vspace{3pt}\\\end{minipage}} & 6.6 (0.19) & 2.6 (0.07) & 31.3 (0.90) & 20.5 (0.59) & 34.2 (0.98) & 18.3 (0.53)\\\hline\end{tabular}\end{center}\end{table*}
    
    \begin{table*}[h]\caption{Pose error and normalized pose error (in parenthesis) for \methodd versus baseline methods, on \textit{simulated} versions of the real contacts used in Table~\ref{table:baseline_results}. Columns labelled SC are evaluated with a single contact, while columns labelled PJ use parallel jaw information.}\label{table:baseline_simreal_results}\renewcommand{\arraystretch}{1.5}\begin{center}\begin{tabular}{|c|c|c c c c c c|}\bottomrule\multicolumn{2}{|c|}{}&\multicolumn{2}{c}{\textbf{\method}} & \multicolumn{2}{c}{\textbf{Pixel}} & \textbf{Classification} & \textbf{Pose}\\\multicolumn{2}{|c|}{}& \makecell{SC\\mm (norm)} & \makecell{PJ\\mm (norm)} & \makecell{SC\\mm (norm)} & \makecell{PJ\\mm (norm)} & \makecell{SC\\mm (norm)} & \makecell{SC\\mm (norm)}\\\cline{1-8}Long Grease &\makecell{\begin{minipage}{14mm}\vspace{3pt}\centering\includegraphics[height=8mm]{figures/scaled_grasp_figures/long_grease.png}\vspace{3pt}\\\end{minipage}} & 1.1 (0.03) & 1.1 (0.03) & 1.8 (0.05) & 1.7 (0.05) & 33.0 (0.94) & 17.0 (0.48)\\Snap Ring &\makecell{\begin{minipage}{14mm}\vspace{3pt}\centering\includegraphics[height=8mm]{figures/scaled_grasp_figures/snap_ring.png}\vspace{3pt}\\\end{minipage}} & 1.0 (0.07) & 1.0 (0.07) & 1.2 (0.08) & 1.2 (0.08) & 6.0 (0.41) & 4.5 (0.31)\\Big Head &\makecell{\begin{minipage}{14mm}\vspace{3pt}\centering\includegraphics[height=8mm]{figures/scaled_grasp_figures/big_head.png}\vspace{3pt}\\\end{minipage}} & 3.9 (0.10) & 3.3 (0.08) & 8.7 (0.22) & 9.6 (0.24) & 34.0 (0.87) & 13.8 (0.35)\\Cotter &\makecell{\begin{minipage}{14mm}\vspace{3pt}\centering\includegraphics[height=8mm]{figures/scaled_grasp_figures/cotter.png}\vspace{3pt}\\\end{minipage}} & 1.3 (0.03) & 1.3 (0.03) & 17.9 (0.46) & 17.7 (0.46) & 35.0 (0.91) & 13.2 (0.34)\\Hanger &\makecell{\begin{minipage}{14mm}\vspace{3pt}\centering\includegraphics[height=8mm]{figures/scaled_grasp_figures/hanger.png}\vspace{3pt}\\\end{minipage}} & 1.2 (0.03) & 1.2 (0.03) & 5.1 (0.15) & 2.4 (0.07) & 33.6 (0.96) & 6.0 (0.17)\\\hline\end{tabular}\end{center}\end{table*}
    
    We next compare \methodd against the three baselines, using the real datasets evaluated in Section~\ref{sec:real_results} and full sized grids. The full set of results are listed in Table~\ref{table:baseline_results}. We also evaluate simulated versions of the contacts in the real datasets, to assess each method's sensitivity to noise in the contact shapes. The results on simulated versions of the contacts in the real datasets are listed in Table \ref{table:baseline_simreal_results}.
    We discuss in detail the results for {\fontfamily{lmtt}\selectfont hanger}, visualized in Figure~\ref{fig:baseline_a}. The trends for this object are representative of the trends we see overall.~\looseness=-1

    For real data, and full-sized grids, the performance {\fontfamily{lmtt}\selectfont classification} (\textcolor{Green}{green} bar), and {\fontfamily{lmtt}\selectfont pixel} with a single contact (\textcolor{orange}{dark orange} bar) are similar to selecting a contact pose randomly from the grid. The median normalized error is 0.98 for {\fontfamily{lmtt}\selectfont classification}, and 0.90 for {\fontfamily{lmtt}\selectfont pixel} with a single contact. {\fontfamily{lmtt}\selectfont Pixel} with parallel jaw contacts and {\fontfamily{lmtt}\selectfont pose} perform better, and have errors almost two times lower than selecting a pose at random from the grid; the median normalized error is 0.59 for {\fontfamily{lmtt}\selectfont pixel} with parallel jaw contacts, and 0.53 for {\fontfamily{lmtt}\selectfont pose}.
    \methodd (\textcolor{blue}{dark blue} bar for single contact, \textcolor{cyan}{light blue} bar for parallel jaw contacts) significantly outperforms all baselines. The median normalized error for the single contact case is 0.19, and for the parallel jaw contact case is 0.07. \methodd outperforms the next best baseline ({\fontfamily{lmtt}\selectfont pose}) by nearly three times for the single contact case, and nearly eight times for the parallel jaw contact case.

    On simulated versions of the same contacts, {\fontfamily{lmtt}\selectfont pose} and {\fontfamily{lmtt}\selectfont pixel} (single contact and parallel jaw contact cases) both perform better than they do with real contacts. The median normalized error for {\fontfamily{lmtt}\selectfont pose} shrinks by over three times, dropping to 0.17 with simulated data. Similarly, for {\fontfamily{lmtt}\selectfont pixel}, the normalized error in simulation shrinks six times for single contact, and over eight times for parallel jaw contacts, dropping to 0.15 and 0.07, respectively. This indicates the sensitivity of both {\fontfamily{lmtt}\selectfont pose} and {\fontfamily{lmtt}\selectfont pixel} to noise in the contact shapes.
    \method, in comparison, is more robust to contact shape noise, particularly in the parallel jaw contact case. With simulated contacts, the median normalized error for the parallel jaw case of \methodd shrinks about two times, dropping to 0.03. 
    Furthermore, despite the better performance of {\fontfamily{lmtt}\selectfont pose} and {\fontfamily{lmtt}\selectfont pixel} on simulated contacts compared with real contacts, \methodd yields the best performance on simulated contacts as well. With simulated contacts, \methodd is more than twice as good as the next best baseline ({\fontfamily{lmtt}\selectfont pixel} with parallel jaw contacts). 
    %
    %
    
    The primary reason for {\fontfamily{lmtt}\selectfont pixel}'s sensitivity to noise in the contact shapes is that it only evaluates the exact 2D location of the contact on the sensor. \method, in comparison, is trained to match contact shapes resulting from slight perturbations of grid poses to the closest match on the grid. This makes\methodd more robust to slight discrepancies between observed and grid contact shapes. 
    %
    
    The sensitivity of {\fontfamily{lmtt}\selectfont pose} to contact shape noise has to do, instead, with the method's ability to generalize to out of distribution data in the form of real contact shapes. Both {\fontfamily{lmtt}\selectfont pose} and \methodd are trained on simulated versions of contact shapes, but \methodd generalizes to real contact shapes much better. \methodd computes the likelihood of an observed contact matching to a discrete set of simulated shapes. Therefore, the match does not need to be exact, just better than other possible options, for the localization to be accurate. The structure imposed by the grid therefore improves the ability of \methodd to generalize to real, noisy contact shapes. Because {\fontfamily{lmtt}\selectfont pose} regresses object poses directly, without the structure of the grid, its performance is more brittle and degrades when using real contacts.
    
    The normalized error of {\fontfamily{lmtt}\selectfont classification}, even with simulated contacts, is 0.96, remaining similar to selecting a pose randomly from the grid. The size of the grid is too large to be handled with {\fontfamily{lmtt}\selectfont classification} (as addressed in Section \ref{sec:baseline_sim}), so it is not possible to comment on the impact of noisy contacts.
    %
    
    For the 5 objects we evaluate ({\fontfamily{lmtt}\selectfont long grease}, {\fontfamily{lmtt}\selectfont snap ring}, {\fontfamily{lmtt}\selectfont big head}, {\fontfamily{lmtt}\selectfont hanger}, and {\fontfamily{lmtt}\selectfont cotter}), \methodd performs significantly better than random (here, we define significantly better as more than twice as good, which can be identified as a median normalized error less than 0.5 in Table \ref{table:baseline_results}) for 4/5 objects in both the single contact and parallel jaw contact cases. In the parallel jaw contact case, the fifth object, {\fontfamily{lmtt}\selectfont cotter}, has median normalized error of 0.51, which is just below twice as a good as random. In comparison, {\fontfamily{lmtt}\selectfont pose}, {\fontfamily{lmtt}\selectfont classification}, and {\fontfamily{lmtt}\selectfont pixel} with a single contact are only significantly better than random for 1/5 objects. {\fontfamily{lmtt}\selectfont Pixel} with parallel jaw contacts is significantly better than random for 3/5 objects. 
    Furthermore, with a single contact, \methodd outperforms all baselines for 4/5 objects. \methodd is between 1.5 and 4 times better than the next best baseline for all objects except for {\fontfamily{lmtt}\selectfont long grease}, which performs marginally worse than {\fontfamily{lmtt}\selectfont pose} in the single contact case (0.72 median normalized error for {\fontfamily{lmtt}\selectfont pose}, compared with 0.76 for \method). With parallel jaw contacts, \methodd outperforms {\fontfamily{lmtt}\selectfont pixel}, which is the only other method compatible with parallel jaw contacts, with  by between 1.5 and 9 times, depending on the object.
    
    A final observation is that for some objects, parallel jaw contact versions of {\fontfamily{lmtt}\selectfont pixel} and \methodd are more robust to contact shape noise than single contact versions. We consider {\fontfamily{lmtt}\selectfont long grease} (results listed in Table~\ref{table:baseline_results}) as an example. \methodd with a single real contact has 0.76 median normalized error, compared with 0.09 for the parallel jaw contact case. This means that the localization performance improves around eight times when using parallel jaw contacts. Similarly, {\fontfamily{lmtt}\selectfont pixel} with a single contact has 0.93 median normalized error, compared with 0.17 median normalized error with parallel jaw contacts; the localization performance is about 5.5 times better in the parallel jaw contact case. 
    We also consider the performance of \method, and {\fontfamily{lmtt}\selectfont pixel}, on simulated contacts to evaluate the impact of contact shape noise on the discrepancy between single and parallel jaw contact cases. With simulated contacts, the localization performance is the same for \methodd for both a single contact and parallel jaw contacts; the median normalized error is 0.03 in both cases. The outcome is similar for {\fontfamily{lmtt}\selectfont pixel}; the median normalized error is 0.05 in both the single contact and parallel jaw contact cases. The consistent localization performance between methods and contact configurations when using simulated contacts indicates that contact shape noise is the differentiating factor between single contact and parallel jaw contact cases when using real contacts. This means that robustness to noise is a key advantage of using parallel jaw contacts over a single contact, for both methods.
    Both \methodd and {\fontfamily{lmtt}\selectfont pixel} have similar performance in the parallel jaw case with real contacts as they do with idealized (simulated) contacts; it is 3 (\methodd) to 3.5 ({\fontfamily{lmtt}\selectfont pixel}) times easier to localize simulated contacts than real contacts. The discrepancy between simulated and real results is much more pronounced in the single contact case of both methods. This implies that the sim-to-real gap can be bridged, in part, by the inclusion of parallel jaw information.

    We conclude this section with some remarks about the challenge of estimating an object's pose from a single noisy contact. \methodd outperforms all baselines on 4/5 objects with a single contact, and all objects with parallel jaw contacts. Ultimately, though, even \methodd struggles to perform significantly better than selecting a pose at random from the grid with a single contact for 1/5 objects (where significantly better than random is defined as more than twice as good). Estimating the pose of an object from a single, noisy, often non-unique contact is challenging. Therefore, perhaps the most important feature of \methodd is that it outputs meaningful pose distributions over possible object poses (Figure~\ref{fig:long_pencil_figure}). This creates a natural framework for incorporating constraints from additional contacts, measurements of the robot state (such as the gripper opening), information from additional sensing modalities, or even previous tactile estimates of the object pose.

\section{Discussion}
This paper presents an approach to tactile pose estimation for objects with known geometry. 
\methodd relies on learning an embedding completely in simulation that facilitates comparing real and simulated contact shapes. 
We can reconstruct contact shapes with high fidelity, using images from the real sensor and in simulation.
We compare the embeddings of an observed contact shape with those of a precomputed dense set of simulated contact shapes to obtain the distribution over a dense set of possible object poses.
%
%
%
The approach therefore allows to reason over pose distributions and to handle additional pose constraints.
%
%
%
%
%

We evaluate \methodd on real grasp datasets for 20 objects, and report the accuracy for three ablations of \method, corresponding to increasing amounts of information from a given grasp. 
First, we evaluate the accuracy using a single tactile image which corresponds to the case where only one contact is available to the algorithm. 
Second, we consider the case where two tactile images (corresponding to a parallel jaw grasp on the object) and the gripper opening are available. 
Third, we evaluate the accuracy after filtering the distribution obtained using parallel jaw information with a coarse prior on the object pose, approximating the case where information from an additional sensing modality (e.g. vision) is available.

We find that the amount of information needed to localize an arbitrary grasp on an object accurately is highly dependent on the object geometry.
Of the 20 objects we evaluate, 9 can be localized accurately with a single contact. For these objects, more than half of arbitrary grasps on the object can be localized accurately. The objects that can be localized accurately with a single contact tend to be smaller, and have significant regions of unique contacts. 
When including parallel jaw information, the number of objects that can be localized accurately using the best match from \methodd increases to 12. The objects that can be localized accurately after including parallel jaw information tend to have pseudosymmetrical features which can be disambiguated with a second contact, or vary significantly in width across the length of the object, which makes the gripper opening a discriminative source of information. 
After filtering the parallel jaw pose distribution with a coarse 10mm prior on the object pose, the number of objects that can be localized accurately increases to 16.
The objects that can be localized accurately after including a coarse prior on the object pose tend to be larger, and have \textit{discrete nonuninquess} (features that are unique, with a discrete number of exceptions on remote regions of the object).

There are, however, four objects which cannot be localized accurately on average even after filtering the parallel jaw distribution with a 10mm prior and selecting the resulting most likely object pose. These objects are large, and have significant, continuous regions of non-unique contacts. The majority of contacts on such objects do not provide enough information to uniquely determine their pose.

Even objects which can be, on average, localized accurately from an arbitrary grasp have regions of non-unique contacts that do not provide enough information to uniquely determine the object pose.
It is therefore important to evaluate the localization accuracy of specific grasps on objects, as well as arbitrary grasps. To this end, we compare the localization accuracy separated out by grasp approach direction, and the localization accuracy of individual grasps on selected objects.
We find that many objects have one direction which is significantly easier to localize (half as much error or less) than others. By examining individual grasps on the object {\fontfamily{lmtt}\selectfont long pencil}, we find that even objects with large, continuous regions of non-unique contacts have regions that are more unique and easier to localize. 

Both of these breakdowns (sets broken out by grasp approach direction, or individual grasps) could ultimately be leveraged in a grasp planning framework, in which grasps that are easier to localize are specifically targeted. In practice, grasps within a given grasp approach direction could be targeted with a very coarse prior on the object pose, while a specific individual grasp could be targeted with somewhat finer information about the object pose.
Because \methodd is trained entirely in simulation, it is possible to evaluate which grasps lead to lower localization error in advance of encountering the object. For instance, the output pose distributions we obtain using simulated and real contact shapes on {\fontfamily{lmtt}\selectfont long pencil} are qualitatively similar (Figure \ref{fig:long_pencil_figure} and \ref{fig:long_pencil_figure_sim}). Simulated contacts can therefore be used to identify unique and non-unique candidate grasps. This feature, too, could be important to leverage in a manipulation planning framework.

Our approach also assumes we have access to accurate geometric models of objects. We demonstrate that \methodd can be effectively combined with object models reconstructed from a 3D scanner.
A key feature of \methodd is the ability to refine coarse estimates of object pose with high resolution tactile information. We find that moderate shape noise does not significantly compromise that ability for any of the 5 objects we evaluate.

Future work could learn embeddings that are even more robust to shape uncertainty by corrupting the contact shapes with noise during training. This would not inherently improve the localization accuracy when considering only information from a single grasp, but could yield distributions that are more representative of our true confidence in the estimate. These distributions could be combined effectively with other information (e.g. pose priors, additional tactile measurements, or dynamics) to converge on a unique estimate of the object pose.

We demonstrate the advantages of \methodd compared with baseline methods for estimating object pose from tactile images. Compared with a standard classification approach, direct pose regression, and direct pixel comparison, \methodd scales better to larger, and more complex problems (e.g. regressing 6D object pose rather than simply translation or rotation in plane) and is more robust to contact shape noise.~\looseness=-1

In summary, we demonstrate the effectiveness of \methodd at pose estimation for 20 objects with information gathered from a single, real grasp on the object. We consider known and reconstructed object shapes, and show that in both cases \methodd outputs meaningful distributions over object pose that can be maximized directly, or combined with additional information to converge on a unique estimate of object pose. While \methodd is trained entirely in simulation, it is robust to real and noisy contact shapes arising from both sensor noise and object shape noise. This robustness stands in contrast baseline methods, which are more sensitive to contact shape noise. This suggests that matching estimated contact shapes to a dense precomputed set opens the door to moving many computations into simulation, without loss of robustness, and improving how robots learn to perceive and manipulate their environment.

\section*{Acknowledgments}
We thank Ian Taylor and Siyuan Dong for helping with the real setup. We also thank Ferran Alet for providing detailed insights and reviewing the paper.

\bibliographystyle{agsm}
\bibliography{ncd-ijrr18}

\newpage
\appendix

\newpage
\clearpage
\section{Contact shape prediction from tactile observations}\label{sec:append_label}

Given a tactile observation, our goal is to extract the contact shape that produces it. 
To that aim, we train an image translation netwrok (pix2pix) based on \citet{isola2018imagetoimage} to map from RGB tactile observations to contact shapes, described below for completeness.

\begin{figure}[h!t]

\centering
	\includegraphics[width=\linewidth]{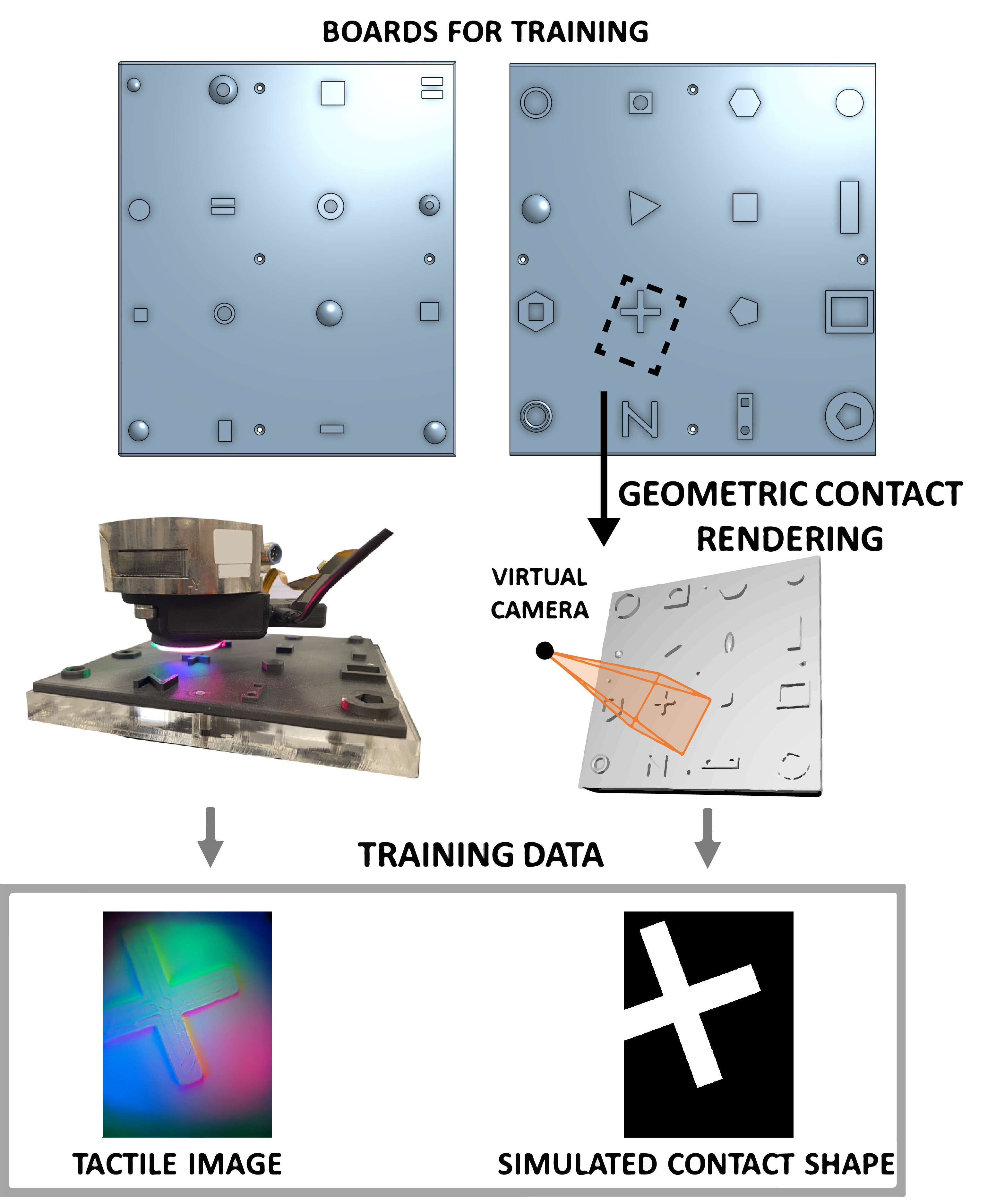}
\centering

\caption{\textbf{Training data set-up to estimate real contact shapes from real tactile observations.} We use the boards at the top to collect real tactile observations from them as well as simulated contact shapes. By placing the boards on a robotic platform, we can move them w.r.t. the sensor to perform calibrated touches on them. From each touch, we recover both a tactile image and a simulated contact shape that we compute using the 3D model of the boards and its pose w.r.t the sensor when the touch happened.} \label{fig:data_collection}
\vspace{-5mm}
\end{figure}

As shown in Figure ~\ref{fig:data_collection}, the input to the NN is an RGB tactile image of size 160x120x3. The output corresponds to a binary contact mask of size 160x120x3, which we flatten to 160x120 to represent the contact shape.\looseness=-1

The training data is collected autonomously in a controlled 4-axis stage that generates controlled touches on known 3D-printed shapes. In our case, we collect the calibration data from two 3D-printed boards with simple geometric shapes on them (see the top of Figure ~\ref{fig:data_collection}). From each controlled touch, we obtain a tactile observation and the pose of the board w.r.t. to the sensor. From this pose, we can later simulate the corresponding contact shape to the tactile observation using geometric contact rendering and the 3D model of the board.

Note that to do tactile localization on any object, we only need to gather calibration data once because the map between tactile observations and contact shapes is object-independent. Empirically, we find that the map is also independent of sensor instance. 

The training input is a 160x240x3 image, consisting of the raw tactile image and the ground truth contact mask stitched together.

Both the generator and descriminator networks of the NN are in the form of convolution-BatchNorm-ReLu modules, as in \citet{isola2018imagetoimage}. The optimizer is Adam with an initial learning rate of 0.0002, that decays linearly. The momentum of Adam is 0.5. The GAN objective is the least squares loss function.

In practice, we collected 10,000 pairs of tactile observations in total (9,000 from the board at left, and 1,000 from the board at right), and trained the NN for 86 epochs.

\section{Collecting ground-truth data for real objects}\label{sec:append_collect}

We collect labelled datasets of tactile observations of grasps on 20 objects mounted in known positions and orientations in the world. Each dataset contains pairs of RGB tactile images, and the corresponding ground-truth object pose relative to the gripper. This allows us to compute pose errors between \methodd predictions from tactile observations, and the true calibrated object poses.

To get calibrated real pairs of contact poses and tactile observations, we first obtain a coarse estimate of the object pose using an overhead camera. We then refine that estimate using \methodd for geometric contact rendering: we move the robot to our coarse estimate of the object pose, grasp the object, and compare the observed contact with the simulated contact shape in the estimated pose. We adjust the estimate of the object pose until the observed contact matches the simulated one. 

\begin{figure}[ht]

\centering
	\includegraphics[width=0.9\linewidth]{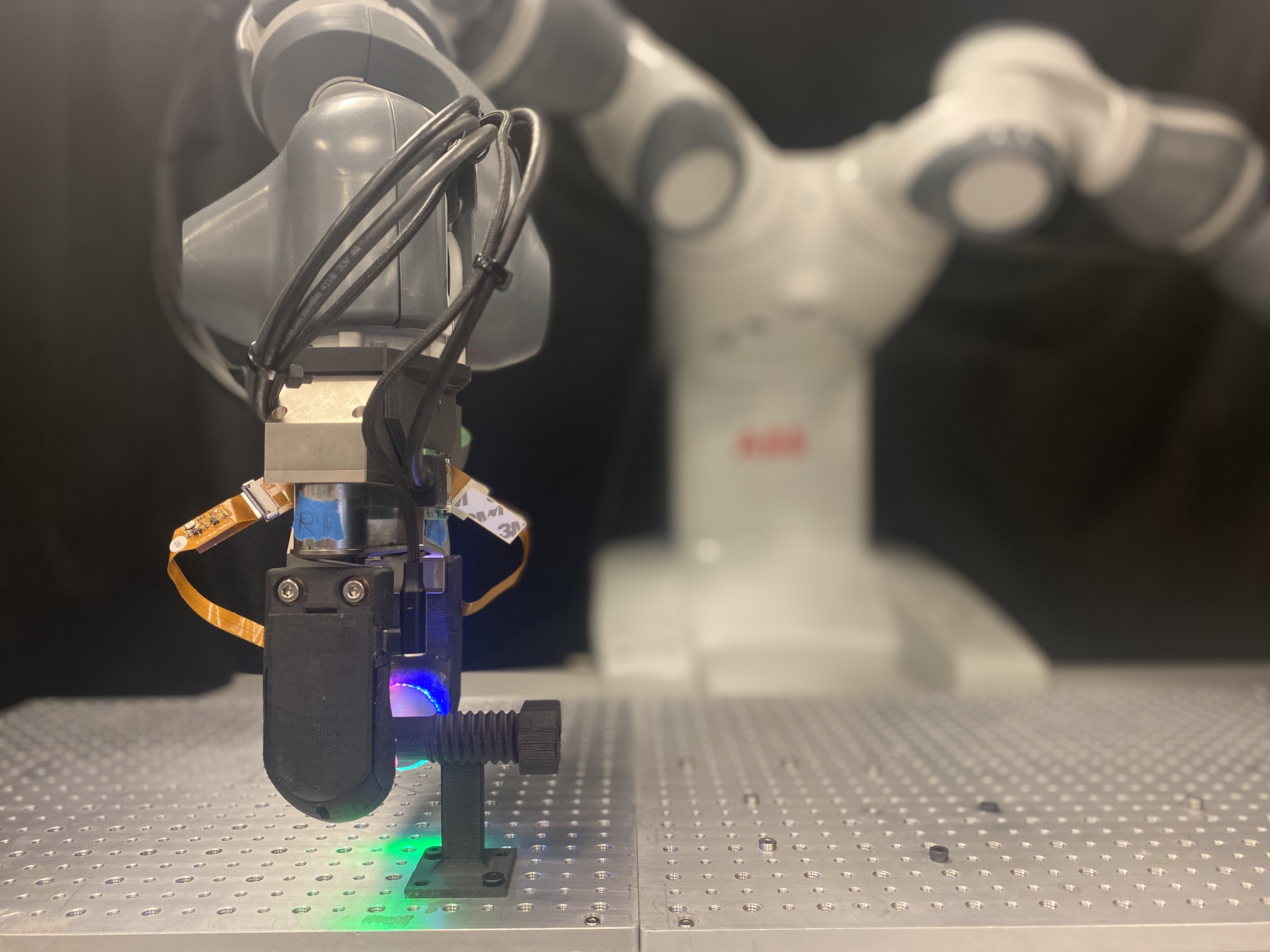}
\centering

\caption{\textbf{Collecting ground-truth data on real objects.} We 3D-print object models mounted on posts and fix them in known location relative to the robot. As a result, we can collect ground-truth data for test objects (the object {\fontfamily{lmtt}\selectfont stud} is shown).} \label{fig:test_data}
\vspace{-5mm}
\end{figure}

The robotic system we use to collect the labelled datasets consists of a dual arm ABB Yumi with two WSG-32 grippers and GelSlim 3.0 tactile sensing fingers.
Figure~\ref{fig:test_data} shows a still of the data collection procedure for the object {\fontfamily{lmtt}\selectfont stud}. 

We collect labelled observations of feasible grasps on each each object. To determine which contacts should appear in the datasets, we first find the \textit{stable poses} of each object. Stable poses are the poses that an object is most likely to fall in when dropped onto a table. We then determine a set of feasible grasp approach directions for each stable pose. We define the \textit{grasp approach direction} as the axis of the grasp during a parallel jaw grasp. 
We break out the labelled datasets by grasp approach direction, to facilitate insight about what categories of grasps are easier to localize for different objects. 
For each grasp approach direction, we collect a set of equispaced observations, that correspond to a given grasp approach direction at various x,y locations relative to the object. In total, these observations correspond to the set of contacts that we are likely to encounter when picking up each object from a table.

\section{Similarity metric for contact shapes}\label{sec:append_sim}

Given a new contact shape, we want to compare it to all pre-computed contact shapes in the grid to find what poses are more likely to produce it. 
To that aim, we use MOCO~\citep{moco}, a widely used tool in contrastive learning, to train a NN that encodes contact shapes into a low dimensional embedding allowing to compare them using the distance between their encodings.

The NN encoder is a pre-trained ResNet-50~\citep{he2016} cropped before the average-pooling layer to preserve spatial information. It encodes contact shapes into vectors of dimension 1000.
To train, we use the same optimizer as~\citet{moco}: stochastic gradient descent with a learning rate that starts at 0.03 and decays over time, momentum of 0.999, and no weight decay.
Each training datapoint comes from selecting a random contact pose, rendering its contact shape, and finding its closest pose in the pre-computed grid. Then our input is the contact shape. The output or label is a vector with all entries zeros except the one that corresponds to the closest element that gets a one. This vector represents the probability of each poses in the grid to be the closest to the target pose. 
The loss function is the cross-entropy loss between the label and the softmax between the target encoder and the ones from the grid (see Fig.~\ref{fig:encoder}). As the NN gets trained, we recompute the encoding of the queue following~\citep{moco}.
We train the models for 15 epochs, as they tend to overfit to simulated data after that. 

To make the training data resemble more the real distribution of contact shapes, we train using as target contact shapes with different depth thresholds, $\Delta d$, selected randomly between 1 and 2 mm. This accounts for changes in the contact force applied to the sensor which can make contacts more or less deep (see Fig.~\ref{fig:encoder}).
Finally, before encoding the contact shapes, we converted them into binary masks to ease the learning process and prevent that numerical mismatches between real and simulated contact shapes.

\section{Multi-contact for N sensors}\label{sec:append_multi}

To derive Equation 1 in~\secref{sec:multi_contact}, we will prove a more general equation for the object pose $x$ when it is in contact with N sensors:
\begin{multline}
P(x | C_1, ..., C_N) =\\ \frac{P(x | C_1) \cdot ... \cdot P(x | C_N) \cdot P_{ta}(x)/ P_{tr}(x)^{N}}{\sum_{x' }P(x' | C_1) \cdot ... \cdot P(x' | C_N) \cdot P_{ta}(x')/ P_{tr}(x')^{N}} \label{eq:full}    
\end{multline}
%
where $P(x | C_i)$ is the likelihood that pose $x$ has produced a contact shape $C_i$ on sensor $i$. $P_{ta}(x)$ is a prior over all the possible contact poses that would result in contact with the N sensors in the current task. $P_{tr}(x)$ is the distribution over the contact poses used to train the similarity function in ~\secref{sec:tactile_mapping}.

Next, we prove Equation~\eqref{eq:full}. Assuming that we have access to the poses of N sensors, given an object pose $x$, the contact shapes $\{C_i\}$ become determined. This allows us to write the joint distribution of pose and contact shapes as:

\begin{multline}
P(x, C_1, ..., C_N) = \\P_{ta}(x) \cdot P(C_1 | x) \cdot ... \cdot P(C_N | x)
\end{multline}

Now we can use Bayes theorem on the $P (C_i | x)$ terms:

\begin{multline}
P(C_i| x )  = P(x | C_i) \cdot P( C_i)/P_{tr}(x)    
\end{multline}

and obtain:

\begin{multline}
P(x, C_1, ..., C_N) =  P(x| C_1) \cdot P(C_1)\cdot ... \\ \cdot P(x| C_N) \cdot P(C_N) \cdot P_{ta}(x)/ P_{tr}(x)^N
\end{multline}

where we take into account that the $P(x|C_i)$ have been trained using a predefined distribution of poses $P_{tr}(x)$ that does not necessarily match $P_{ta}(x)$.
Now, we can compute $P(x| C_1, ... , C_N)$ using again the Bayes theorem:

\begin{multline}
P(x| C_1, ... , C_N) = \frac{P(x, C_1, ... , C_N)}{P(C_1, ... , C_N)} \\= \frac{P(x, C_1, ... , C_N)}{\sum_{x'}P(x',C_1, ... , C_N)}
\end{multline}
%
and thus
\begin{multline}
P(x| C_1, ... , C_N) = \\\frac{\left(\Pi_i^N  P(x| C_i) \cdot P(C_i) \right) \cdot  P_{ta}(x)/ P_{tr}(x)^{N} }{\sum_{x'} \left( \Pi_i^N P(x'| C_i) \cdot P(C_i) \right) \cdot  P_{ta}(x')/ P_{tr}(x')^{N}}    
\end{multline}
We can now cancel the terms $P(C_i)$ that appear both in the numerator and denominator, getting:
\begin{multline}
P(x| C_1, ... , C_N) = \\\frac{\left(\Pi_i^N  P(x| C_i) \right) \cdot  P_{ta}(x)/ P_{tr}(x)^{N} }{\sum_{x'} \left( \Pi_i^N P(x'| C_i) \right) \cdot  P_{ta}(x')/ P_{tr}(x')^{N}}
\end{multline}

This concludes the proof of Equation~\eqref{eq:full}. In practice, we discretize the set of contact poses $\{ x \}$ that result in contact with the N sensors using the grid over of sensor 1 (this is an arbitrary choice). Then, we use the transformation between sensor 1 and sensor $i$ to compute each $P(x| C_i)$ using the closest pose to $x$ in the grid of sensor $i$. Note that often a pose that results in contact with sensor $1$ will not contact sensor $i$. In that case, we do not consider that pose as we are only interested in poses that contact the N sensors. Finally, because the grids are dense and structured, finding the closest pose to $x$ in a grid has a minor effect on performance and is fast to compute.

We can use Equation~\eqref{eq:full} to derive Equation~\eqref{eq:multi-contact} in~\secref{sec:multi_contact} under two extra assumptions. First, we observe that the denominator is constant because all contact shapes $C_i$ are given. Next, if we have no prior over the contact poses during training, then $P_{tr}(x)$ is constant, and that leads to Equation~\ref{eq:multi-contact} repeated here for completeness: 

\begin{multline}
P(x | C_1, ... , C_N) \propto P(x | C_1) \cdot ... \cdot P(x | C_N) \cdot P_{ta}(x)
\end{multline}
\section{Results by grasp approach direction}\label{sec:direction_appendix}
\begin{table}[h]\caption{Results by grasp approach direction for Parallel Jaw ablation of \methodd}\label{table:grasp_approach_directions}\renewcommand{\arraystretch}{1.5}\begin{center}
\resizebox{\columnwidth}{!}{%
\begin{tabular}{|c|c|c c c c|}\bottomrule\multicolumn{2}{|c}{}&\multicolumn{4}{|c|}{Grasp Approach Direction mm (norm)}\\\cline{1-6}& & 1 & 2 & 3 & 4\\\cline{3-6}Long Grease &\makecell{\vspace{3pt}\begin{minipage}{10mm}\centering\includegraphics[height=5mm]{figures/scaled_grasp_figures/size_line_long_grease.png}\vspace{3pt}\end{minipage}} & 1.2 (0.0) & 9.9 (0.3) & &\\Grease &\makecell{\vspace{3pt}\begin{minipage}{10mm}\centering\includegraphics[height=5mm]{figures/scaled_grasp_figures/size_line_grease.png}\vspace{3pt}\end{minipage}} & 0.8 (0.1) & 2.0 (0.2) & &\\Stud &\makecell{\vspace{3pt}\begin{minipage}{10mm}\centering\includegraphics[height=5mm]{figures/scaled_grasp_figures/size_line_stud.png}\vspace{3pt}\end{minipage}} & 13.6 (0.3) & 13.2 (0.3) & &\\Pin &\makecell{\vspace{3pt}\begin{minipage}{10mm}\centering\includegraphics[height=5mm]{figures/scaled_grasp_figures/size_line_pin.png}\vspace{3pt}\end{minipage}} & 5.5 (0.2) & 5.7 (0.2) & &\\Hanger &\makecell{\vspace{3pt}\begin{minipage}{10mm}\centering\includegraphics[height=5mm]{figures/scaled_grasp_figures/size_line_hanger.png}\vspace{3pt}\end{minipage}} & 2.7 (0.1) & 1.1 (0.0) & 1.3 (0.0) & 6.2 (0.2)\\Hydraulic &\makecell{\vspace{3pt}\begin{minipage}{10mm}\centering\includegraphics[height=5mm]{figures/scaled_grasp_figures/size_line_hydraulic.png}\vspace{3pt}\end{minipage}} & 9.1 (0.4) & 1.7 (0.1) & &\\Hose &\makecell{\vspace{3pt}\begin{minipage}{10mm}\centering\includegraphics[height=5mm]{figures/scaled_grasp_figures/size_line_hose.png}\vspace{3pt}\end{minipage}} & 46.1 (0.7) & 37.2 (0.6) & &\\Round Hose &\makecell{\vspace{3pt}\begin{minipage}{10mm}\centering\includegraphics[height=5mm]{figures/scaled_grasp_figures/size_line_round_hose.png}\vspace{3pt}\end{minipage}} & 35.2 (0.7) & 38.9 (0.7) & &\\Holder &\makecell{\vspace{3pt}\begin{minipage}{10mm}\centering\includegraphics[height=5mm]{figures/scaled_grasp_figures/size_line_cable_f.png}\vspace{3pt}\end{minipage}} & 2.0 (0.1) & 2.9 (0.1) & &\\Round Cable &\makecell{\vspace{3pt}\begin{minipage}{10mm}\centering\includegraphics[height=5mm]{figures/scaled_grasp_figures/size_line_cable_g.png}\vspace{3pt}\end{minipage}} & 15.2 (0.8) & 2.2 (0.1) & &\\Couple &\makecell{\vspace{3pt}\begin{minipage}{10mm}\centering\includegraphics[height=5mm]{figures/scaled_grasp_figures/size_line_couple.png}\vspace{3pt}\end{minipage}} & 13.9 (0.6) & 19.6 (0.7) & 21.4 (0.8) &\\Round Couple &\makecell{\vspace{3pt}\begin{minipage}{10mm}\centering\includegraphics[height=5mm]{figures/scaled_grasp_figures/size_line_round_couple.png}\vspace{3pt}\end{minipage}} & 4.8 (0.2) & 14.8 (0.6) & 17.8 (0.7) &\\Cable Clip &\makecell{\vspace{3pt}\begin{minipage}{10mm}\centering\includegraphics[height=5mm]{figures/scaled_grasp_figures/size_line_cable_clip.png}\vspace{3pt}\end{minipage}} & 10.6 (0.6) & 12.1 (0.7) & &\\\hline
\end{tabular}
}
\end{center}\end{table}


\section{Round clip discussion}\label{sec:round_clip_appendix}
\begin{figure}[ht]

\centering
	\includegraphics[width=0.9\linewidth,trim={15mm 0 0 0}]{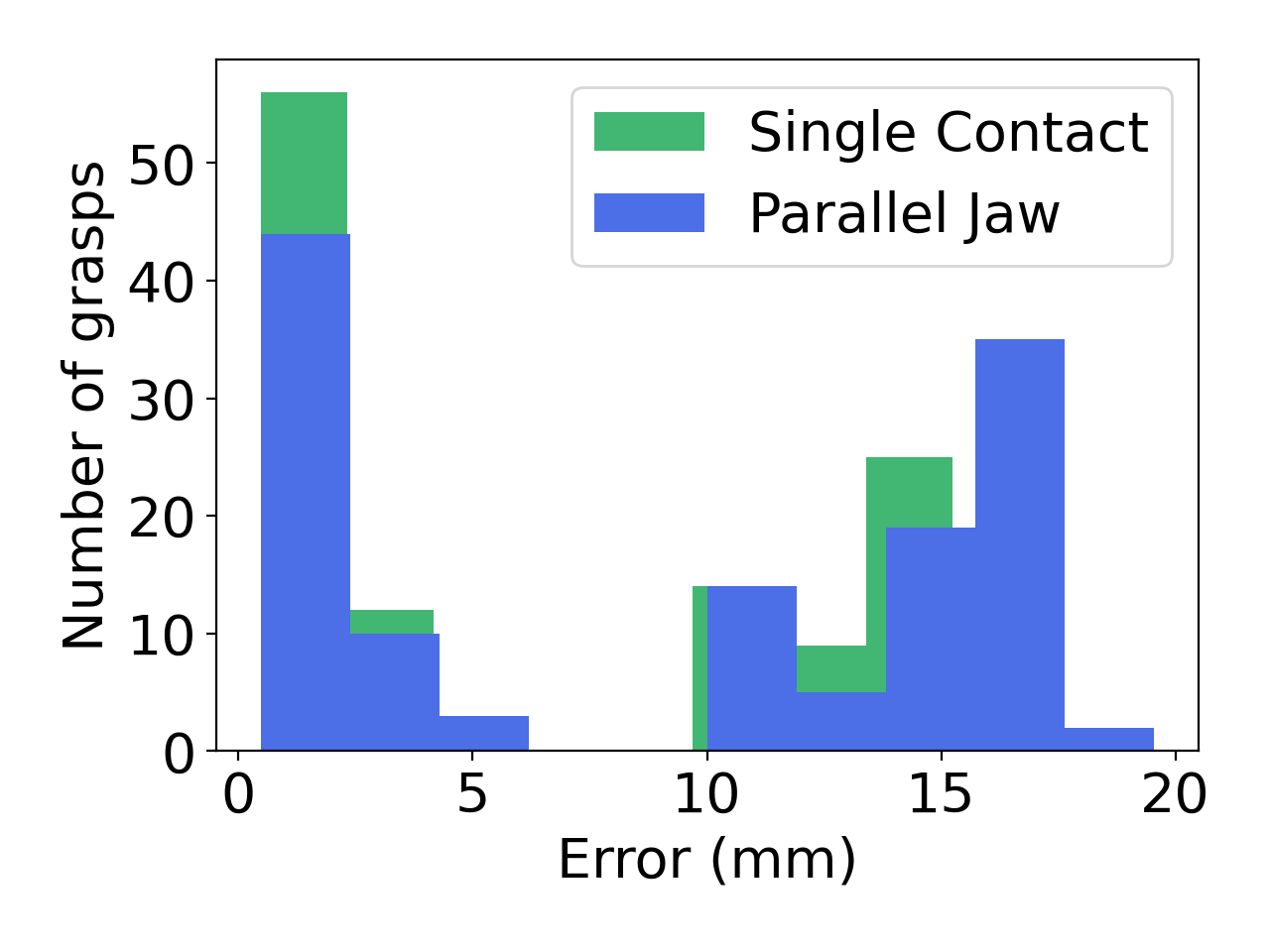}
\centering

\caption{Distribution of errors for single contact vs. parallel jaw contact case for {\fontfamily{lmtt}\selectfont round clip}.} \label{fig:rc_appendix}
\vspace{-5mm}
\end{figure}

Our error metric - median error of the best match - does not capture the error distributions for {\fontfamily{lmtt}\selectfont round clip} well. As discussed in Section \ref{sec: grasp_appoach} and tabulated in Table \ref{table:grasp_approach_directions},  {\fontfamily{lmtt}\selectfont round clip} is much easier to localize from one grasp approach direction than the other. This results in similar, bimodal error distributions in both the single contact and parallel jaw contact cases, as seen in Figure \ref{fig:rc_appendix}. Instead, we compare the means of the two distributions and observe the underlying similarity between the two distributions more clearly: the mean error of the best match in the single contact case is $7.6$ mm, compared with $9.2$ mm in the parallel jaw contact case.\looseness=-1

\end{document}